\documentclass{article} 
\usepackage{iclr2026_conference,times}
\makeatletter
\providecommand{\iclrruler}[1]{}   
\renewcommand{\iclrruler}[1]{}     
\makeatother

\usepackage{amsmath,amsfonts,bm}

\def\eqref#1{equation~\ref{#1}}

\def\1{\bm{1}}

\DeclareMathAlphabet{\mathsfit}{\encodingdefault}{\sfdefault}{m}{sl}
\SetMathAlphabet{\mathsfit}{bold}{\encodingdefault}{\sfdefault}{bx}{n}

\newcommand{\R}{\mathbb{R}}

\usepackage{hyperref}
\usepackage{url}
\usepackage[utf8]{inputenc} 
\usepackage[T1]{fontenc}    
\usepackage{hyperref}       
\usepackage{booktabs}       
\usepackage{amsfonts}       
\usepackage{nicefrac}       
\usepackage{microtype}      
\usepackage{xcolor}         

\usepackage{bm}
\usepackage{graphicx}
\usepackage{subfigure}
\usepackage{dsfont}
\usepackage{amsmath}
\usepackage{amsthm}
\usepackage{natbib}
\usepackage[toc,page]{appendix}
\newtheorem{theorem}{Theorem}

\newtheorem{definition}{Definition}
\newtheorem{condition}{Condition}
\usepackage{amsmath, amssymb}
\newtheorem{lemma}{Lemma}
\theoremstyle{plain}
\usepackage{mathtools}
\newtheorem{corollary}[theorem]{Corollary} 
\theoremstyle{remark}
\newtheorem{remark}{Remark}

\newtheorem{assumption}{Assumption}

\usepackage[capitalize,noabbrev]{cleveref}

\usepackage{multirow}
\usepackage[inline]{enumitem}
\newenvironment{inlinelist}{\begin{enumerate*}[label=\emph{(\roman{*})}]}{\end{enumerate*}}

\title{Membership Privacy Risks of Sharpness Aware Minimization}

\author{%
  Young In Kim \\
  Department of Computer Science\\
  Purdue University\\
  West Lafaytte, IN 47906, USA\\
  \texttt{kim3531@purdue.edu} \\
  \And
  Andrea Agiollo \\
  Department of Computer Science\\
  Delft University of Technology\\
  Delft, Netherlands\\
  \texttt{A.Agiollo-1@tudelft.nl}\\
  \And
  Pratiksha Agrawal \\
  Department of Computer Science\\
  Purdue University\\
  West Lafaytte, IN 47906, USA\\
  \texttt{agraw180@purdue.edu} \\
  \And
  Johannes O. Royset \\
  Department of Industrial and \\
  Systems Engineering\\
  University of Southern California\\
  Los Angeles, CA 90015, USA\\
  \texttt{royset@usc.edu} \\
  \And
  Rajiv Khanna\\
  Department of Computer Science\\
  Purdue University\\
  West Lafaytte, IN 47906, USA\\
  \texttt{rajivak@purdue.edu} \\
}

\begin{document}

\maketitle

\begin{abstract}
Optimization algorithms that seek flatter minima, such as Sharpness-Aware Minimization (SAM), are credited with improved generalization and robustness to noise. We ask whether such gains impact membership privacy. Surprisingly, we find that SAM is more prone to Membership Inference Attacks (MIA) than classical SGD across multiple datasets and attack methods, despite achieving lower test error. This suggests that the geometric mechanism of SAM that improves generalization simultaneously exacerbates membership leakage. We investigate this phenomenon through extensive analysis of memorization and influence scores. Our results reveal that SAM is more capable of capturing atypical subpatterns, leading to higher memorization scores of samples. Conversely, SGD depends more heavily on majority features, exhibiting worse generalization on atypical subgroups and lower memorization. Crucially, this characteristic of SAM can be linked to lower variance in the prediction confidence of unseen samples, thereby amplifying membership signals. Finally, we model SAM under a perfectly interpolating linear regime and theoretically show that sharpness regularization inherently reduces variance, guaranteeing a higher MIA advantage for confidence and likelihood ratio attacks.
\end{abstract}

\section{Introduction}

Sharpness-Aware Minimization (SAM) has emerged as a prominent optimization technique for improving generalization in deep learning by encouraging flatter minima -- i.e., similar loss values for weight perturbations of certain degree around the optima -- in the loss landscape~\citep{norton2021diametrical,foret2020sharpness,wu2020adversarial,kim22fisher,du2022sharpnessaware,kwon21b}. Flatter optima have been linked to robustness to noise and improved test performance~\citep{chen2023doessharpnessawareminimizationgeneralize,foret2020sharpness, baek2024samrobustlabelnoise}, while a tension exists between whether SAM's implicit bias is geared more towards diversity~\citep{springer2024sharpnessawareminimizationenhancesfeature} or simplicity~\citep{Andriushchenko2023LowRank, chang2025unifiedstabilityanalysissam} of features.

Models that generalize well are thought to rely less on memorizing specific training examples, which should also improve privacy. Consider membership inference attacks (MIAs) in which an attacker exploits the model behavior gap between training and unseen data to infer if a data point was part of the training data or not~\citep{shokri2017membership}. Intuitively, when a model strongly overfits (training error $\ll$ test error), MIA would become easier. \citet{MIAandOverfitting} formally showed that, under certain assumptions, the advantage of a threshold-based MIA is upper bounded by the model’s generalization error. In light of this, one would naturally expect that a technique like SAM -- which demonstrably improves generalization -- should also decrease a model’s susceptibility to MIAs. 

Contrary to this expectation, we find that models trained with SAM are actually \emph{more} vulnerable to MIAs than SGD consistently across diverse datasets and attack methods, even as they achieve better generalization (see \Cref{dsq-attacks,tab:mia_online}). Furthermore, this finding challenges the notion that \emph{flatter minima=good} from a privacy standpoint, calling for a deeper investigation into the relationship between generalization, memorization, and privacy to unearth this phenomenon both empirically and theoretically. Our work is the first to systematically demonstrate higher membership privacy leakage for a sharpness-based algorithm known to generalize better, connecting loss sharpness to MIA vulnerability. We note that there have been previous works that exhibit utility–privacy tradeoffs~\citep{long2018understanding,carlini2021extracting,relaxloss,ccl} or demonstrate that higher generalization does not necessarily decrease privacy leakage~\citep{pmlr-v139-kaya21a,Del_Grosso_2023}.  

\begin{table}[t]
  \caption{Attack accuracy of direct threshold MIA on SGD and SAM showing tradeoffs in test accuracy and MIA attacks. In \textcolor{blue}{blue} we highlight the best performing model on the test set and in \textcolor{red}{red} the model against which MIA is more successful. SAM is more prone to direct threshold attacks.}
  \label{dsq-attacks}
  \centering
  \begin{small}
  \begin{tabular}{ccccccc}
    \toprule 
    Dataset     & Algo  & NN   & Confidence    & Entropy & M-entropy &Test Acc\\
    \midrule
    \multirow{2}{*}{CIFAR-100}  & SGD    & 76.62\%        & 77.19\%  &76.61\%  & 77.30\% &80.30\%\\
                
               & SAM     & \textcolor{red}{77.99\%}        & \textcolor{red}{79.10\%} & \textcolor{red}{78.66\%}  & \textcolor{red}{79.25\%} & \textcolor{blue}{81.60\%} \\

    \midrule
    \multirow{2}{*}{CIFAR-10}  & SGD    & 50\%     &59.37\%   &59.09\% &59.51\% &96.00\%\\                 
               & SAM      & 50.08\%      & \textcolor{red}{61.64\%}   & \textcolor{red}{61.64\%} & \textcolor{red}{61.70\%} & \textcolor{blue}{96.72\%} \\ 
                           
    \midrule
    \multirow{2}{*}{Purchase-100} & SGD     &66.00\%        &66.76\%   &64.78\% &67.13\% &85.50\%\\                    
              & SAM    & \textcolor{red}{66.62\%}     & \textcolor{red}{67.30\%}   & \textcolor{red}{65.35\%} & \textcolor{red}{67.54\%} & \textcolor{blue}{85.54\%}\\

    \midrule
    \multirow{2}{*}{Texas-100}    & SGD        & \textcolor{red}{59.81\%}         &65.20\%   &55.74\% &65.13\% &50.83\%\\             
               & SAM        &59.56\%      & \textcolor{red}{66.59\%} & \textcolor{red}{57.14\%} & \textcolor{red}{65.42\%} & \textcolor{blue}{51.34\%} \\   
                
    \midrule
    \multirow{2}{*}{EyePacs}    & SGD        &73.62\%           &73.40\%  &68.50\% &73.40\% &73.67\%\\   
              & SAM        & \textcolor{red}{77.73\%}        & \textcolor{red}{77.07\%}  & \textcolor{red}{73.37\%} & \textcolor{red}{77.36\%} & \textcolor{blue}{75.41\%} \\

    \bottomrule
  \end{tabular}
  \end{small}
\end{table}

The difference in model behavior on data points that were part of the training set versus those that were not can be more precisely quantified using \emph{memorization scores} \citep{feldman2020does, feldman2020neural}, defined via Leave-One-Out (LOO) error. Memorization scores measure the change in model performance when a specific training sample is removed, and thus serve as a proxy for how much the model has memorized that sample. Motivated by this connection, we analyze the memorization scores of samples trained with SAM and find that \emph{SAM exhibits more memorization than SGD}, indicating a stronger reliance on individual training samples. While this increased memorization provides a plausible explanation for SAM’s heightened vulnerability to MIA, it raises a key question: "How can a model that memorizes more generalize better?"

We hypothesize that the answer lies in \emph{what} is being memorized. Under overparameterization  \citep{allen2019learning}, models can learn not only noise in the data, but also atypical patterns in under-represented
subpopulations through memorization---i.e., few white tiger images with numerous yellow tiger images. This distinction is important as real world datasets are known to have a long tail of such rare subclasses \citep{feldman2020does}. We conjecture that SAM is capable of doing more structured memorization, selectively focusing on atypical subclass patterns, which contributes positively to generalization. Corroborating this hypothesis, we observe that SAM’s memorization score distribution is concentrated in the mid range, rather than the high end---which is typically associated with noise memorization (refer to~\Cref{ssec:mem_analysis}). This suggests that SAM emphasizes samples that are neither trivially learned nor purely noisy, but instead represent rare, but generalizable sub-patterns.

To further validate this finding, we analyze influence scores---which measure the impact of individual training samples on test predictions (see~\Cref{ssec:infl_vs_mem}). Our results show that, for SAM, samples corresponding to moderate memorization exert higher influence on test predictions compared to SGD, confirming that SAM’s generalization gains derive from its ability to better capture rare sub-patterns. Conversely, for SGD, the lower influence of such points implies that the majority pattern is learned dominantly: since this feature is redundant across many samples, the marginal influence of any single point is diluted. These results seem to suggest more that SAM's implicit bias is towards diversity as opposed to simplicity. 

We support our intuition further by introducing a novel metric that quantifies the degree of memorization involved in predicting a test sample in~\Cref{sec:doesSAMmem}. Using this metric, we dissect SAM’s performance gains and identify that SAM’s improvements mostly stem from its performance on atypical test samples that depend heavily on a handful of memorized training points. Meanwhile, SGD performs slightly better on typical samples that rely more on broadly learned features. Stronger influence of minority samples can lead to greater membership privacy risk due to the increased retention of information an attacker can exploit. 

However, what intrinsic mechanism drives this structured memorization? Results for confidence threshold attack indicates that SGD produces more predictions with extreme confidence  compared to SAM. These samples reside beyond the threshold and are source of attacker's error. This suggests a distinct geometric effect: \emph{SAM induces a shrinkage in the variance of the model's output predictions}. This property translates to structured memorization and suppression of majority sub-feature. Relying heavily on a single/few majority subclass feature to classify diverse inputs requires the model to assign large weights to that feature. Geometrically, this creates a steep decision boundary and, consequently, high output variance under perturbation. By penalizing this sharpness, SAM prohibits the amplification of the majority feature, effectively forcing the model to distribute its reliance across diverse, subclass-specific features.

Completing this conjecture, we provide a theoretical foundation for this mechanism in~\Cref{sec:thms}. We prove that the interpolating solution favored by sharpness-aware geometry inherently reduces the variance of the output logits. We then demonstrate how this variance reduction amplifies the attacker's advantage for both confidence-based and likelihood ratio attacks. Our proofs highlight a strong result: \emph{SAM is more vulnerable at any fixed threshold}. We empirically corroborate this with ROC curves in Figure~\ref{fig:roc}. Lastly, we theoretically analyze a dataset with majority and minority subclasses and show how capturing minority feature better leads to enhanced generalization in Section~\ref{app:msa_theory}.

\paragraph{Contributions}
In summary, our contributions are the following: 
\begin{inlinelist}
  \item we are the first to empirically demonstrate that SAM-trained models exhibit higher membership privacy risk than SGD-trained models, serving as a cautionary tale against \emph{flatter minima=good} notion from a privacy standpoint; 
  \item we offer a detailed and conceptually grounded analysis of the root causes of SAM's generalization-memorization relationship, suggesting SAM's implicit bias towards diversity;
  \item we introduce a novel methodology to dissect generalization gains, proving that SAM’s generalization gains stem from its performance on unseen atypical samples;
  \item we theoretically show variance shrinkage effect of interpolating sharpness-aware solutions and how it increases MIA risk for both confidence and likelihood ratio attacks; and
  \item we theoretically formulate a data distribution composed of subclasses where stronger alignment with minority subclass features enhances generalization.
\end{inlinelist}

\begin{table}[t]
\centering
\caption{Comparison of online shadow-model MIA on SGD and SAM. In \textcolor{blue}{blue} we highlight the best performing model on the test set, and in \textcolor{red}{red} the model with higher privacy leakage (higher AUC, Attack Accuracy, and TPR@0.1\%FPR). SAM is more prone to shadow-model attacks.}
\label{tab:mia_online}
\begin{small}
\setlength{\tabcolsep}{4pt}
\begin{tabular}{llcccccccc}
\toprule
 & & \multicolumn{4}{c}{SGD} & \multicolumn{4}{c}{SAM} \\
\cmidrule(lr){3-6} \cmidrule(lr){7-10}
Dataset & Attack & Test Acc & AUC & Attack Acc & TPR@.1 & Test Acc & AUC & Attack Acc & TPR@.1 \\
\midrule

\multirow{2}{*}{CIFAR-100} 
 & RMIA  & \multirow{2}{*}{67.7\%} & 90.4\% & 80.8\% & 21.0\% & \multirow{2}{*}{\textcolor{blue}{69.1\%}} & \textcolor{red}{91.6\%} & \textcolor{red}{82.2\%} & \textcolor{red}{23.4\%} \\
 & LiRA  &                         & 92.6\% & 82.9\% & 27.0\% &                                            & \textcolor{red}{93.7\%} & \textcolor{red}{84.1\%} & \textcolor{red}{31.0\%} \\
\midrule

\multirow{2}{*}{CIFAR-10} 
 & RMIA  & \multirow{2}{*}{92.3\%} & 71.4\% & 63.5\% & 4.8\% & \multirow{2}{*}{\textcolor{blue}{93.1\%}} & \textcolor{red}{74.9\%} & \textcolor{red}{65.9\%} & \textcolor{red}{6.7\%} \\
 & LiRA  &                         & 73.0\% & 64.2\% & 8.8\% &                                            & \textcolor{red}{76.4\%} & \textcolor{red}{66.7\%} & \textcolor{red}{12.5\%} \\
\midrule

\multirow{2}{*}{Purchase100} 
 & RMIA  & \multirow{2}{*}{76.5\%} & 68.8\% & 62.7\% & 1.5\% & \multirow{2}{*}{\textcolor{blue}{77.4\%}} & \textcolor{red}{70.2\%} & \textcolor{red}{63.7\%} & \textcolor{red}{1.7\%} \\
 & LiRA  &                         & 68.9\% & 62.6\% & 1.4\% &                                            & \textcolor{red}{70.2\%} & \textcolor{red}{63.4\%} & \textcolor{red}{1.6\%} \\
\midrule

\multirow{2}{*}{Texas100} 
 & RMIA  & \multirow{2}{*}{46.9\%} & 79.8\% & 70.8\% & 2.9\% & \multirow{2}{*}{\textcolor{blue}{49.2\%}} & \textcolor{red}{80.6\%} & \textcolor{red}{71.5\%} & \textcolor{red}{2.8\%} \\
 & LiRA  &                         & 80.8\% & 71.3\% & 6.9\% &                                            & \textcolor{red}{81.6\%} & \textcolor{red}{72.0\%} & \textcolor{red}{8.2\%} \\

\bottomrule
\end{tabular}
\end{small}
\end{table}

\section{Background \& Preliminaries}
\paragraph{Memorization \& Influence scores}
For a training algorithm $\mathcal{A}$ that is used to train the model $f(\cdot)$ using dataset $\mathcal{D}$ = $((\mathbf{x}_1, y_1),...,(\mathbf{x}_n,y_n))$, the amount of label memorization by $\mathcal{A}$ on a sample ($\mathbf{x}_i, y_i$) $\in$ $\mathcal{D}$ is defined by \Cref{eq:mem}. The probability is taken over randomness of the algorithm such as weight initialization. 
\begin{align} \label{eq:mem}
    mem(\mathcal{A}, \mathcal{D}, i) := \underset{f \leftarrow \mathcal{A(D)}}{Pr}[f(\mathbf{x}_i)=y_i] - \underset{f \leftarrow \mathcal{A}(\mathcal{D} \setminus {(\mathbf{x}_i, y_i)})}{Pr}[f(\mathbf{x}_i)=y_i]
\end{align}
\\
Influence score of a training example $(\mathbf{x}_i, y_i)$ on test example $(\mathbf{x}_j', y_j')$ is defined by:
\begin{align}\label{eq:infl}
    infl(\mathcal{A,D,}i,j) = \underset{f \leftarrow \mathcal{A(D)}}{Pr}[f(\mathbf{x}_j')=y_j'] - \underset{f \leftarrow \mathcal{A}(\mathcal{D} \setminus {(\mathbf{x}_i, y_i)})}{Pr}[f(\mathbf{x}_j')=y_j']
\end{align}

\paragraph{Sharpness Aware Minimization (SAM)} 
Consider a model $f$ : $X \rightarrow Y$ parameterized by a weight vector $\mathbf{w}$ and a per-sample loss function $l$: $W \times X \times Y \rightarrow R_+$. Given a dataset S = \{($\mathbf{x}_1$, $y_1$),..., ($\mathbf{x}_n$, $y_n$)\} sampled i.i.d. from a data distribution, the training loss is defined as $L_S(\mathbf{w}) = \sum_{i=1}^{n} l(y_i, f(\mathbf{x}_i, \mathbf{w}))/n$. Sharpness Aware Minimization combines traditional loss with sharpness term to minimize the difference between maximum loss in the vicinity (a Ball of radius $\rho$: $B(\rho)$) of the current minima. Formally, it is defined as the following: 
\begin{equation}
L_{SAM}(w) = \min_w L_S(\mathbf{w}) + [\max_{\epsilon \in B(\rho)}L_S(\mathbf{w}+\epsilon) - L_S(\mathbf{w})] = \min_w \max_{\epsilon \in B(\rho)}L_S(\mathbf{w}+\epsilon)
\end{equation}
               
\subsection{Membership Inference Attacks} \label{sec:MIattacks}
Consider a victim model $f_{v}$ trained on dataset $D \sim \mathcal{D}$ and attack model $f_{a}$. In a black-box setting, an attacker infers whether a sample $(\mathbf{x},y)$ belongs to $D$ (IN) or not (OUT). In this paper, we consider two types of attacks: \emph{Direct threshold attacks}, which directly learns a threshold from obtained member/non-member data, such as confidence and entropy attacks; and \emph{Shadow model attacks}, which train proxy models to calibrate membership scores, such as Likelihood Ratio Attack (LiRA) and Robust Membership Inference Attack (RMIA) \citep{lira, rmia}.

We quantify privacy risk using the empirical attack accuracy, defined as the average of the true positive (TPR) and true negative rates (1-FPR):
\begin{equation}\label{eq:mia_acc}
\small
\mathrm{Acc}_{\mathrm{MIA}}
= \frac{1}{2n_{\mathrm{m}}}
    \sum_{i=1}^{n_{\mathrm{m}}}
      \mathds{1}\big[ f_{\mathrm{a}}(\mathbf{x}_i,y_i) = 1 \big] +\frac{1}{2n_{\mathrm{nm}}}
    \sum_{j=1}^{n_{\mathrm{nm}}}
      \mathds{1}\big[ f_{\mathrm{a}}(\mathbf{x}_j,y_j) = 0 \big],
\end{equation}
where $n_m, n_{nm}$ are the counts of IN and OUT samples. Additionally, metrics such as Area Under ROC Curve (AUC) and TPR at low FPR are employed to characterize vulnerability. Further details are provided in \Cref{app:mia_attacks}.

\section{Privacy Risks of SAM}\label{sec:sam_mia_exp}

Inspired by the link between SAM and generalization and how MIAs should exploit poor generalization, we here scrutinize the membership privacy risk of SAM by comparing the membership attack accuracy (see \Cref{eq:mia_acc} and \Cref{tab:mia_online}) of different MIAs against SAM- and SGD-trained models across five different benchmark datasets for direct threshold attacks and four different benchmark datasets for shadow model attacks. 

We utilize datasets and target models that are widely employed in studies on MIAs and defenses \citep{MIAandOverfitting, centerbasedrelaxedlearningmembership, relaxloss, jia2019memguard}. Furthermore, we assume that the attacker has access to some portion of the training data and non-training data that it uses to train the attack models---a common assumption in the MIA literature.  

\paragraph{Datasets} We use CIFAR-10, CIFAR-100 and Purchase-100 along with two medical datasets Texas-100 and EyePacs. For direct threshold attacks, we follow~\citet{tang2022mitigating} to determine the partition between training and test data and to determine the subset that constitutes the attacker's prior knowledge \footnote{We adopt and extend the code in \url{https://github.com/inspire-group/MIAdefenseSELENA}}. For shadow model attacks, we use a different dataset split to account for shadow model training \footnote{We adopt and extend the code in \url{https://github.com/orientino/lira-pytorch}}. More details about the datasets and the experimental setup can be found in \Cref{data}.

\paragraph{Methods} 
For direct threshold attacks, we train a set of models and choose the one achieving highest validation accuracy. We then employ different MIA methods -- namely NN-based, confidence-based, entropy-based and modified entropy-based attacks (see \Cref{app:mia_attacks} for a detailed formulation of each MIA) -- to evaluate the attack accuracy on the target model. For shadow model attacks, we generate 256 random half-splits of the training dataset into member set and non-member set and train a model for each split. One model is chosen as the target model and all the other models are used as shadow models for reference. More details about the experimental settings can be found in \Cref{setup}.

\paragraph{Results} 
\label{sec:res}
\emph{Direct Threshold Attacks}
We report the attack accuracy and test accuracy for each model in \Cref{dsq-attacks}. We observe that while SAM achieves higher generalization performance, it also incurs highest attack accuracy for almost all settings. To further investigate the connection between flatness of minima and membership privacy, we report the results for other sharpness-aware optimizers and custom designed optimizer that explicitly aims to find sharper minima in Section~\ref{app:other_sharp}. The results support a relationship between loss landscape geometry and membership privacy, with other optimizers exhibiting similar behavior. For confidence threshold attack, we observe that SGD model incur higher number of extremely confident predictions compared to SAM for both correctly classified and wrongly classified non-members. Because the threshold is typically set near a high value (i.e. 0.92), non-member samples with high confidence are missed by the attacker. From this observation, we conjecture that variance of the model output is an important factor in MIA risk. To confirm that these results are not model-dependent, we report an ablation study in \Cref{diffarch} and verify that similar findings can be observed for different model architectures over the same datasets. 

\emph{Shadow Model Attacks} 
Table~\ref{tab:mia_online} summarizes the results for online Robust MIA (RMIA) and Likelihood Ratio Attack (LiRA), reporting Attack Accuracy, AUC, and TPR at 0.1\% FPR averaged over 10 attack splits~\citep{rmia,lira}. Consistent with the previous results, SAM achieves higher generalization while incurring non-trivially higher privacy leakage across nearly all settings. We note that the reported test accuracies differ from Table~\ref{dsq-attacks} due to difference in data splits for shadow model training. 

The results for offline attacks are presented in Table~\ref{tab:mia_offline}. Analogous to the online setting, SAM exhibits superior generalization paired with heightened MIA vulnerability. Since shadow model attacks represent the state-of-the-art in membership inference, these results suggest that SAM's vulnerability is not merely an artifact of global threshold shifts. Instead, it points to an intrinsic geometric property of SAM that persists even under rigorous, sample-specific SOTA attacks.

\section{SAM learns Atypical Subclass Features More}
\label{sec:SAMatyp}

To investigate the source of SAM’s increased membership privacy risk, we analyze its optimization behavior through the lens of sample memorization and influence. We follow the procedure of \citet{feldman2020neural} to compute the memorization and influence scores for SAM-trained models on CIFAR-100. We then compare these scores against the publicly available scores for SGD-trained models on the same dataset\footnote{\url{https://pluskid.github.io/influence-memorization/}}, enabling a direct comparison of sample-level behavior between the two optimizers.

\subsection{SAM Memorizes Atypical Sub-patterns More}\label{ssec:mem_analysis}

\begin{figure}
  \centering
  \subfigure[\label{fig:memsgdsamdist}]{
  \begin{minipage}[c]{0.3\textwidth}
  \centering
  \includegraphics[width=0.95\textwidth]{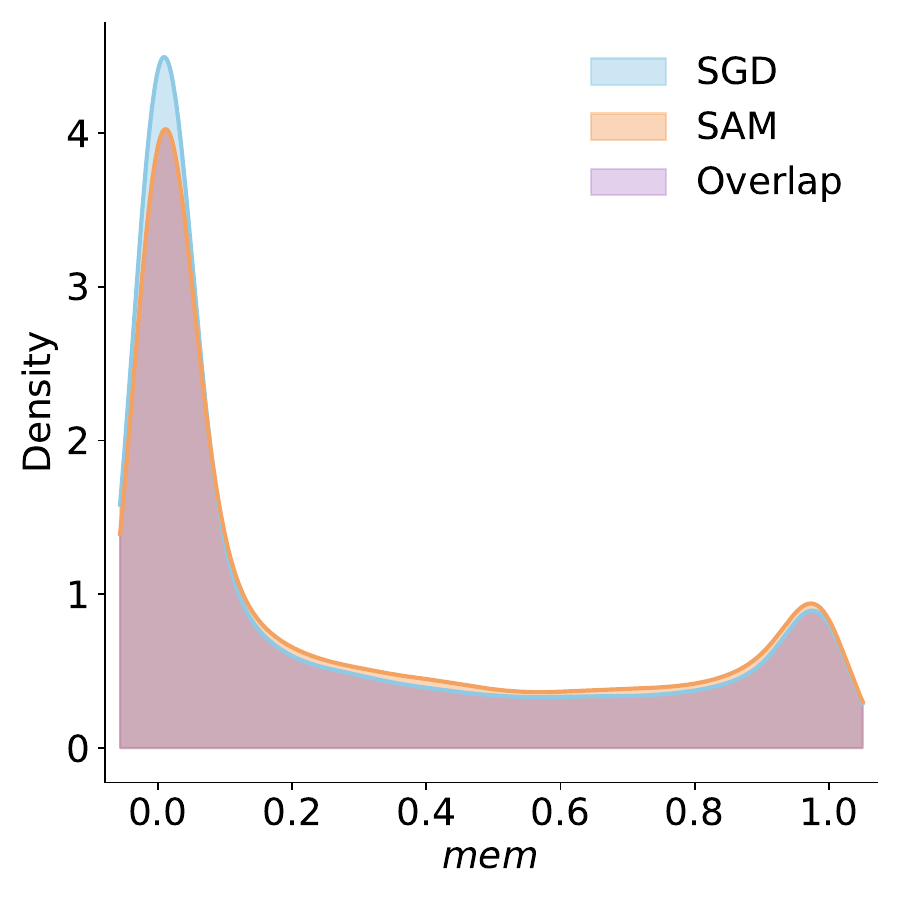}
  \end{minipage}
  }
  \hfill 
  \subfigure[\label{fig:memsgdsamscatter}]{
  \begin{minipage}[c]{0.3\textwidth}
  \centering    
  \includegraphics[width=0.95\textwidth]{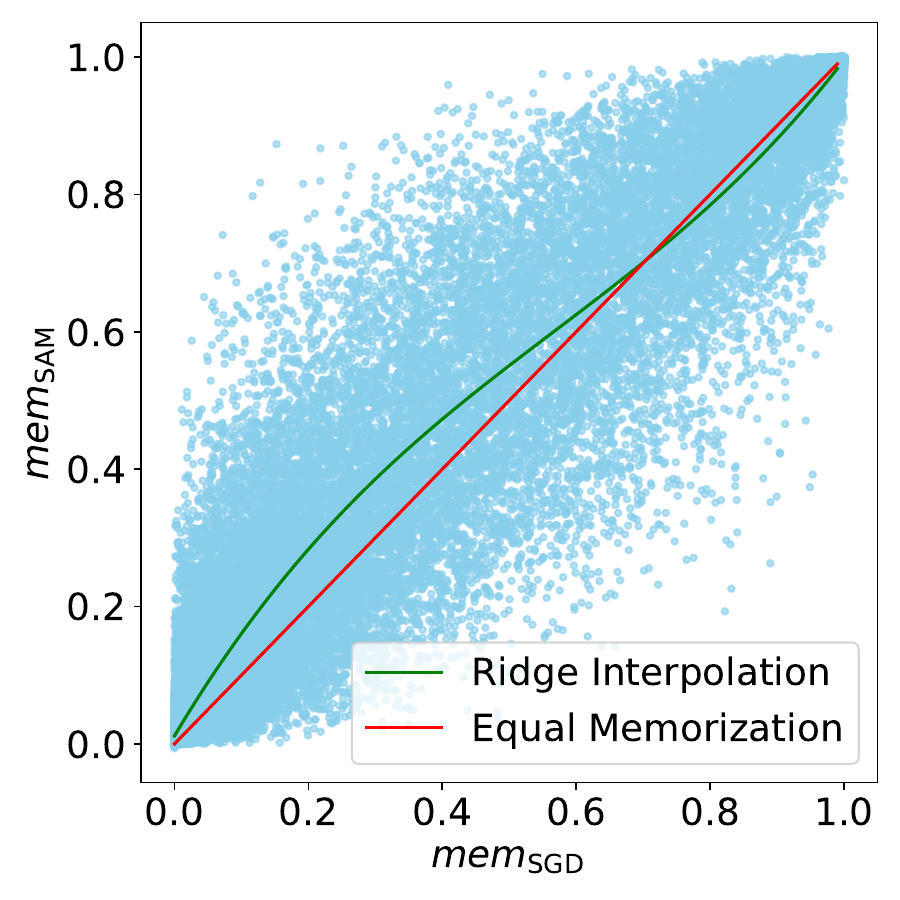}
  \end{minipage}
  }
  \hfill 
  \subfigure[\label{fig:memsgdsamtiger}]{
  \begin{minipage}[c]{0.3\textwidth}
  \centering    
  \includegraphics[width=0.95\textwidth]{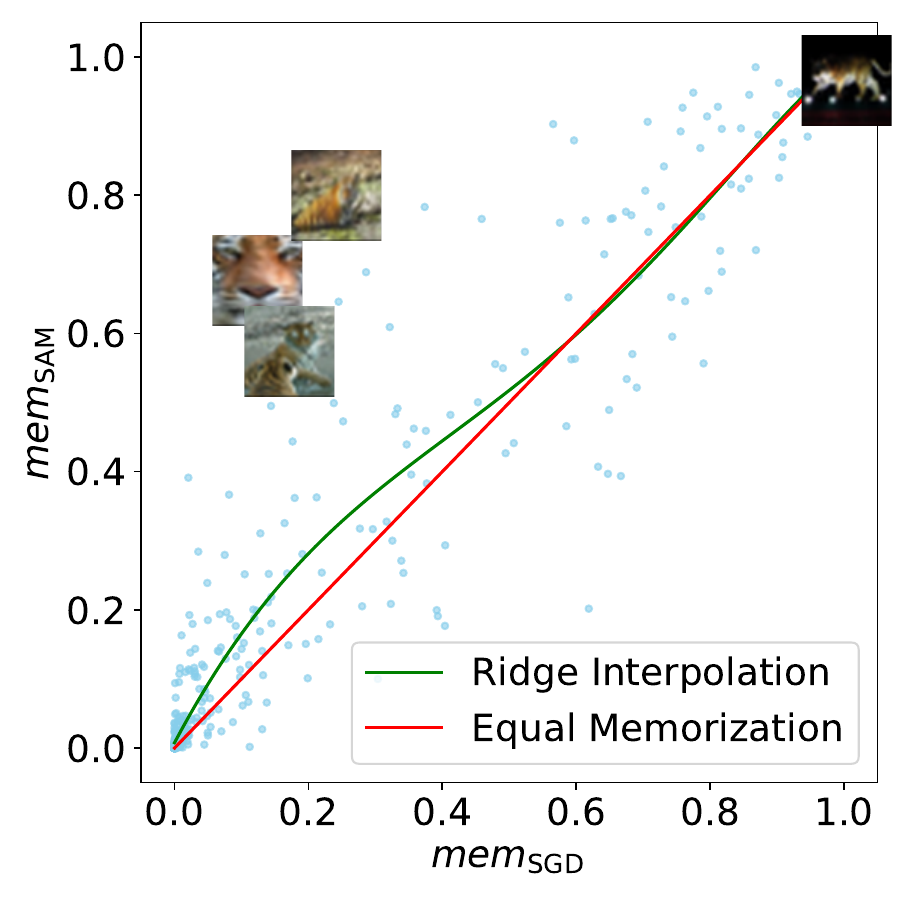}
  \end{minipage}
  }
  \caption{(a): Memorization score density plot for SAM vs SGD. SAM has less density in the lowest range, but more density spread evenly across the remaining range. (b): Memorization scores of CIFAR-100 training samples under SAM and SGD. The regression curve (in \textcolor{green}{green}) shows a consistent deviation from the identity line (in \textcolor{red}{red}), indicating that SAM memorizes a larger subset of samples in the lower score range which are likely to be atypical subclass samples. (c): Visualization of samples more memorized by SAM for the tiger class following the same setting of (b).  }
\end{figure}
We first focus on comparing the memorization behavior of SAM and SGD. \Cref{fig:memsgdsamdist} shows kernel density estimates of memorization scores for both models. Although the overall shapes of the distributions are similar -- reflecting the long-tailed nature of the dataset --, SAM exhibits a lower density at the lowest end of the spectrum, with the mass redistributed more evenly across the rest of the range. This indicates that SAM assigns higher memorization scores more broadly, suggesting a structured memorization of more diverse patterns compared to SGD.

To further investigate this behavior, we plot the memorization scores of individual CIFAR-100 samples under both SAM and SGD in \Cref{fig:memsgdsamscatter}. Each sample is represented as a blue dot, with its x- and y-coordinates corresponding to its memorization score under SGD and SAM, respectively. The red diagonal line denotes equal memorization across both optimizers. Samples above this line and to the left (top-left quadrant) are more memorized by SAM, while those below and to the right (bottom-right quadrant) are more memorized by SGD.

A regression analysis over all samples -- shown via the green curve -- reveals a consistent deviation from the identity -- red -- line, skewed towards the top-left quadrant. This indicates a systematic increase in memorization for a large subset of samples under SAM. Crucially, this deviation is not concentrated at the high end of the memorization spectrum. This finding -- together with the kernel density plot -- supports our hypothesis that SAM does not simply memorize pure noise samples, but rather focuses on non-dominant, atypical subclass samples that are underrepresented in the training distribution. Indeed, if SAM were picking up sample-specific noise, we would expect a sharp concentration of kernel density at the highest end of the spectrum and a deviation of the regression curve in the top-right quadrant.

\Cref{fig:memsgdsamtiger} illustrates this phenomenon within the \emph{tiger} class. Samples with higher SAM memorization relative to SGD (top-left region) tend to depict clean samples containing atypical sub-patterns---e.g., close-ups of tiger heads, tigers in water, or multiple tigers in a single image. These are visually distinct yet semantically consistent with the class label. In contrast, samples with high memorization under both SAM and SGD (top-right region) often contain sample-specific noise, -- e.g., a tiger with shiny paws on a pitch-black background --, which are less likely to generalize.

\subsection{SAM Increases Influence of High Memorization Samples} \label{ssec:infl_vs_mem}

\begin{figure}
  \centering
  \subfigure[\label{fig:inflmemsgdtop20}]{
  \begin{minipage}[c]{0.3\textwidth}
  \centering
  \includegraphics[width=\linewidth]{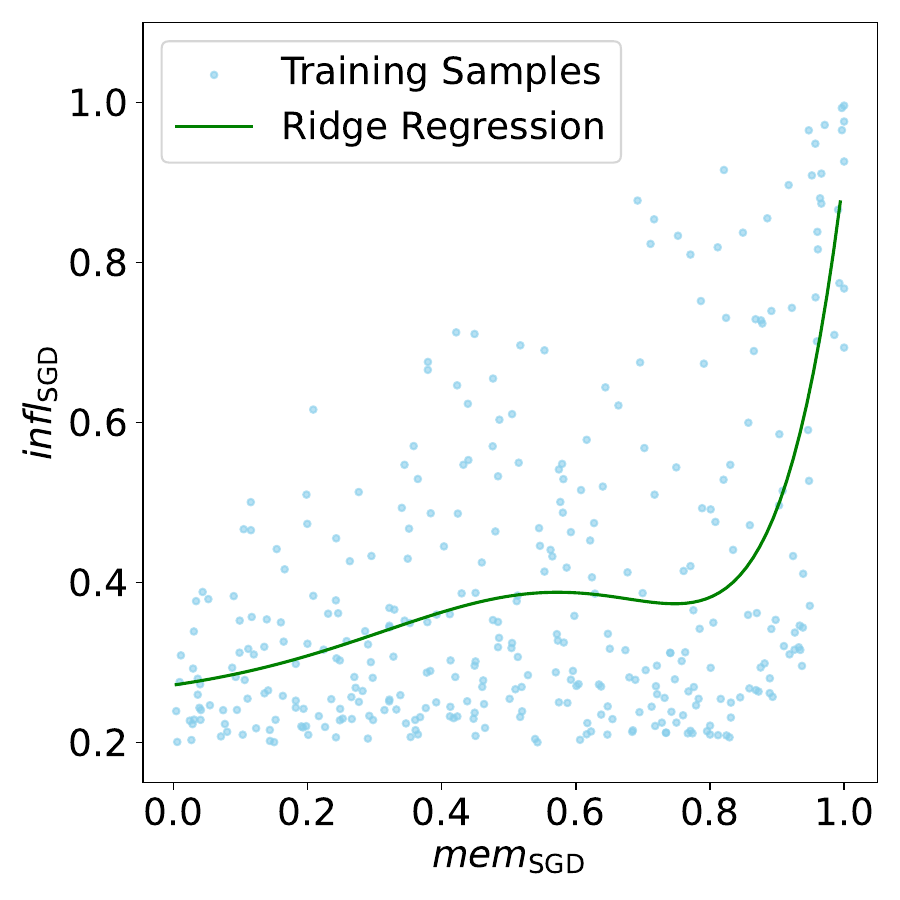}
  \end{minipage}
  }
  \hfill 
  \subfigure[\label{fig:inflmemsamtop20}]{
  \begin{minipage}[c]{0.3\textwidth}
  \centering    
  \includegraphics[width=\linewidth]{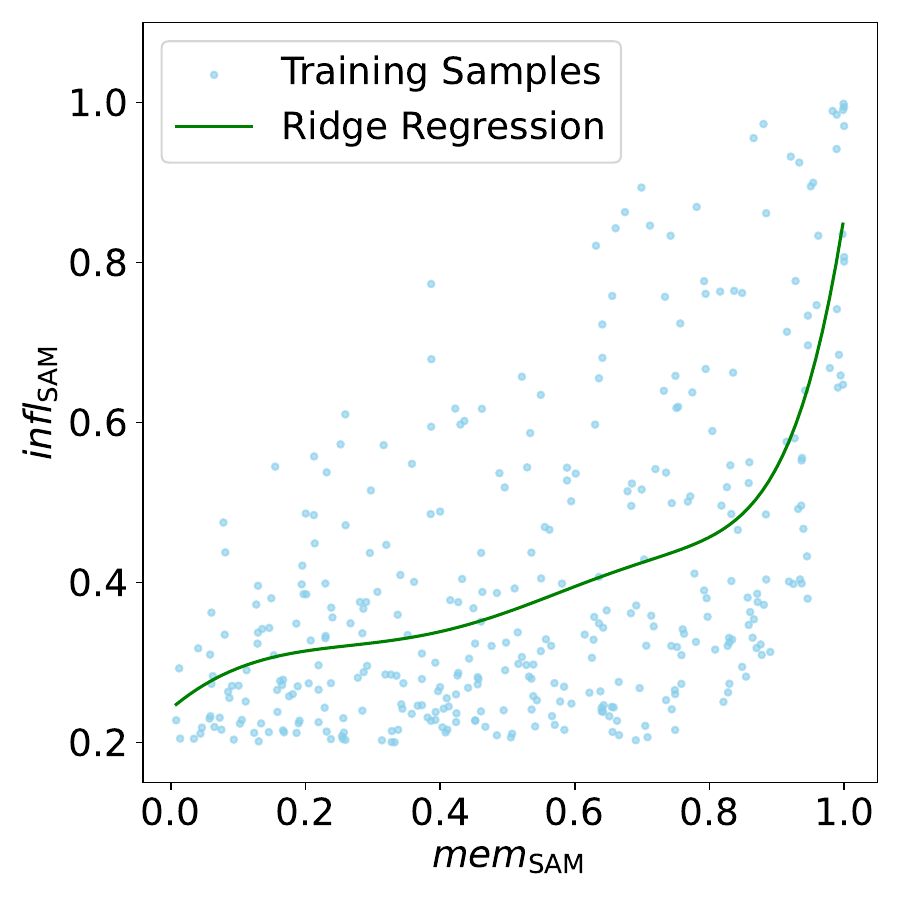}
  \end{minipage}
  }
  \hfill 
  \subfigure[\label{fig:inflmemdiff}]{
  \begin{minipage}[c]{0.3\textwidth}
    \centering
  \includegraphics[width=\linewidth]{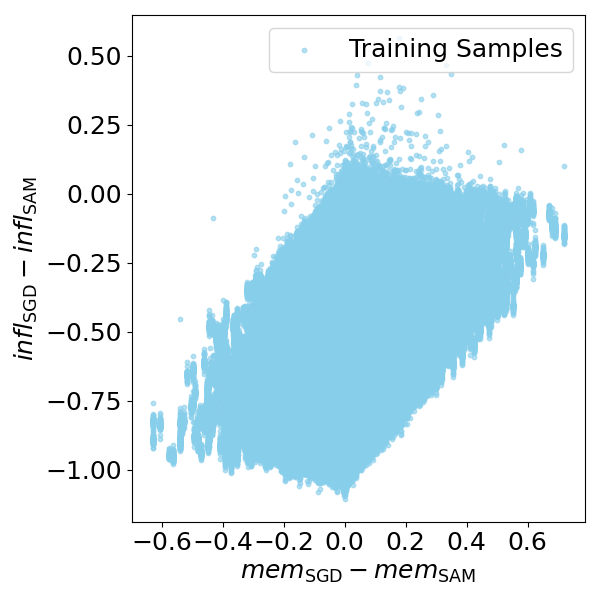}
  \end{minipage}}
  \caption{(a) and (b): Distribution of the influence scores of the 20 most influential training samples over each memorization interval for SGD (a) and SAM (b). The regression analysis (\textcolor{green}{green} lines) shows that SAM maintains a smoother influence distribution, relying more on mid-to-high (0.6 - 0.85) memorization samples (subclass features) than SGD. (c) Difference in influence scores between SAM and SGD as a function of memorization score differences. SAM downweighs influence for low-memorized samples and selectively amplifies the influence of mid-to-high memorization samples.}
\end{figure}

We here analyze how memorization affects the influence of training samples on test predictions, using the influence metric defined in \Cref{eq:infl}. Following the setup of \citet{feldman2020neural}, we first filter training–test sample pairs with influence scores above 0.2 to exclude non-influential cases. We then group training samples by memorization intervals -- defined as $l < \text{mem}(\mathcal{A}, \mathcal{D}, i) < u$, with $l$ and $u$ ranging from 0 to 1 in steps of 0.05 --- and, for each interval, we select the 20 training samples achieving the highest influence score on test data. This yields a distribution of influence scores of the most influential training samples conditioned on their memorization levels.

\Cref{fig:inflmemsgdtop20,fig:inflmemsamtop20} show the resulting distributions for SGD and SAM, respectively. As in \Cref{fig:memsgdsamscatter}, we fit a regression curve (green line) to highlight the trends. For SGD, influence scores incur in a steep transition from lowly influent samples to highly influent points at the upper end of the memorization spectrum. This indicates SGD's reliance on a very narrow set of highly memorized samples. In contrast, SAM exhibits a smoother transition curve, with a larger set of high (and mid-to-high) memorization samples contributing more consistently to test predictions. This supports our earlier finding that SAM emphasizes a set of atypical, non-dominant subclass patterns.

To further validate this, we examine the difference in influence scores between SAM and SGD as a function of their memorization score differences (\Cref{fig:inflmemdiff}). Training samples which are more memorized by SGD tend to have lower influence under SAM, suggesting that SAM down-weighs the influence of its low-memorized samples. Conversely, samples with similar memorization under both optimizers but higher influence under SAM tend to lie in the mid-to-high memorization range. These are precisely the samples containing atypical subpatterns that SAM selectively amplifies, confirming our intuitions.

\subsection{SAM's Generalization Gain Comes From Higher Memorization of Sub-patterns}
\label{sec:doesSAMmem}
\begin{figure}
\centering

\subfigure[Bicycle]{
\begin{minipage}[c]{1.0\textwidth}
\centering
\fbox{\includegraphics[height=1cm,width=1cm]{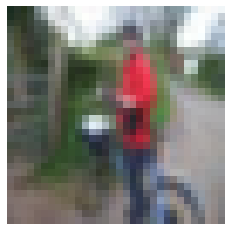}}
\includegraphics[height=1cm,width=1cm]{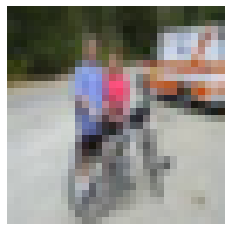}
\includegraphics[height=1cm,width=1cm]{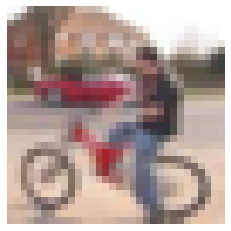}
\includegraphics[height=1cm,width=1cm]{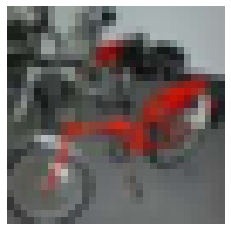}
\includegraphics[height=1cm,width=1cm]{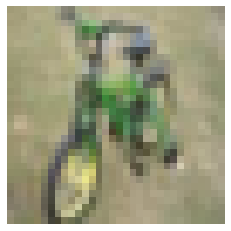}
\includegraphics[height=1cm,width=1cm]{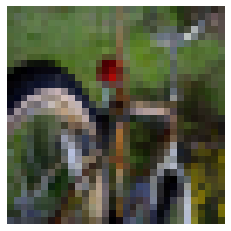}
\includegraphics[height=1cm,width=1cm]{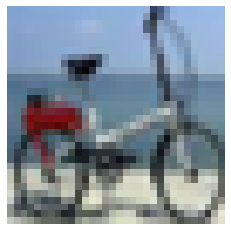}
\includegraphics[height=1cm,width=1cm]{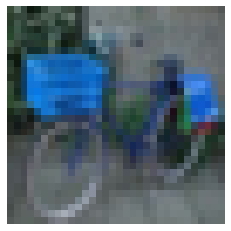}
\includegraphics[height=1cm,width=1cm]{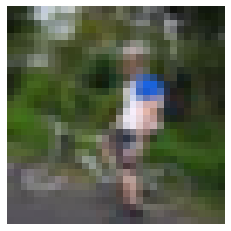}
\includegraphics[height=1cm,width=1cm]{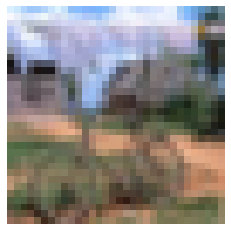}
\includegraphics[height=1cm,width=1cm]{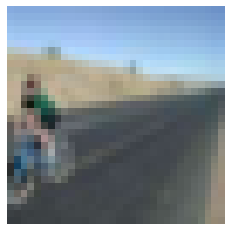}
\hfill \\
\fbox{\includegraphics[height=1cm,width=1cm]{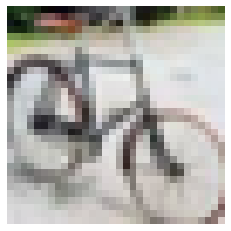}}
\includegraphics[height=1cm,width=1cm]{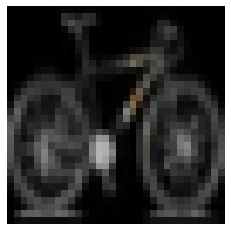}
\includegraphics[height=1cm,width=1cm]{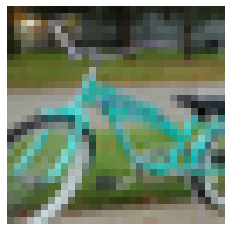}
\includegraphics[height=1cm,width=1cm]{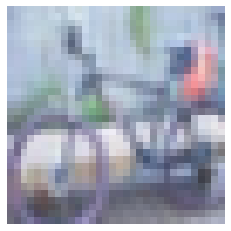}
\includegraphics[height=1cm,width=1cm]{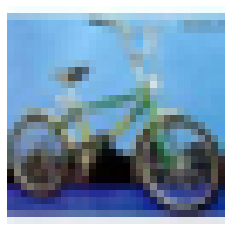}
\includegraphics[height=1cm,width=1cm]{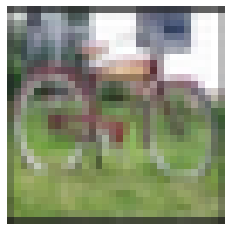}
\includegraphics[height=1cm,width=1cm]{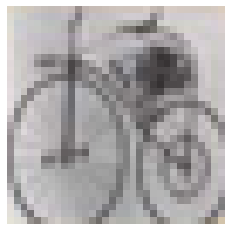}
\includegraphics[height=1cm,width=1cm]{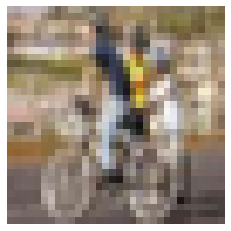}
\includegraphics[height=1cm,width=1cm]{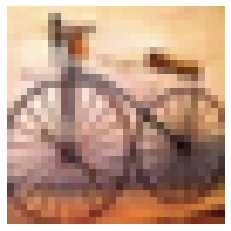}
\includegraphics[height=1cm,width=1cm]{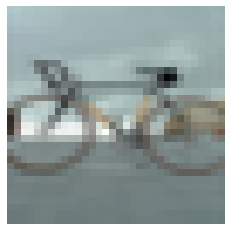}
\includegraphics[height=1cm,width=1cm]{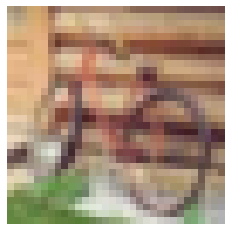}
\end{minipage}
}
\hfill

\hfill
\subfigure[Tiger]{
\begin{minipage}[c]{1.0\textwidth}
\centering
\fbox{\includegraphics[height=1cm,width=1cm]{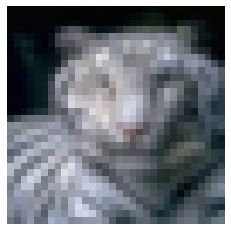}}
\includegraphics[height=1cm,width=1cm]{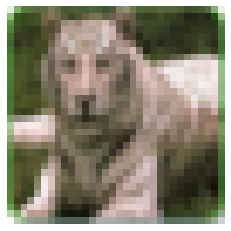}
\includegraphics[height=1cm,width=1cm]{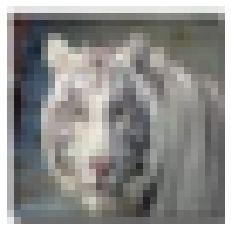}
\includegraphics[height=1cm,width=1cm]{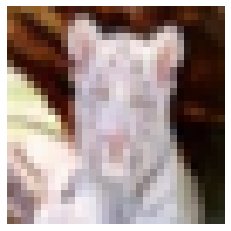}
\includegraphics[height=1cm,width=1cm]{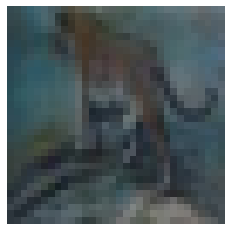}
\includegraphics[height=1cm,width=1cm]{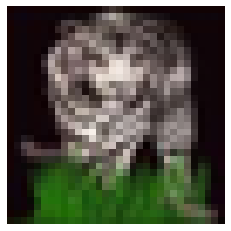}
\includegraphics[height=1cm,width=1cm]{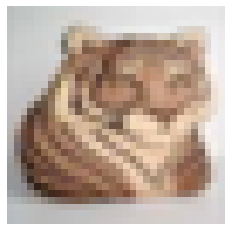}
\includegraphics[height=1cm,width=1cm]{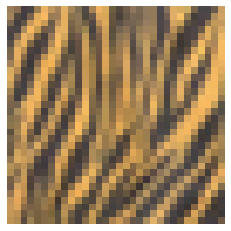}
\includegraphics[height=1cm,width=1cm]{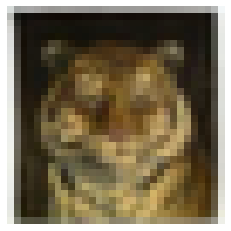}
\includegraphics[height=1cm,width=1cm]{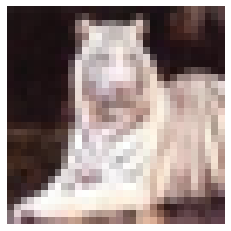}
\includegraphics[height=1cm,width=1cm]{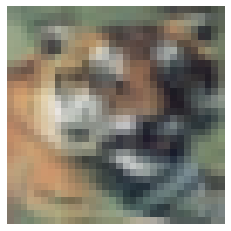}
\hfill \\
\fbox{\includegraphics[height=1cm,width=1cm]{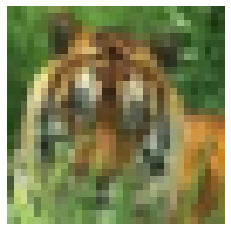}}
\includegraphics[height=1cm,width=1cm]{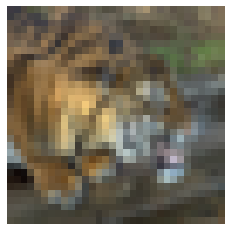}
\includegraphics[height=1cm,width=1cm]{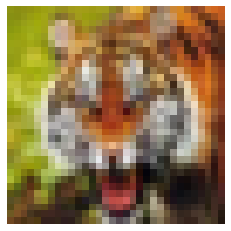}
\includegraphics[height=1cm,width=1cm]{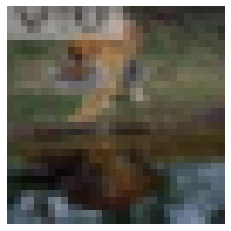}
\includegraphics[height=1cm,width=1cm]{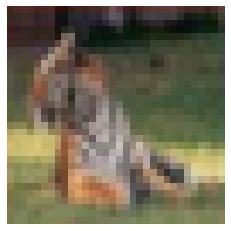}
\includegraphics[height=1cm,width=1cm]{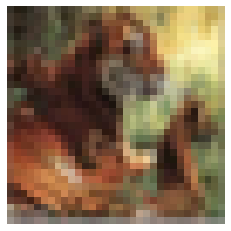}
\includegraphics[height=1cm,width=1cm]{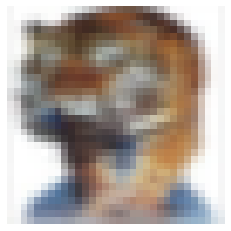}
\includegraphics[height=1cm,width=1cm]{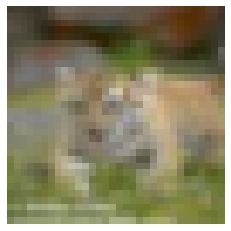}
\includegraphics[height=1cm,width=1cm]{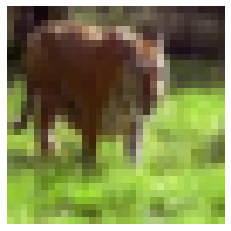}
\includegraphics[height=1cm,width=1cm]{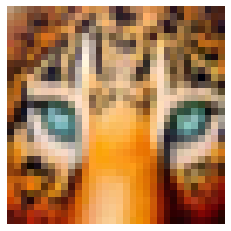}
\includegraphics[height=1cm,width=1cm]{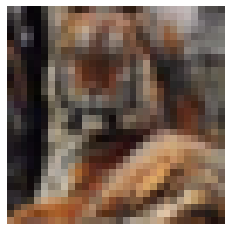}
\end{minipage}
}
\caption{Test images (boxed) from buckets 1 and 5 and their respective top-10 influential training images. For each object the top row is an image from bucket 1 and the bottom row is an image from bucket 5. For bucket 1 images (higher memorization,top row), notice that the images are atypical for their classes, and there is a near duplicate in the training data that was important for generalizing on this test image. For bucket 5 images, on the other hand, the top influential images are reminiscent of the test image at a conceptual level.}
\label{fig:atypicalexamples}
\end{figure}

In this section, we dissect the generalization gains of SAM at a finer granularity by constructing a metric that divides the test data points into groups based on the amount of memorization used for predicting them. We then compare the performance on each group between SGD and SAM. 

We measure the typicality of a test data point as the entropy of its corresponding training samples' influence scores. We rely on this measure since, in practice, the prediction of a typical unseen sample would be evenly influenced by numerous training data points within the same class, while atypical counterparts would be heavily influenced by a handful of training samples that are themselves also atypical. To measure the even spread, we normalize the influence scores and leverage entropy. Even influence spread would follow a more uniform distribution resulting in high entropy, while uneven spread incurs low entropy. Formally, for each test data point $i$, let $\mathcal{S}_i$ be the set of influence scores of all the training points in the same class. Let $S_{i,j}$ be the influence score of $j_{th}$ training point and $m$ be the number of training points in the same class. Then, our entropy metric $\mathcal{I}_{ent}$ is defined as:
\begin{align}
\mathcal{I}_{ent}[i] = \sum_{j=1}^m - p_{i,j} \cdot \log{p_{i,j}}, \label{entr}
\text{   where, }  p_{i,j} = \frac{ \mathcal{S}_{i,j} }{\sum_j  \mathcal{S}_{i,j} }
\end{align}

\begin{figure}
  \centering
  \subfigure[\label{fig:bucketvsaccuracy}]{
  \begin{minipage}[c]{0.3\textwidth}
  \centering
  \includegraphics[width=\linewidth]{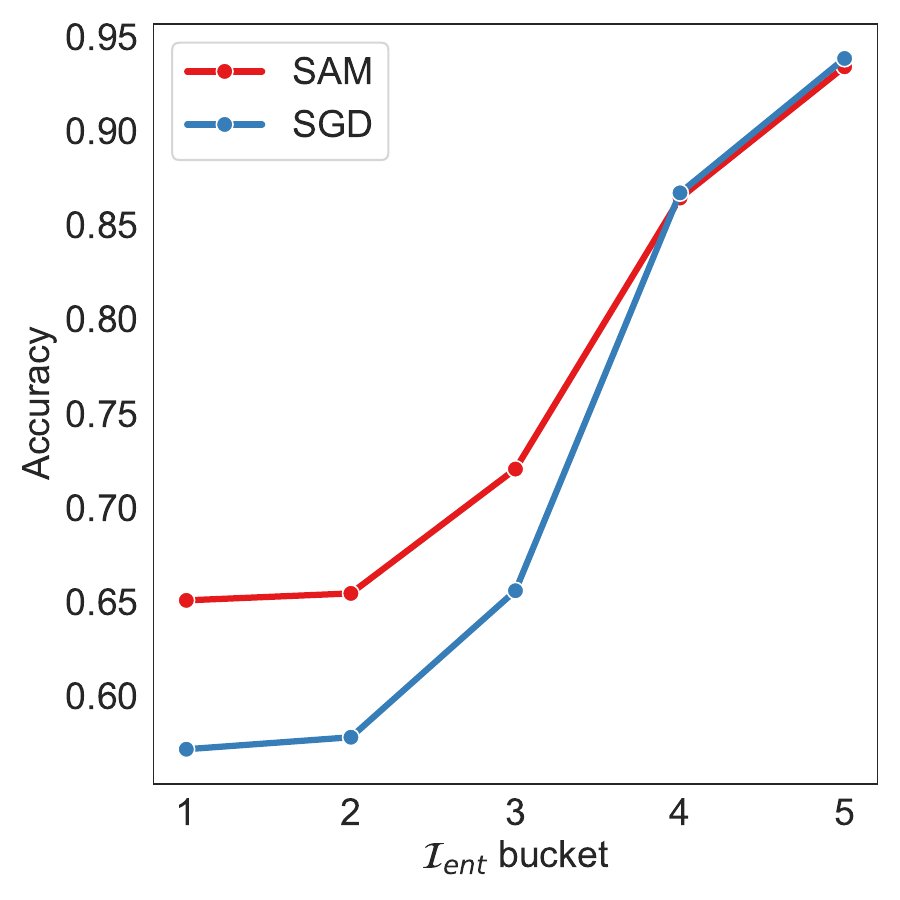}
  \end{minipage}
  }
  \hfill 
  \subfigure[\label{fig:memorizationscorebucket1}]{
  \begin{minipage}[c]{0.3\textwidth}
  \centering    
  \includegraphics[width=\linewidth]{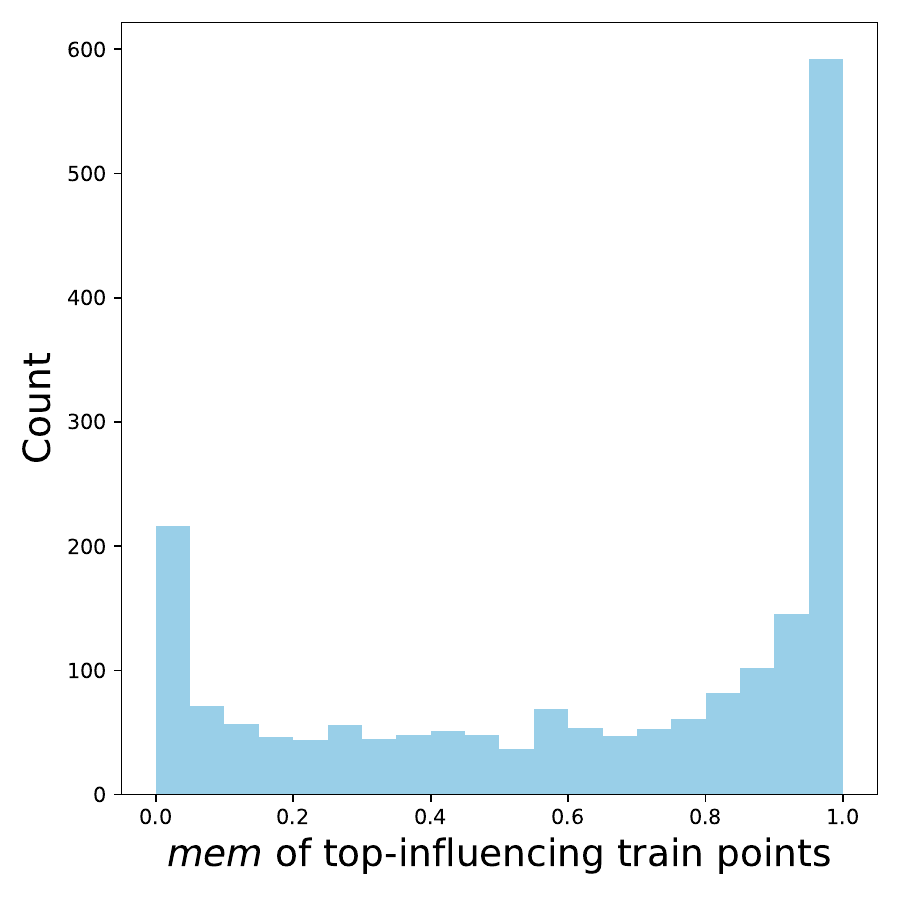}
  \end{minipage}
  }
  \hfill 
  \subfigure[\label{fig:memorizationscorebucket2}]{
  \begin{minipage}[c]{0.3\textwidth}
    \centering
  \includegraphics[width=\linewidth]{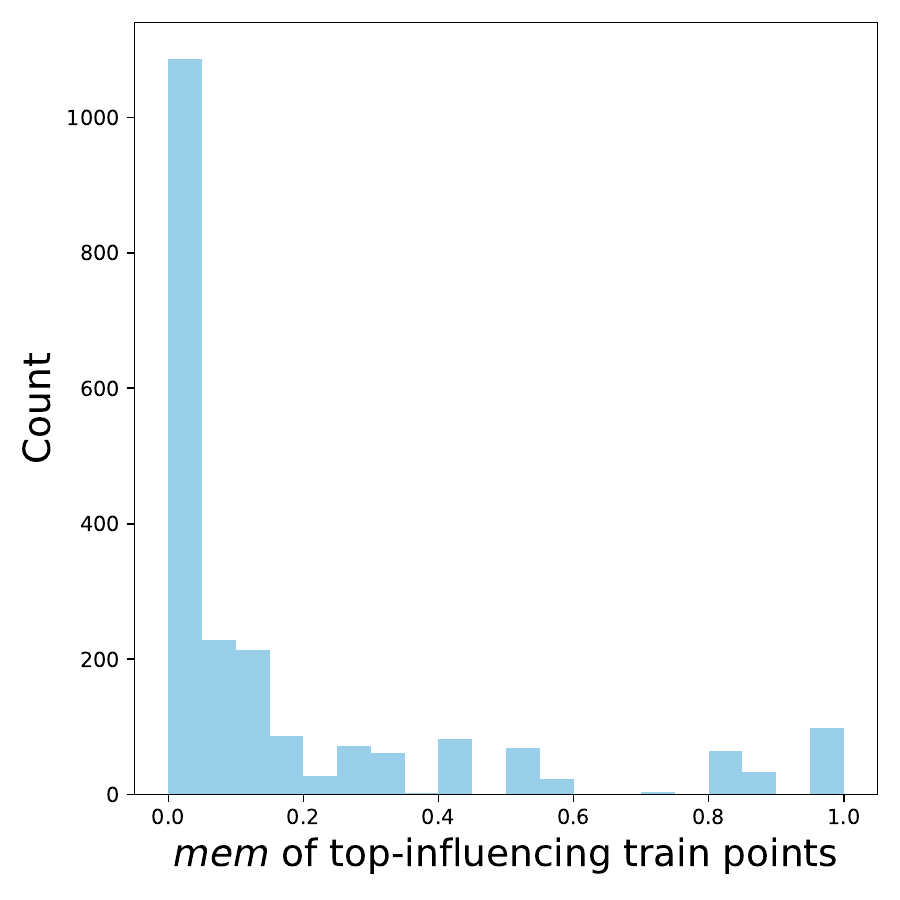}
  \end{minipage}}
  \caption{(a): Test accuracy on $\mathcal{I}_{ent}$ groups as evaluated by~\eqref{entr}. SAM's performance gains comes from it correctly predicting more atypical data points that need memorization of atypical sub-patterns to be classified correctly. (b) and (c): Distribution of top-1 most influential training point's memorization scores for $\mathcal{I}_{ent}$ buckets 1 and 5. Testing samples falling in the lower (higher) numbered buckets are influenced by training points with higher (lower) memorization.}
\end{figure}

We group test data points into 5 buckets in the order of lowest $\mathcal{I}_{ent}$ to highest $\mathcal{I}_{ent}$. We present some test images and their top-10 influential training images in \Cref{fig:atypicalexamples} from bucket 1 and bucket 5. The figure illustrates that images from bucket 1 tend to be atypical images -- e.g., bicycle alongside people,  white tiger, etc, -- for their respective labels while images from bucket 5 tend to be more typical images---e.g., typical bicycle and yellow tiger. For quantitative verification, we plot the distribution of memorization scores of the highest influencing training points from each bucket. We observe that lower numbered buckets are associated with high memorization and vice versa (see \Cref{fig:memorizationscorebucket1,fig:memorizationscorebucket2}). The results for other buckets interpolate between those of bucket 1 and 5, and are skipped for brevity. 

We compare the generalization gains of SAM against SGD on each of these buckets and show the results in \Cref{fig:bucketvsaccuracy}\footnote{These results do not consider image transformations (e.g. random crop, rotations), however we have also replicated the experiment with transformations, obtaining a similar trend.}. For test data points in bucket 5, SGD achieves a negligible performance gain, while for bucket 1 SAM achieves a significant gain w.r.t. SGD. Thus, the performance gains of SAM can be attributed to it correctly predicting more atypical data points which need more memorization of atypical sub-patterns to be classified correctly. For further validation, we generate a synthetic dataset and illustrate SAM's capability of learning atypical subclasses better than SGD in \Cref{sec:synth}. We theoretically analyze how stronger alignment with minority subclass features can lead to better generalization in Section~\ref{app:msa_theory}.

\paragraph{Summary of experimental findings} These results -- together with those in \Cref{ssec:mem_analysis,ssec:infl_vs_mem} -- provide strong empirical evidence that \emph{SAM’s increased performance derives from better capturing atypical but informative sub-patterns, allowing predictions to be less dominated by the majority feature}. For SGD, we posit that this results in a higher output sensitivity, yielding more confident predictions on unseen data. Paradoxically, this high variance acts as a cloak for membership privacy, as it mimics the high confidence usually reserved for members. In the next section, we theoretically analyze how sharpness-regularization can inherently suppress prediction variance on unseen data.

\section{Theoretical Analysis}\label{sec:thms}
In this section, we provide a theoretical foundation for the variance--shrinkage
effect of sharpness-aware minimization and its implication for membership
inference risk. To obtain a clean and analyzable characterization of SAM's geometric bias, we study perfectly interpolating regime of overparameterized linear models. Closest to our analysis in spirit is that of~\citet{privacytheory}, which explains how parameter size and ridge regression affects membership risk via a variance gap. In contrast, we introduce a curvature-aligned geometry modeling SAM and show that it provably increases membership risk.

Our proofs are written in the regression form $X\theta=y$. We note that the
connection to classification is direct. For separable binary classification
with losses such as logistic or exponential, gradient descent is known to
implicitly maximize the $\ell_2$ margin, and the resulting predictor is
geometrically equivalent to the minimum-$\ell_2$-norm interpolator of a
corresponding regression problem
\citep{soudry2024implicitbiasgradientdescent,gunasekar2017implicitregularizationmatrixfactorization,muthukumar2021classificationvsregressionoverparameterized}.
Motivated by this equivalence, we can consider classification as regression to
high-magnitude targets $y_i\in\{-M,+M\}$ for a large constant $M\gg 1$ under MSE
loss. In this regime, training points are interpolated to $\pm M$
and therefore have fixed high confidence, whereas test points
 yield outputs that fluctuate around the decision boundary
($0$). 

\paragraph{Model and notation.}
We work in finite dimension \(d\gg n\).
Let the population feature covariance be \(\Sigma\in\R^{d\times d}\) with \(\Sigma\succ0\).
Draw i.i.d.\ samples
\[
x_i\sim\mathcal{N}(0,\Sigma),\qquad y_i=\theta^{*\top}x_i+\xi_i,\quad
\xi_i\sim\mathcal{N}(0,\sigma_y^2),\ \text{independent of }x_i,\quad i=1,\dots,n.
\]
Let \(X\in\R^{n\times d}\) have rows \(x_i^\top\), \(y=(y_1,\dots,y_n)^\top\),
and assume \(\mathrm{rank}(X)=n\).
Define the orthogonal projector onto the data span and the covariance matrix 
\[
P := X^\top(XX^\top)^{-1}X,\qquad P^2=P=P^\top,\qquad
\widehat{\Sigma}:=\frac1n X^\top X,\qquad
\Sigma:=\mathbb{E}[xx^\top].
\]
The model is defined as $f_G(x) := \widehat\theta_G^\top x$. In the squared-loss linear model, the population Hessian $H$ equals $\Sigma$
and the empirical Hessian $\widehat{H}$ equals $\widehat{\Sigma}$.

Now we introduce the geometries we employ to model SGD and SAM. 

\paragraph{Min-\(G\) interpolation.}
For any symmetric positive definite matrix \(G\succ0\), consider the minimum-\(G\)-norm interpolant:
\begin{equation}
\label{eq:minG}
\widehat\theta_G\ :=\ \arg\min_{\theta\in\mathbb{R}^d}\ \frac12\,\theta^\top G\,\theta
\quad\text{s.t.}\quad X\theta=y .
\end{equation}
We compare the standard Euclidean case, 
\[
G_0:=I_d
\]
with the \emph{sharpness-aware} geometry,
\[
G_\eta := I + \eta H = I + \eta \Sigma,\qquad \eta>0,
\]

Classical implicit-bias results identify standard SGD with the
Euclidean case under step size and weight initialization assumptions~\citep{zhang2017musings}. Sharpness-aware geometry, on the other hand, reflects SAM's penalty on local
sharpness, which can be shown by expanding the minimization objective. We prove that, relative to the Euclidean interpolation, the geometry
$G_\eta$ strictly reduces the variance of non-member outputs, thereby enlarging
the separation between members and non-members and increasing the advantage of
membership inference attacks.

We first establish that SAM strictly reduces the variance of the model's output on unseen data. By Lemma~\ref{lem:test_dist}, for a non-member sample $X_{\mathrm{out}}$, and the output follows $f_G(X_{\mathrm{out}})\ \sim\ \mathcal{N}\!\big(0,\ \sigma_G^2\big)$.

\begin{theorem}[Variance strictly decreases for SAM geometry]
\label{thm:variance-strict}
Under Assumption~\ref{asm:overlap}, there exists $\eta_0>0$ such that for all
$\eta\in(0,\eta_0]$,
\[
\sigma_{G_\eta}^2 < \sigma_{G_0}^2
\qquad\text{with probability } 1-o(1).
\]
\end{theorem}

\begin{remark}
Theorem~\ref{thm:variance-strict} formalizes a simple geometric picture.
The minimum-$G$ interpolant balances fitting the training span  against penalizing components in directions where $G$ is large.
Moving from $G_0$ to $G_\eta$ increases the penalty
precisely along high-curvature directions of the Hessian.
Lemma~\ref{lem:theta-derivative} shows that, under the overlap condition in
Assumption~\ref{asm:overlap}, this reweighting strictly suppresses the
$(I-P)\Sigma\widehat\theta_{G_0}$ component of the interpolant. Since
non-member predictions depend on $\widehat\theta_G$ only through the quadratic
form $\widehat\theta_G^\top\Sigma\widehat\theta_G$, this suppression translates
directly into a strict decrease of the output variance on unseen data.    
\end{remark}

Next, we consider confidence-threshold attack and likelihood ratio attack.
\paragraph{Confidence-threshold attack.}
The attacker uses the confidence score
$\mathrm{Conf}_G(x):=|f_G(x)|$ and predicts ``member'' iff $\mathrm{Conf}_G(x)\ge\tau$.
Let $(X_{\mathrm{in}},Y_{\mathrm{in}})$ be a random training pair and
$X_{\mathrm{out}}\sim\mathcal N(0,\Sigma)$ an independent non-member. Define attack advantage
\[
\mathrm{Adv}^{\mathrm{conf}}_G:=\sup_{\tau\ge0}\big(\mathrm{TPR}_G(\tau)-\mathrm{FPR}_G(\tau)\big).
\]

\begin{theorem}[SAM strictly increases confidence-based MIA advantage]
\label{thm:adv-conf}
Under Assumption~\ref{asm:overlap}, for all sufficiently small $\eta>0$,
\[
\mathrm{Adv}^{\mathrm{conf}}_{G_\eta}
\;>\;
\mathrm{Adv}^{\mathrm{conf}}_{G_0}
\qquad
\text{with probability }1-o(1).
\]
\end{theorem}

\begin{remark}
For the confidence-threshold attack, interpolation implies that member
confidences are geometry-invariant: the training logits are fixed  (up to label noise) for any choice of $G$. Thus, changing the
geometry from $G_0$ to $G_\eta$ leaves $\mathrm{TPR}_G(\tau)$ unchanged for
every threshold $\tau$, while Theorem~\ref{thm:variance-strict} strictly
reduces the non-member variance and hence lowers $\mathrm{FPR}_G(\tau)$ at
every $\tau>0$.
In other words, SAM pushes non-member scores closer to the decision boundary, sharpening the
separation between the two confidence distributions.
\end{remark}

\paragraph{Likelihood Ratio (LR) attack.}
The oracle LRA predicts ``member'' iff $\Lambda_G(f_G(x))\ge t$.
Define
\[
\Lambda_G(s):=\log\frac{p_{\mathrm{in}}(s)}{p_{\mathrm{out}}(s;G)}, \qquad
\mathrm{Adv}^{\mathrm{LR}}_G:=\sup_{t\in\mathbb R}
\big(\Pr(\Lambda_G(f_G(X_{\mathrm{in}}))\ge t)-\Pr(\Lambda_G(f_G(X_{\mathrm{out}}))\ge t)\big) .
\]

\begin{theorem}[SAM strictly increases LR-attack advantage]
\label{thm:adv-lr}
Under Assumption~\ref{asm:overlap}, for all sufficiently small $\eta>0$,
\[
\mathrm{Adv}^{\mathrm{LR}}_{G_\eta}
\;>\;
\mathrm{Adv}^{\mathrm{LR}}_{G_0}
\qquad\text{with probability }1-o(1),
\]
where $G_\eta=I+\eta\Sigma$.
\end{theorem}
\begin{remark}
The argument is similar to Theorem~\ref{thm:adv-conf}. Theorem~\ref{thm:adv-lr} analyzes an oracle LR attacker that uses a single
global IN/OUT distribution.
Corollary~\ref{cor:lira} strengthens this to sample-adaptive IN/OUT
distributions, matching the per-query calibration used by LiRA/RMIA. We note that while Theorem~\ref{thm:adv-conf} and Theorem~\ref{thm:adv-lr} state an improvement in the
supremum advantage, the proof in fact shows a
stronger statement: for all sufficiently small $\eta>0$, \emph{the entire ROC curve
of the confidence-based attack under $G_\eta$ strictly dominates that under
$G_0$}. This is empirically verified in Figure~\ref{fig:roc}, where SAM's ROC curve is above that of SGD's for nearly the \emph{entire} range across most settings. 
\end{remark}

\section{Conclusion and Future Work}
This work investigates the mechanism behind SAM’s dual effect: superior generalization coupled with heightened privacy leakage. As algorithms seeking flatter minima are widely employed, our work serves as a cautionary tale users should be aware of. Furthermore, extending our findings to devise an effective defense against Membership Inference Attacks would be beneficial to the community. Finally, future research is encouraged to further scrutinize the implicit bias of SAM. 

\bibliography{main} 
\bibliographystyle{iclr2026_conference} 
\clearpage

\appendix

\section{Synthetic Dataset} \label{sec:synth}
\begin{figure}

\subfigure[Class contours\label{fig:toya}]{
\begin{minipage}[c]{0.3\textwidth}
\centering
\includegraphics[height=3.8cm, width=4.5cm]{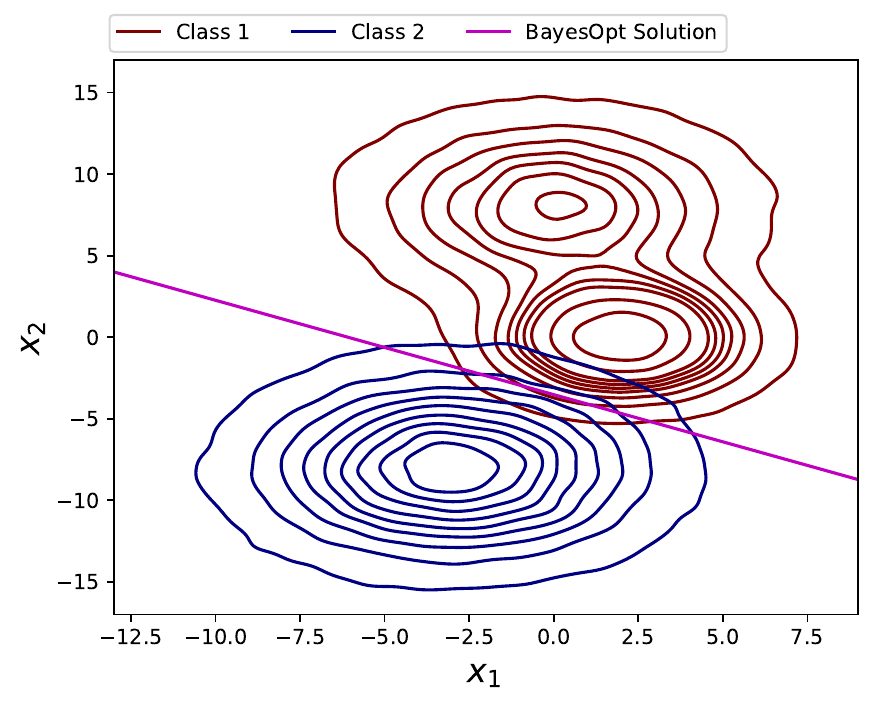}
\end{minipage}
}
\hfill
\subfigure[SAM vs SGD (Few atypical examples) \label{fig:toyb}]{
\begin{minipage}[c]{0.3\textwidth}
\centering
\includegraphics[scale=0.3]{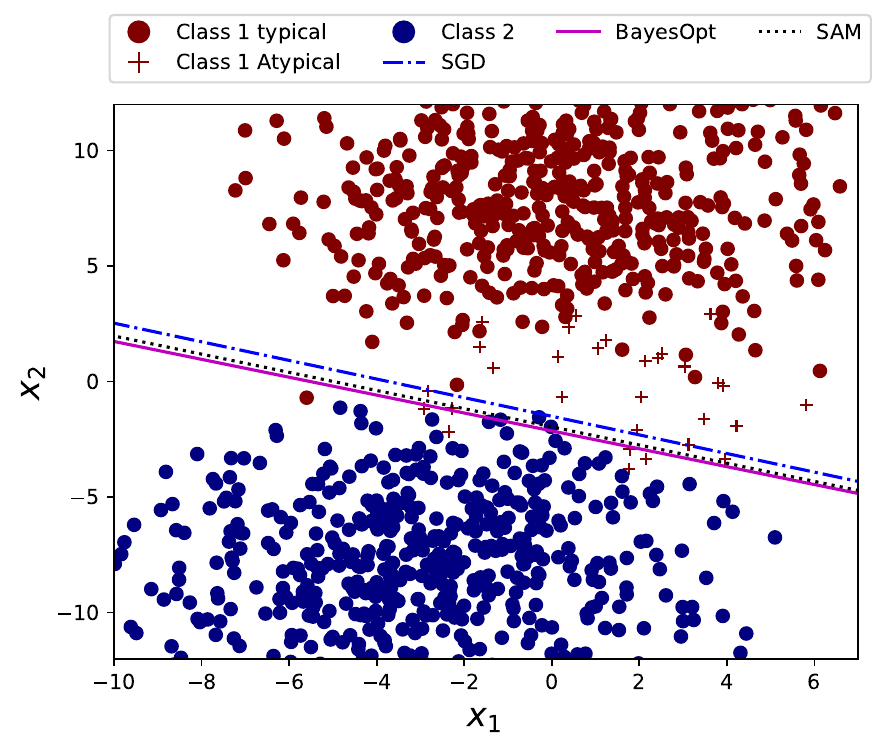}
\end{minipage}
}
\hfill
\subfigure[SAM vs SGD (Large sample size) \label{fig:toyc}]{
\begin{minipage}[c]{0.3\textwidth}
\centering
\includegraphics[scale=0.3]{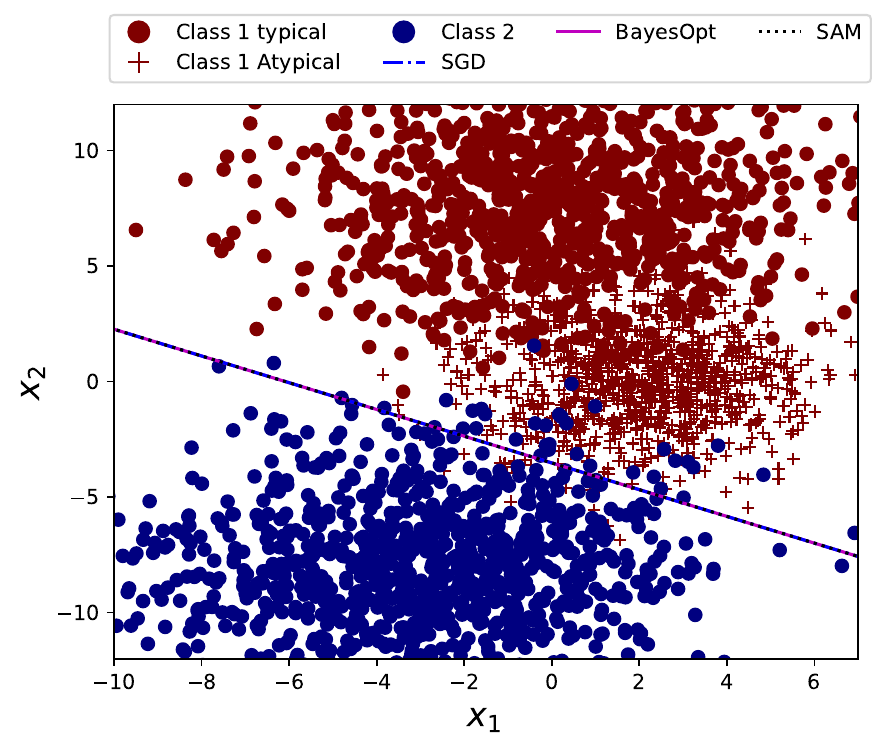}
\end{minipage}
}
\caption{A synthetic construction illustrating the generalization ability of SAM over SGD for atypical examples. Fig (a) shows class density contours of a two-class, 2-dimensional classification problem, along with the Bayes Optimal solution. The red class has two `clusters', one representing typical examples and one representing atypical examples. Fig (b) shows an instance of data sampled from densities shown in (a); the larger cluster of red dots represent typical examples in the red class, and the red `+' points represent a lot fewer atypical examples. SAM generalizes better than SGD in this case. Fig (c) shows that if there are enough samples generated from both typical and atypical clusters, SAM and SGD coincide with the Bayes Optimal classifier.}
\label{fig:toy}
\end{figure}

In this section, we provide a simple synthetic construction that illustrates how SAM can achieve better generalization performance vs vanilla SGD. The example is illustrated in \Cref{fig:toy}. The data is generated from two-dimensional densities illustrated in \Cref{fig:toya}. The densities are supported in two dimensions labelled as $x_1$ and $x_2$. There are two classes - the red class and the blue class. \Cref{fig:toya} also shows the Bayes Optimal classifier. The red class has two `clusters', one representing the typical examples (e.g. yellow tigers), and the other representing the atypical examples (e.g. white tigers). The data is sampled in such a way that we have several samples from the typical cluster, while there are only a few samples from the atypical cluster in the red class. This is shown in \Cref{fig:toyb}. \Cref{fig:toyb} further shows that seeking flatter minima using the SAM optimizer learns a classifier that is closer to the Bayes Optimal classifier than the classifier learnt using vanilla SGD, and thus the former generalizes better. This difference in performance vanishes in \Cref{fig:toyc} when we have a large sample size for the atypical examples as well. 

This synthetic construction shows that one possible reason that SAM can perform better is if it tends to memorize atypical samples more than vanilla SGD. In other words, the gain in generalization could potentially come from those atypical data subgroups. In the next subsection, we empirically verify this conjecture for the CIFAR-100 dataset.

\section{Better minority subclass alignment leads to higher generalization}
\label{app:msa_theory}
In this section, we theoretically analyze a setting involving majority subclass samples, minority subclass samples, and pure noise samples to illustrate how an overfitting model generalizes better. Driven by the motivations in the previous sections, we discern a model by the amount of minority subclass feature it captures. \Cref{thm:gen} shows that this property leads to higher generalization. The proofs are in Section~\ref{sec:gen_proof}.

\begin{definition}[Data Model] \label{def:datamodel}
Training dataset $D=\{(\mathbf{x}_i,y_i)\}_{i=1}^n$ is sampled i.i.d.\ from the data distribution $\mathcal{D}$. We define $\mathbf{x} = [\mathbf{x}_1,\mathbf{x}_2,\mathbf{x}_3]$, $y \in \{\pm 1\}$, $\mathbf{x}_1 \in \mathbb{R}^{d_1}$, $\mathbf{x}_2 \in \mathbb{R}^{d_2}$, $\mathbf{x}_3 \in \mathbb{R}^{d_3}$. We consider a overparameterized regime where $d_1, d_2, d_3 \gg n$. We consider a linear predictor for classification, 
$f_\mathbf{w}(\mathbf{x}) = \operatorname{sign}\{\langle \mathbf{w}, \mathbf{x}\rangle\}$ with $\mathbf{w}=[\mathbf{w}_1,\mathbf{w}_2,\mathbf{w}_3]$. We assume perfect interpolation satisfying finite margin $\forall i$ $0 \leq m_0 \leq m_i \leq M < \infty$, where margin $m_i = y_i\langle\mathbf{w}, \mathbf{x_i}\rangle$.
Let $\mathcal{M}, \mathcal{S}, \mathcal{N}$ denote the majority, minority, 
and noise subsets, respectively. $D = \mathcal{M}\cup\mathcal{S}\cup\mathcal{N}$. We define $\sigma(z) = 1/(1+\exp^{-z})$.

$\mathcal{D}$ is a mixture distribution
\[
\mathcal{D} = p_{\mathcal{M}} \,\mathcal{D}_{\mathcal{M}} 
+ p_{\mathcal{S}} \,\mathcal{D}_{\mathcal{S}} 
+ p_{\mathcal{N}} \,\mathcal{D}_{\mathcal{N}},
\]
with mixture weights $p_{\mathcal{M}}+p_{\mathcal{S}}+p_{\mathcal{N}}=1,$ and $p_{\mathcal{M}} \gg p_{\mathcal{S}},p_{\mathcal{N}}$. 
Here, $\mathcal{D}_{\mathcal{M}}$ generates majority samples with 
$\mathbf{x}_1=y\bm{\mu}_1$, $\mathbf{x}_2 \sim \mathcal{N}(0,I_{d_2})$, $\mathbf{x}_3 \sim \mathcal{N}(0,I_{d_3})$ where $\bm{\mu}_1$ is a fixed vector; 
$\mathcal{D}_{\mathcal{S}}$ generates minority samples with 
$\mathbf{x}_1=\bm{\nu}$, $\mathbf{x}_2=y\bm{\mu}_2$, $\mathbf{x}_3 \sim \mathcal{N}(0,I_{d_3})$ where $\bm{\mu}_2$ is a fixed vector and $\bm{\nu}$ is a random vector anti-aligned with $\mathbf{w}_1$ ($\langle \mathbf{w}_1, \bm{\nu} \rangle < 0$); 
and $\mathcal{D}_{\mathcal{N}}$ generates pure noise samples with 
$\mathbf{x} \sim \mathcal{N}(0,I_{d_1 + d_2 + d_3})$. Let $(n_{\mathcal{M}},n_{\mathcal{S}},n_{\mathcal{N}})$ denote the counts of 
majority, minority, and noise samples in $S$. 
\end{definition}

Our model easily translates to a logistic regression model if we set $f_\mathbf{w}(\mathbf{x}) = \sigma(\langle \mathbf{w}, \mathbf{x}\rangle)$ and change the label to $y \in \{0,1\}$. The setup in which a part of the input contains the true signal has been commonly used in previous works \citep{chen2023doessharpnessawareminimizationgeneralize, kou2023benignoverfittingtwolayerrelu} but we generalize their setup by including and analyzing subclasses in the design. Concretely, one can think of a minority sample as belonging to a long-tail subgroup of the class that requires a different feature to be recognized. We formalize an anti-alignment condition: the minority features are arranged such that any model that heavily prioritizes the majority feature $u$ will perform poorly on the minority. Intuitively, fitting the minority subgroup requires the model to memorize an alternative pattern that is independent of (or even interfering with) the main decision boundary for the majority. We assume the minority subgroup is very small, so that by default a standard empirical risk minimizer might deem it negligible. This resonates with the long-tail phenomena observed in practice – a handful of unusual examples exist that a model could easily ignore without significant impact on overall training loss. However, those examples are crucial for tail generalization: they represent rare yet valid concepts that an ideal model should learn. Our assumptions reflect prior findings that real datasets contain such long-tailed subpopulations which must be memorized to achieve the best possible accuracy~\cite{feldman2020does}. Formally,

\begin{condition}
For each minority point $i\in\mathcal S$, define
\[
B_i \;\coloneqq\; -\,y_i\langle \mathbf{w}_1,\bm{\nu}_i\rangle \;>\; 0,
\qquad
A \;\coloneqq\; \langle \mathbf{w}_2,\bm{\mu}_2\rangle .
\]
Let $B$ be a random variable with CDF $F_B(A) = \Pr(B < A)$ such that $B \stackrel{d}{=} B_i$ (i.e., $F_B$ is the law/distribution of the $B_i$’s when $i$ is drawn uniformly from $\mathcal S$).
We assume
\[
A \;<\; B_{\max},
\qquad
B_{\max} \coloneqq \sup\{\, b \in \mathbb{R} \mid F_B(b) < 1 \,\},
\]
\end{condition}

This condition assures that the majority feature still dominates globally. Furthermore, we formulate $\bm{\nu}$ as a random variable to effectively capture multiple atypical subclasses. For example, there can be a wide range of tigers that are purely white or yellowish-white. By modeling $\bm{\nu}$ as a random variable, it provides variation on the strength of anti-alignment with the majority feature. 

Driven by the empirical motivations in \Cref{sec:SAMatyp}, we define an ordering of the models as how much minority subclass alignment (MSA) they achieved. Formally,

\begin{definition}[Minority Subclass Alignment Order]
Given two interpolating solutions $\mathbf{w}^{(A)},\mathbf{w}^{(B)}$ trained on the same $S$, define
\[
A^{(A)} \;\coloneqq\; \langle \mathbf{w}^{(A)}_2,y\bm{\mu}_2\rangle,\qquad
A^{(B)} \;\coloneqq\; \langle \mathbf{w}^{(B)}_2,y\bm{\mu}_2\rangle.
\]
We write
\[
\mathbf{w}^{(A)} \;\overset{\mathrm{MSA}}{\succcurlyeq}\; \mathbf{w}^{(B)}
\]
and say that $\mathbf{w}^{(A)}$ has higher \emph{minority subclass alignment} than $\mathbf{w}^{(B)}$ if
\[
A^{(A)} \;\ge\; A^{(B)} \quad\text{and}\quad A^{(A)},A^{(B)} < B_{\max}.
\]
\end{definition}

\begin{theorem}[Higher MSA $\implies$ Better Generalization]\label{thm:gen}
Let $\mathbf{w}^{(A)},\mathbf{w}^{(B)}$ be two interpolating solutions trained on the same $D$. Under \Cref{def:datamodel} and regulatory conditions, if $\mathbf{w}^{(A)} \;\overset{\mathrm{MSA}}{\succcurlyeq}\; \mathbf{w}^{(B)}$, then
\[
\Pr\big(y\langle \mathbf{w}^{(A)},\mathbf{x}\rangle>0\mid (\mathbf{x},y)\sim\mathcal D)
\;\ge\;
\Pr\big(y\langle \mathbf{w}^{(B)},\mathbf{x}\rangle>0\mid (\mathbf{x},y)\sim\mathcal D),
\]
with strict inequality if $F_B((A^{(B)},A^{(A)}])>0$.
\end{theorem}

\Cref{thm:gen} shows that the interpolating model that aligns with the minority subclass feature more generalizes better.

\section{Proofs}
\label{apd:proofs}

\begin{assumption}
\label{asm:overlap}
We make the following assumptions:
\begin{enumerate}
    \item[(i)] (Kernel overlap) Let $P:=X^\top(XX^\top)^{-1}X$ be the orthogonal
    projector onto $\mathrm{span}(X^\top)$.
    The population-sharpness direction has nontrivial mass outside the data span:
    \[
        \|(I-P)\,\Sigma\,\widehat{\theta}_{G_0}\|
        \ \ge\ c \ >\ 0
    \]
    with probability $1-o(1)$ for some constant $c$.
    \item[(ii)] (Bounded interpolator) The Euclidean interpolator satisfies
    $\|\widehat{\theta}_{G_0}\| = O_{\mathbb{P}}(1)$.
\end{enumerate}
\end{assumption}

\begin{remark}[Justification of assumptions]
Assumption~\ref{asm:overlap}(i) is generic in the overparameterized regime
($d\gg n$): the vector $\Sigma\widehat{\theta}_{G_0}$ is a population signal
direction, while $\mathrm{span}(X^\top)$ is a random $n$-dimensional subspace.
Under mild spectral regularity of $\Sigma$, their alignment is not perfect with
high probability, yielding a non-negligible $(I-P)$ component.

Assumption~\ref{asm:overlap}(ii) is standard for benign overfitting and ensures
finite test variance. If
$\|\widehat{\theta}_{G_0}\|$ diverged, the Euclidean interpolator would have
exploding prediction variance and thus be unstable on unseen data.
\end{remark}

\begin{lemma}
\label{lem:KKT}
The unique solution of \eqref{eq:minG} is
\[
\widehat{\theta}_G
=
G^{-1}X^\top\Bigl(XG^{-1}X^\top\Bigr)^{-1}y,
\qquad\text{and}\qquad
X\widehat{\theta}_G = y.
\]
\end{lemma}
\begin{proof}
Form the Lagrangian
$L(\theta,\lambda)=\tfrac12\theta^\top G\theta+\lambda^\top(X\theta-y)$.
The KKT conditions are
$G\theta+X^\top\lambda=0$ and $X\theta=y$.
Eliminate $\theta=-G^{-1}X^\top\lambda$ to get
$-XG^{-1}X^\top\lambda=y$, hence
$\lambda=-(XG^{-1}X^\top)^{-1}y$.
Substituting back yields the stated $\widehat{\theta}_G$.
Since $G\succ0$, the objective is strictly convex and the solution is unique.
\end{proof}

\begin{lemma}
\label{lem:theta-derivative}
Let $\widehat{\theta}(\eta) := \widehat{\theta}_{G_\eta}$ with
$G_\eta = I + \eta\Sigma$.
Then
\[
\widehat{\theta}'(0)
:=
\frac{d}{d\eta}\widehat{\theta}(\eta)\Big|_{\eta=0}
=
-(I-P)\,\Sigma\,\widehat{\theta}_{G_0}.
\]
\end{lemma}
\begin{proof}
By Lemma~\ref{lem:KKT},
\[
\widehat{\theta}(\eta)
=
G_\eta^{-1}X^\top\!\bigl(XG_\eta^{-1}X^\top\bigr)^{-1}y.
\]
Differentiate at $\eta=0$.

(i) $\frac{d}{d\eta}G_\eta^{-1}=-G_\eta^{-1}\Sigma G_\eta^{-1}$, so
$\bigl.\frac{d}{d\eta}G_\eta^{-1}\bigr|_{\eta=0}=-\Sigma$.

(ii) Let $A(\eta):=XG_\eta^{-1}X^\top$. Then
$A'(0)=-X\Sigma X^\top$ and
\[
\bigl.\tfrac{d}{d\eta}A(\eta)^{-1}\bigr|_{\eta=0}
=
A(0)^{-1}(X\Sigma X^\top)A(0)^{-1}.
\]

Combining (i)–(ii),
\begin{align*}
\widehat{\theta}'(0)
&=
-\Sigma\,X^\top(XX^\top)^{-1}y
+
X^\top(XX^\top)^{-1}(X\Sigma X^\top)(XX^\top)^{-1}y \\
&=
-(I-P)\Sigma\,\widehat{\theta}_{G_0}.
\end{align*}
\end{proof}

\begin{lemma}[Distribution of model output on non-member data]
\label{lem:test_dist}
Let $X_{\mathrm{out}} \sim \mathcal{N}(0,\Sigma)$ independently of the training
set. Conditioned on $\widehat{\theta}_G$, the prediction
$f_G(X_{\mathrm{out}})=\widehat{\theta}_G^\top X_{\mathrm{out}}$ satisfies
\[
f_G(X_{\mathrm{out}})
\sim
\mathcal{N}\!\bigl(0,\sigma_G^2\bigr),
\qquad
\sigma_G^2 := \widehat{\theta}_G^\top \Sigma\,\widehat{\theta}_G .
\]
\end{lemma}
\begin{proof}
Condition on $\widehat{\theta}_G$. Since $X_{\mathrm{out}}$ is Gaussian and
$f_G$ is linear in $X_{\mathrm{out}}$, the claim follows with variance
$\widehat{\theta}_G^\top\Sigma\widehat{\theta}_G$.
\end{proof}

\begin{lemma}[First derivative of non-member variance at $\eta=0$]
\label{lem:var-derivative}
With $G_\eta = I + \eta\Sigma$, let
$\sigma^2(\eta):=\widehat{\theta}(\eta)^\top\Sigma\,\widehat{\theta}(\eta)$.
Then
\[
\sigma^{2\,\prime}(0)
=
2\,\widehat{\theta}_{G_0}^\top\Sigma\,\widehat{\theta}'(0)
=
-2\,\big\|(I-P)\Sigma\,\widehat{\theta}_{G_0}\big\|^2 .
\]
\end{lemma}
\begin{proof}
Differentiate:
$\sigma^{2\,\prime}(0)=2\,\widehat{\theta}_{G_0}^\top\Sigma\,\widehat{\theta}'(0)$
since $\Sigma$ is symmetric.
Insert Lemma~\ref{lem:theta-derivative}:
\[
\sigma^{2\,\prime}(0)
=
-2\,\widehat{\theta}_{G_0}^\top\Sigma(I-P)\Sigma\,\widehat{\theta}_{G_0}
=
-2\|(I-P)\Sigma\widehat{\theta}_{G_0}\|^2,
\]
using symmetry and idempotence of $(I-P)$.
\end{proof}

\paragraph{Proof of Theorem~\ref{thm:variance-strict}}
\begin{proof}
By Lemma~\ref{lem:var-derivative},
\[
\sigma^{2\,\prime}(0)
=
-2\|(I-P)\Sigma\widehat{\theta}_{G_0}\|^2 .
\]
Assumption~\ref{asm:overlap}(i) yields
$\|(I-P)\Sigma\widehat{\theta}_{G_0}\|\ge c$, hence
$\sigma^{2\,\prime}(0)\le -2c^2<0$ with probability $1-o(1)$.

Since $\widehat{\theta}(\eta)$ is smooth in $\eta$ for $G_\eta\succ0$,
$\sigma^2(\eta)$ is differentiable at $0$. Therefore, on the same high-probability
event, there exists $\eta_0>0$ such that for all $\eta\in(0,\eta_0]$,
\[
\sigma^2(\eta)
=
\sigma^2(0) + \eta\,\sigma^{2\,\prime}(0) + o(\eta)
<
\sigma^2(0).
\]
This implies $\sigma_{G_\eta}^2<\sigma_{G_0}^2$ for all sufficiently small
$\eta>0$ with probability $1-o(1)$.
\end{proof}

\subsection{Confidence-threshold attack}
\label{sec:conf-attack}

Let $\mathcal{I}_{\mathrm{in}}$ be the index set of member training points, and
let $\{(x_i,y_i)\}_{i\in\mathcal{I}_{\mathrm{in}}}$ denote the training samples.
For a geometry $G\succ 0$, let $\widehat{\theta}_G\in\mathbb{R}^d$ be the
(interpolating) solution of \eqref{eq:minG}.  
Define the signed score and confidence by
\[
f_G(x):=\widehat{\theta}_G^\top x,
\qquad
\mathrm{Conf}_G(x):=\bigl|f_G(x)\bigr|
=\bigl|\widehat{\theta}_G^\top x\bigr|.
\]

We model a black-box confidence-threshold attacker as follows.  
Let $I$ be a random index drawn from $\mathcal{I}_{\mathrm{in}}$
(e.g., uniformly), and let the random member pair be
\[
(X_{\mathrm{in}},Y_{\mathrm{in}}):=(x_I,y_I).
\]
Let $X_{\mathrm{out}}\sim\mathcal{N}(0,\Sigma)$ be an independent non-member
(test) input, independent of the training set and algorithmic randomness.
For a threshold $\tau\ge 0$, define
\begin{align}
\mathrm{TPR}_G(\tau)
&:=\Pr\!\left(\mathrm{Conf}_G(X_{\mathrm{in}})\ge \tau\right),\\
\qquad
\mathrm{FPR}_G(\tau)
&:=\Pr\!\left(\mathrm{Conf}_G(X_{\mathrm{out}})\ge \tau\right),\\
\qquad
\mathrm{Adv}^{\mathrm{conf}}_G
&:=\sup_{\tau\ge 0}\Bigl(\mathrm{TPR}_G(\tau)-\mathrm{FPR}_G(\tau)\Bigr).
\end{align}

\begin{lemma}
\label{lem:TPR-equal}
For any geometry $G\succ 0$,
\[
\mathrm{Conf}_G(X_{\mathrm{in}})=|Y_{\mathrm{in}}|
\quad\text{almost surely}.
\]
\end{lemma}

\begin{proof}
Because $\widehat{\theta}_G$ interpolates the training data, we have
$\widehat{\theta}_G^\top x_i=y_i$ for every $i\in\mathcal{I}_{\mathrm{in}}$.
Therefore for the random member index $I$,
\[
\mathrm{Conf}_G(X_{\mathrm{in}})
=\bigl|\widehat{\theta}_G^\top x_I\bigr|
=|y_I|
=|Y_{\mathrm{in}}|
\quad\text{a.s}.
\]
\end{proof}

\begin{lemma}
\label{lem:half-normal}
Conditioned on $\widehat{\theta}_G$,
\[
\mathrm{Conf}_G(X_{\mathrm{out}})
\overset{d}{=}|Z|,
\ \ Z\sim\mathcal{N}(0,\sigma_G^2),
\]
\end{lemma}

\begin{proof}
By Lemma~\ref{lem:test_dist} and absolute value. 
\end{proof}

\begin{lemma}
\label{lem:gauss-tail}
Let $0<\sigma_1<\sigma_2$ and let $Z_k\sim\mathcal{N}(0,\sigma_k^2)$
for $k\in\{1,2\}$. Then for every $\tau>0$,
\[
\Pr(|Z_1|\ge \tau)<\Pr(|Z_2|\ge \tau).
\]
\end{lemma}

\begin{proof}
Let $U\sim\mathcal{N}(0,1)$. Then $Z_k\overset{d}{=}\sigma_k U$, so
\[
\Pr(|Z_k|\ge \tau)=\Pr\!\left(|U|\ge \frac{\tau}{\sigma_k}\right).
\]
The function $t\mapsto \Pr(|U|\ge t)$ is strictly decreasing on $(0,\infty)$.
Since $\tau/\sigma_1>\tau/\sigma_2$, the claim follows.
\end{proof}

\begin{lemma}
\label{lem:tau-positive}
Assume $\Pr(|Y_{\mathrm{in}}|>0)>0$ (true whenever labels have any continuous
noise). Then
\[
\mathrm{Adv}^{\mathrm{conf}}_G
=
\sup_{\tau>0}\Bigl(\mathrm{TPR}_G(\tau)-\mathrm{FPR}_G(\tau)\Bigr).
\]
\end{lemma}

\begin{proof}
At $\tau=0$, $\mathrm{TPR}_G(0)=\mathrm{FPR}_G(0)=1$, so the gap equals $0$.
Because $\Pr(|Y_{\mathrm{in}}|>0)>0$, we have $\mathrm{TPR}_G(\tau)>0$ for some
$\tau>0$. Also, by Lemma~\ref{lem:half-normal},
$\mathrm{FPR}_G(\tau)\to 0$ as $\tau\to\infty$.
Hence there exists a $\tau>0$ such that
$\mathrm{TPR}_G(\tau)-\mathrm{FPR}_G(\tau)>0$.
Therefore the supremum cannot be attained at $\tau=0$, and we may restrict to
$\tau>0$.
\end{proof}

\paragraph{Proof of Theorem~\ref{thm:adv-conf}}

\begin{proof}
By Lemma~\ref{lem:TPR-equal}, member confidences are geometry-invariant, so
\[
\mathrm{TPR}_{G_\eta}(\tau)=\mathrm{TPR}_{G_0}(\tau)
\quad\text{for all }\tau\ge 0.
\]
By Theorem~\ref{thm:variance-strict}, for all sufficiently small $\eta>0$,
\[
\sigma_{G_\eta}^2<\sigma_{G_0}^2
\quad\text{with probability }1-o(1).
\]
Condition on this high-probability event.  
Lemma~\ref{lem:half-normal} implies
$\mathrm{Conf}_{G_\eta}(X_{\mathrm{out}})\overset{d}{=}|Z_\eta|$ with
$Z_\eta\sim\mathcal{N}(0,\sigma_{G_\eta}^2)$, and similarly for $G_0$.
Then Lemma~\ref{lem:gauss-tail} yields that for every $\tau>0$,
\[
\mathrm{FPR}_{G_\eta}(\tau)
=\Pr(|Z_\eta|\ge \tau)
<
\Pr(|Z_0|\ge \tau)
=\mathrm{FPR}_{G_0}(\tau).
\]
Therefore, for every $\tau>0$,
\[
\mathrm{TPR}_{G_\eta}(\tau)-\mathrm{FPR}_{G_\eta}(\tau)
>
\mathrm{TPR}_{G_0}(\tau)-\mathrm{FPR}_{G_0}(\tau).
\]
Taking the supremum over $\tau>0$ and using Lemma~\ref{lem:tau-positive},
\[
\mathrm{Adv}^{\mathrm{conf}}_{G_\eta}
=
\sup_{\tau>0}\Bigl(\mathrm{TPR}_{G_\eta}(\tau)-\mathrm{FPR}_{G_\eta}(\tau)\Bigr)
>
\sup_{\tau>0}\Bigl(\mathrm{TPR}_{G_0}(\tau)-\mathrm{FPR}_{G_0}(\tau)\Bigr)
=
\mathrm{Adv}^{\mathrm{conf}}_{G_0}.
\]
This holds with probability $1-o(1)$.
\end{proof}

\subsection{Likelihood-ratio attack}
\label{sec:lra-attack}

    

We consider an oracle likelihood-ratio attack that uses the model output $f_G(x)=\widehat\theta_G^\top x$.
The attacker knows the true member and non-member score distributions and
performs the Neyman--Pearson likelihood-ratio test.

\paragraph{Score distributions.}
Recall the data model
$x\sim\mathcal N(0,\Sigma)$ and $y=\theta^{*\top}x+\xi$ with
$\xi\sim\mathcal N(0,\sigma_y^2)$ independent of $x$.
Let $(X_{\mathrm{in}},Y_{\mathrm{in}})$ be a random member training pair
obtained by sampling a random index from the training set.
Unconditionally over the training data draw, we have the same marginal law
as a fresh sample:
\[
X_{\mathrm{in}}\sim\mathcal N(0,\Sigma),
\qquad
Y_{\mathrm{in}}=\theta^{*\top}X_{\mathrm{in}}+\xi.
\]

\begin{lemma}[Member score is geometry-invariant Gaussian]
\label{lem:lra-member}
For any geometry $G$, the member score satisfies
\[
f_G(X_{\mathrm{in}})=Y_{\mathrm{in}}
\quad\text{a.s.}
\qquad\text{and hence}\qquad
f_G(X_{\mathrm{in}})
\sim \mathcal N(0,v_{\mathrm{in}}),
\]
where
\(
v_{\mathrm{in}}:=\theta^{*\top}\Sigma\theta^*+\sigma_y^2.
\)
\end{lemma}

\begin{proof}
Interpolation gives $X\widehat\theta_G=y$, so for every training point
$\widehat\theta_G^\top x_i=y_i$. Thus for a random member index $I$,
\[
f_G(X_{\mathrm{in}})
=\widehat\theta_G^\top x_I
=y_I
=Y_{\mathrm{in}}
\quad\text{a.s.}
\]
Unconditionally, $Y_{\mathrm{in}}=\theta^{*\top}X_{\mathrm{in}}+\xi$
with $X_{\mathrm{in}}\sim\mathcal N(0,\Sigma)$ and
$\xi\sim\mathcal N(0,\sigma_y^2)$ independent, so it is mean-zero Gaussian
with variance $v_{\mathrm{in}}$.
\end{proof}

\begin{lemma}[Non-member score is geometry-dependent Gaussian]
\label{lem:lra-nonmember}
Let $X_{\mathrm{out}}\sim\mathcal N(0,\Sigma)$ be independent of the training
set. Conditioned on $\widehat\theta_G$,
\[
f_G(X_{\mathrm{out}})
=\widehat\theta_G^\top X_{\mathrm{out}}
\sim \mathcal N(0,v_{\mathrm{out}}(G)),
\qquad
v_{\mathrm{out}}(G):=\widehat\theta_G^\top\Sigma\widehat\theta_G.
\]
\end{lemma}

\begin{proof}
Same argument as Lemma~\ref{lem:test_dist}. \qedhere
\end{proof}

\paragraph{Oracle likelihood-ratio test.}
Let $p_{\mathrm{in}}$ and $p_{\mathrm{out}}$ be the densities of
$\mathcal N(0,v_{\mathrm{in}})$ and $\mathcal N(0,v_{\mathrm{out}}(G))$,
respectively. The oracle LR score is
\[
\Lambda_G(s):=\log\frac{p_{\mathrm{in}}(s)}{p_{\mathrm{out}}(s)}.
\]

\begin{lemma}
\label{lem:lra-threshold}
For zero-mean Gaussians with variances $v_{\mathrm{in}}>0$ and
$v_{\mathrm{out}}(G)>0$, the LR test
$\Lambda_G(s)\ge t$ is equivalent to $|s|\ge \tau$ for some $\tau\ge0$.
Moreover, the optimal test for any fixed FPR is of this form.
\end{lemma}

\begin{proof}
For $s\in\mathbb R$,
\[
\Lambda_G(s)
=
-\frac12\log v_{\mathrm{in}}+\frac12\log v_{\mathrm{out}}(G)
-\frac{s^2}{2v_{\mathrm{in}}}+\frac{s^2}{2v_{\mathrm{out}}(G)}
=
C_G
+\frac{s^2}{2}\Bigl(\frac1{v_{\mathrm{out}}(G)}-\frac1{v_{\mathrm{in}}}\Bigr),
\]
where $C_G$ does not depend on $s$.
Thus $\Lambda_G(s)\ge t$ is equivalent to $s^2\ge \tau^2$ for some $\tau\ge0$,
i.e. $|s|\ge\tau$. Neyman--Pearson gives optimality of the LR test. Note that as we are in a setting where members are highly confident, $v_{\mathrm{in}} > v_{\mathrm{out}}$.
\end{proof}

Define the LR-attack advantage as the best achievable TPR--FPR gap over all
two-sided thresholds:
\[
\mathrm{Adv}^{\mathrm{LR}}_G
:=
\sup_{\tau\ge0}
\Bigl(
\Pr(|f_G(X_{\mathrm{in}})|\ge\tau)
-
\Pr(|f_G(X_{\mathrm{out}})|\ge\tau)
\Bigr).
\]

\paragraph{Proof of Theorem~\ref{thm:adv-lr}}

\begin{proof}
By Lemma~\ref{lem:lra-member},
\[
f_{G_\eta}(X_{\mathrm{in}})
\sim \mathcal N(0,v_{\mathrm{in}})
\quad\text{for all }\eta\ge0,
\]
so the member tail
\(\Pr(|f_{G_\eta}(X_{\mathrm{in}})|\ge\tau)\)
is geometry-invariant for every $\tau\ge0$.

By Lemma~\ref{lem:lra-nonmember},
\[
f_{G_\eta}(X_{\mathrm{out}})
\sim \mathcal N(0,v_{\mathrm{out}}(G_\eta)),
\qquad
v_{\mathrm{out}}(G_\eta)=\sigma_{G_\eta}^2.
\]
Theorem~\ref{thm:variance-strict} gives that, for all sufficiently small
$\eta>0$,
\[
v_{\mathrm{out}}(G_\eta) < v_{\mathrm{out}}(G_0)
\quad\text{with probability }1-o(1).
\]
Condition on this high-probability event. Then Lemma~\ref{lem:gauss-tail}
(applied to $|f_G(X_{\mathrm{out}})|$) yields that for every $\tau>0$,
\[
\Pr(|f_{G_\eta}(X_{\mathrm{out}})|\ge\tau)
<
\Pr(|f_{G_0}(X_{\mathrm{out}})|\ge\tau).
\]
Therefore, for every $\tau>0$,
\[
\Pr(|f_{G_\eta}(X_{\mathrm{in}})|\ge\tau)
-
\Pr(|f_{G_\eta}(X_{\mathrm{out}})|\ge\tau)
>
\Pr(|f_{G_0}(X_{\mathrm{in}})|\ge\tau)
-
\Pr(|f_{G_0}(X_{\mathrm{out}})|\ge\tau).
\]
Taking the supremum over $\tau>0$ on both sides gives
\(
\mathrm{Adv}^{\mathrm{LR}}_{G_\eta}
>
\mathrm{Adv}^{\mathrm{LR}}_{G_0}.
\)
(As in Lemma~\ref{lem:tau-positive}, $\tau=0$ yields zero gap, so the supremum
is attained for some $\tau>0$.)
This holds with probability $1-o(1)$.
\end{proof}

\begin{corollary}[Sample-adaptive LR monotonicity]
\label{cor:lira}
Fix a query point $z$:
\[
f_{\mathrm{in},z} \sim \mathcal{N}(0, v_{\mathrm{in},z}),\qquad
f_{\mathrm{out},z}(G) \sim \mathcal{N}(0, v_{\mathrm{out},z}(G)).
\]
Let $G_a,G_b$ be two geometries such that
$v_{\mathrm{out},z}(G_a) < v_{\mathrm{out},z}(G_b)$.
Then for every false positive rate $\alpha\in(0,1)$, the optimal
likelihood-ratio test at level~$\alpha$ attains strictly larger
true positive rate under $G_a$ than under $G_b$.
Equivalently, the per-sample ROC curve under $G_a$ strictly dominates that
under $G_b$, so any MIA advantage is strictly larger under $G_a$.
\end{corollary}

\begin{proof}
Fix $z$ and $k\in\{a,b\}$.
The LR between $\mathcal{N}(0,v_{\mathrm{in},z})$ and
$\mathcal{N}(0,v_{\mathrm{out},z}(G_k))$ is
\[
\Lambda_z(s;G_k)
=
\frac12\log\frac{v_{\mathrm{out},z}(G_k)}{v_{\mathrm{in},z}}
+
\frac{s^2}{2}\Bigl(\frac{1}{v_{\mathrm{out},z}(G_k)}
                   -\frac{1}{v_{\mathrm{in},z}}\Bigr).
\]
If $v_{\mathrm{in},z}\neq v_{\mathrm{out},z}(G_k)$ then $\Lambda_z(s;G_k)$ is a
strictly monotone function of $|s|$, so by the Neyman--Pearson lemma the
optimal level-$\alpha$ test is equivalent to a two-sided magnitude test
$|s|\ge\tau_k(\alpha)$ for some unique threshold $\tau_k(\alpha)>0$.

Write $f_{\mathrm{out},z}(G_k)\overset{d}{=}
\sqrt{v_{\mathrm{out},z}(G_k)}\,U$ with $U\sim\mathcal{N}(0,1)$.  The
constraint
$\Pr(|f_{\mathrm{out},z}(G_k)|\ge\tau_k(\alpha))=\alpha$
is then
\[
\alpha
=
\Pr\Bigl(|U|\ge \tfrac{\tau_k(\alpha)}
                    {\sqrt{v_{\mathrm{out},z}(G_k)}}\Bigr).
\]
Since $u\mapsto\Pr(|U|\ge u)$ is strictly decreasing on $(0,\infty)$ and
$v_{\mathrm{out},z}(G_a)<v_{\mathrm{out},z}(G_b)$, this forces
\[
\frac{\tau_a(\alpha)}{\sqrt{v_{\mathrm{out},z}(G_a)}}
=
\frac{\tau_b(\alpha)}{\sqrt{v_{\mathrm{out},z}(G_b)}}
\quad\Rightarrow\quad
\tau_a(\alpha)<\tau_b(\alpha).
\]

The member distribution $f_{\mathrm{in},z}$ is the same under $G_a$ and $G_b$, so
\[
\Pr\bigl(|f_{\mathrm{in},z}|\ge\tau_a(\alpha)\bigr)
>
\Pr\bigl(|f_{\mathrm{in},z}|\ge\tau_b(\alpha)\bigr).
\]
Thus at every FPR level $\alpha$ the optimal LR/LiRA test has strictly larger
TPR under $G_a$, which implies strict ROC dominance and the claimed advantage
comparison.
\end{proof}

\subsection{Higher Subclass Alignment leads to higher generalization }
\label{sec:gen_proof}

\begin{lemma}[High-dimensional near-orthogonality]\label{lm:HNO}
Let $\mathbf{x}_1,\dots,\mathbf{x}_N\in\mathbb R^d$ have i.i.d.\ $\mathcal N(0,1)$ entries. Then, as $d\to\infty$,
\[
\|\mathbf{x}_i\|^2 = d\,(1+o(1)) \quad\text{and}\quad 
\frac{\langle \mathbf{x}_i,\mathbf{x}_j\rangle}{\|\mathbf{x}_i\|\,\|\mathbf{x}_j\|}=o(1)
\]
for each fixed $i\neq j$, with probability tending to $1$. Moreover, the two conclusions hold \emph{uniformly over all $i\neq j$} with probability tending to $1$ provided $\log N = o(d)$.
\end{lemma}

\begin{proof}
For norms, $\|\mathbf{x}_i\|^2\sim\chi^2(d)$. Laurent--Massart's inequality implies that for all $t>0$,
\[
\Pr\!\big(\|\mathbf{x}_i\|^2-d \ge 2\sqrt{dt}+2t\big)\le e^{-t},\qquad
\Pr\!\big(d-\|\mathbf{x}_i\|^2 \ge 2\sqrt{dt}\big)\le e^{-t}.
\]
Taking $t=\varepsilon^2 d$ gives $\|\mathbf{x}_i\|^2=d(1\pm O(\varepsilon))$ with probability at least $1-2e^{-\varepsilon^2 d}$; hence $\|\mathbf{x}_i\|^2=d(1+o(1))$ w.h.p. and $\|\mathbf{x}_i\|=\sqrt d+\mathcal O_p(1)$.

For inner products, write $\langle \mathbf{x}_i,\mathbf{x}_j\rangle=\sum_{k=1}^d Z_k$ with $Z_k:=\mathbf{x}_{i,k}\mathbf{x}_{j,k}$, which are i.i.d., mean $0$, and sub-exponential. Bernstein's inequality yields
\[
\Pr\!\big(|\langle \mathbf{x}_i,\mathbf{x}_j\rangle|\ge t\big)\le 2\exp\!\Big(-c\min\{t^2/d,\; t\}\Big)
\]
for a universal $c>0$. Taking $t=C\sqrt d$ shows $|\langle \mathbf{x}_i,\mathbf{x}_j\rangle|=\mathcal O_p(\sqrt d)$. Combining with $\|\mathbf{x}_i\|\,\|\mathbf{x}_j\|=d(1+o_p(1))$,
\[
\frac{|\langle \mathbf{x}_i,\mathbf{x}_j\rangle|}{\|\mathbf{x}_i\|\,\|\mathbf{x}_j\|}=\mathcal O_p(d^{-1/2})=o_p(1).
\]
A union bound over the $\binom{N}{2}$ pairs then gives the uniform statement whenever $\log N=o(d)$, since both tails are $\exp(-\Theta(d))$.
\end{proof}

\paragraph{Noise weights}
By the representer theorem, write
\[
\mathbf{w}_3=\sum_{j=1}^n \beta_j\, y_j\, \mathbf{x}_{3,j}.
\]
Then, for any training point $i$,
\[
y_i\langle \mathbf{w}_3,\mathbf{x}_{3,i}\rangle
=
\beta_i\,\|\mathbf{x}_{3,i}\|^2
\;+\;
\zeta_i,
\qquad
\zeta_i \;\coloneqq\; \sum_{j\neq i}\beta_j\, y_i y_j\, \langle \mathbf{x}_{3,j},\mathbf{x}_{3,i}\rangle .
\]
Under \Cref{lm:HNO}, $\|\mathbf{x}_{3,i}\|^2=(1\pm o(1))\,d_3$ and the cross inner products are
$o(\|\mathbf{x}_{3,i}\|\,\|\mathbf{x}_{3,j}\|)=o(d_3)$

\paragraph{Condition 1.}
For each minority point $i\in\mathcal S$, define
\[
B_i \;\coloneqq\; -\,y_i\langle \mathbf{w}_1,\bm{\nu}_i\rangle \;>\; 0,
\qquad
A \;\coloneqq\; \langle \mathbf{w}_2,\bm{\mu}_2\rangle .
\]
Let $B$ be a random variable with CDF $F_B(A) = \Pr(B < A)$ such that $B \stackrel{d}{=} B_i$ (i.e., $F_B$ is the law/distribution of the $B_i$’s when $i$ is drawn uniformly from $\mathcal S$).
We assume
\[
A \;<\; B_{\max},
\qquad
B_{\max} \triangleq \sup\{\, b \in \mathbb{R} \mid F_B(b) < 1 \,\},
\]

\begin{remark}
For each minority sample $i$, the anti-alignment magnitude
\(
B_i \triangleq -\,y_i\langle \mathbf{w}_1,\bm{\nu}_i\rangle>0
\)
summarizes how strongly the majority anchor opposes the minority anchor for that sample.
We assume $\{B_i\}_{i\in\mathcal S}$ are i.i.d.\ draws from a common distribution $F_B$ supported on $(0,B_{\max}]$.
Unseen minority test point has margin
\(
m'=-B+A+\zeta', \text{ with } B\sim F_B,
\)
so
\(
\Pr_{\mathcal S}(m'>0) = \Pr(B<A) = F_B(A)
\)
(up to the $o(1)$ fluctuation $\zeta'$ from \Cref{lm:HNO}).
Intuitively, $F_B(A)$ is the fraction of minority subclasses whose majority anti-alignment is not too strong relative to the learned shared minority signal $A$.
\end{remark}

\begin{definition}
[Generalization gap]
The generalization gap for a training point $i$ from distribution $\mathcal{D}_{\mathcal{K}} \in {\mathcal{D}_{\mathcal{S}}, \mathcal{D}_{\mathcal{M}}, \mathcal{D}_{\mathcal{N}}}$ is defined as
\begin{align}
\mathrm{R}^{\mathcal{K}}_i(w)\;\coloneqq\;
\Pr\{y_i\langle \mathbf{w},\mathbf{x}_i\rangle>0\}
-\Pr\!\big(y'\langle \mathbf{w},\mathbf{x}'\rangle>0 \big),
\quad
(\mathbf{x}',y') \sim \mathcal{D_{\mathcal{K}}}.
\end{align}
\end{definition}

\begin{assumption}[Majority alignment]\label{asm:maj_align}
We consider models whose first block weights align with the majority subclass feature:
\[
\mathbf{w}_1 \;=\; \alpha\,\bm{\mu} \qquad (\alpha>0),
\]
This captures the implicit bias of common ERM procedures (e.g., logistic regression trained by gradient descent, or minimum-$\ell_2$-norm interpolation) to align with the dominant signal in the data.
\end{assumption}   

\begin{assumption}[Majority dominance]\label{asm:maj_dom}
We assume the majority signal dominates the stochastic parts:
\[
\frac{\langle \mathbf{w}_1,\bm{\mu}_1\rangle^2}{\|\mathbf{w}_2\|^2+\|\mathbf{w}_3\|^2}\;\xrightarrow[]{}\;\infty.
\]
\end{assumption}

\begin{assumption}[Perfect interpolation and finite-margin] \label{asm:finite_margin}
A trained model $\mathbf{w}$ has finite margins. That is, there exist $0<m_0\le M<\infty$ such that
\[
m_0 \;\le\; y_i\langle \mathbf{w},\mathbf{x}_i\rangle \;\le\; M \qquad \forall i.
\]
\end{assumption}

\begin{lemma}[Generalization gap of majority subclass samples]\label{lem:maj}
Let $i\in\mathcal M$. Then with probability $1-o(1)$,
\[
\mathrm{R}^{\mathcal{M}}_i(\mathbf{w})=o(1)
\]
\end{lemma}
\begin{proof}
By Assumption~\ref{asm:finite_margin}, $y_i\langle \mathbf{w},\mathbf{x}_i\rangle>0$, hence $\Pr\{y_i\langle \mathbf{w},\mathbf{x}_i\rangle>0\}=1$.

\smallskip
\noindent
Draw $(x',y')\sim\mathcal D_{\mathcal M}$, so $x'_1=y'\bm{\mu}_1$, $x'_2\sim\mathcal N(0,I_{d_2})$, $x'_3\sim\mathcal N(0,I_{d_3})$, independent of $y'$. Then
\[
y'\langle w,x'\rangle
=\langle \mathbf{w}_1,\bm{\mu}_1\rangle + y'\langle \mathbf{w}_2,x'_2\rangle + y'\langle \mathbf{w}_3,x'_3\rangle
=: \langle \mathbf{w}_1,\bm{\mu}_1\rangle + Z,
\]
where $Z$ is a mean-zero sub-Gaussian random variable with variance proxy
\(
\mathrm{VarProxy}(Z)=\|\mathbf{w}_2\|^2+\|\mathbf{w}_3\|^2,
\)
since $y'\langle \mathbf{w}_2,\mathbf{x}'_2\rangle\sim \mathcal N(0,\|\mathbf{w}_2\|^2)$ and $y'\langle \mathbf{w}_3,\mathbf{x}'_3\rangle\sim \mathcal N(0,\|\mathbf{w}_3\|^2)$ are independent and centered.

For any $a>0$, a standard sub-Gaussian tail bound yields
\[
\Pr\{Z\le -a\}\;\le\;\exp\!\Big(-\,\tfrac{a^2}{2(\|\mathbf{w}_2\|^2+\|\mathbf{w}_3\|^2)}\Big).
\]
Taking $a=\langle \mathbf{w}_1,\bm{\mu}_1\rangle$,
\[
\Pr\big\{y'\langle w,\mathbf{x}'\rangle\le 0\big\}
=\Pr\{Z\le -\langle \mathbf{w}_1,\bm{\mu}_1\rangle\}
\;\le\;
\exp\!\Big(-\,\frac{\langle \mathbf{w}_1,\bm{\mu}_1\rangle^2}{2(\|\mathbf{w}_2\|^2+\|\mathbf{w}_3\|^2)}\Big).
\]
Hence
\[
\Pr\{y'\langle w,\mathbf{x}'\rangle>0\}
\;\ge\;
1-\exp\!\Big(-\,\frac{\langle \mathbf{w}_1,\bm{\mu}_1\rangle^2}{2(\|\mathbf{w}_2\|^2+\|\mathbf{w}_3\|^2)}\Big).
\]

By Assumption~\ref{asm:maj_dom},
\[
\mathrm{R}^{\mathcal{M}}_i(\mathbf{w})
= 1 - \Pr\{y'\langle \mathbf{w},\mathbf{x}'\rangle>0\}
\;\le\;
\exp\!\Big(-\,\frac{\langle \mathbf{w}_1,\bm{\mu}_1\rangle^2}{2(\|\mathbf{w}_2\|^2+\|\mathbf{w}_3\|^2)}\Big)
= o(1)
\]

\end{proof}

\begin{lemma}[Generalization gap of minority subclass samples]\label{lem:min}
Let $i\in\mathcal S$.
\[
\mathrm{R}^{\mathcal S}_i(\mathbf{x})
\;=\;
 1
- F_B(A)
\quad \text{(up to $o(1)$ terms).}
\]
\end{lemma}

\begin{proof}
By Assumption~\ref{asm:finite_margin}, $y_i\langle \mathbf{w},\mathbf{x}_i\rangle>0$, hence $\Pr\{y_i\langle \mathbf{w},\mathbf{x}_i\rangle>0\}=1$.

Draw $(\mathbf{x}',y')\sim \mathcal{D}_{\mathcal{S}}$. $\mathbf{x}_3'$ is independent of $\{\mathbf{x}_{3,j}\}$ and mean-zero; hence
\(
y\langle \mathbf{w}_3,\mathbf{x}_3'\rangle=\sum_j \beta_j y y_j \langle \mathbf{x}_{3,j},\mathbf{x}_3'\rangle
\)
is a mean-zero fluctuation with variance vanishing relative to $\|\mathbf{x}_3'\|^2$ ; set this fluctuation to $\zeta' = o_\mathbb{P}(1)$ using \Cref{lm:HNO}. Then, $y'\langle \mathbf{w},\mathbf{x}'\rangle=-B+A+\zeta'$.

\[
\Pr\!\big(y'\langle w,\mathbf{x}'\rangle>0\big)
=
\Pr(B<A-\zeta') \in \big(F_B(A-\varepsilon),\,F_B(A+\varepsilon)\big).
\]
Letting $\varepsilon\downarrow 0$ gives the stated identities up to $o(1)$.
\end{proof}

\begin{lemma}[Generalization gap for pure noise samples]\label{lm:noise}
Let $i\in\mathcal N$ (pure noise). 
\[
\mathrm{R}^{\mathcal N}_i(\mathbf{w})
\;=\;
\frac{1}{2}
\]
\end{lemma}

\begin{proof}
By Assumption~\ref{asm:finite_margin}, $y_i\langle \mathbf{w},\mathbf{x}_i\rangle>0$, hence $\Pr\{y_i\langle \mathbf{w},\mathbf{x}_i\rangle>0\}=1$.

For an unseen noise $(\mathbf{x}',y') \sim \mathcal{D}_{\mathcal{N}}$, each term $y'\langle \mathbf{w}_k,\mathbf{w}_k'\rangle$ ($k=1,2,3$) is a centered continuous symmetric random variable (linear form of a mean-zero isotropic vector, independent of $y'$). The sum remains centered and symmetric; hence $\Pr(y'\langle \mathbf{w},\mathbf{x}'\rangle>0)=1/2$.
\end{proof}

\paragraph{Proof of \Cref{thm:gen}}
\begin{proof}
By \Cref{lem:min}, minority test accuracy equals $F_B(A)$. This is monotone increasing in $A$. The majority subclass and noise samples yield the same result for both models by \Cref{lem:maj,lm:noise}.
\end{proof}

\section{Additional related works}
\label{sec:additionalRelatedWorks}
\subsection{Connection of Flatter minima with Generalization gap }
There have been numerous studies \citep{foret2020sharpness,izmailov2018averaging,cha2021swad,norton2021diametrical,wu2020adversarial} which account for the worst-case empirical risks within neighborhoods in parameter space. Diametrical Risk Minimization (DRM) was first proposed by \citep{norton2021diametrical} and they asserted that the practical and theoretical performance of Empirical Risk Minimization (ERM) tends to suffer when dealing with loss functions that exhibit poor behavior characterized by large Lipschitz moduli and spurious sharp minimizers. They tackled this concern by employing DRM, which offers generalization bounds that are unaffected by Lipschitz moduli, applicable to both convex and non-convex problems. Another algorithm that improves generalization is Sharpness Aware Minimization (SAM) \citep{foret2020sharpness} which performs gradient descent while regularizing for the highest loss in the neighborhood of radius $\rho$ of the parameter space.
\citep{izmailov2018averaging} proposed Stochastic Weight Averaging (SWA) that performs averaging of weights with a cyclical or constant learning rate which leads to better generalization than conventional training. They also prove that the optima chosen by the single model is in fact a flatter minima than the SGD solution. Further, \citep{cha2021swad} argues that simply performing the Empirical Risk Minimization (ERM) is not enough to achieve at a good generalization, in particular, domain generalization. Hence, they introduce SWAD which seeks for flatter optima and hence, will generalize well across domain shifts.

\subsection{Different Membership Inference attacks} \label{app:mia_attacks}
There are many variants of Direct Single-query attacks (DSQ) based on the approach of the attack and below we describe the ones used in our experiments:

\paragraph{NN-based attack \citep{shokri2017membership,tang2022mitigating,nasr2018machine}} This is the first MIA proposed by \citet{shokri2017membership} where they use a binary classifier to distinguish between the training members and the non-members using the victim model's behavior on these data points. The adversary can utilize the prediction vectors from the target model and incorporate them along with the one-hot encoded ground truth labels as inputs. Then, they can construct a neural network $(I_{NN})$ called attack model.
\paragraph{Confidence-based attack \citep{yeom2020overfitting,salem2018ml,song2021systematic}} If the highest prediction confidence of an input record exceeds a predetermined threshold, the adversary considers it a member; otherwise, it is inferred as a non-member. This approach is based on the understanding that the target model is trained to minimize prediction loss using its training data, implying that the maximum confidence score of a prediction vector for a training member should be near 1. The attack $I_{conf}$ is defined as follows:
\begin{align}
    I_{conf}\hat{p}(y|\mathbf{x}) = \mathds{1}(\text{max }\hat{p}(y|\mathbf{x}) \geq \tau)
\end{align}
Here, $\mathds{1}(\cdot)$ is an indicator function which returns 1 if the predicate inside it holds True else the function evaluates to 0.

\paragraph{Entropy-based attack \citep{nasr2019comprehensive, song2021systematic, tang2022mitigating}} When the prediction entropy of an input record falls below a predetermined threshold, the adversary considers it a member. Conversely, if the prediction entropy exceeds the threshold, the adversary infers that the record is a non-member. This inference is based on the observation that there are notable disparities in the prediction entropy distributions between training and test data. Typically, the target model exhibits higher prediction entropy on its test data compared to its training data. The entropy of a prediction vector $p(\hat{y}|x)$ is defined as follows:
\begin{align}
    H(p(\hat{y}|\mathbf{x})) = - \sum_i(p_i log(p_i))
\end{align}
where $p_i$ is the confidence score in $p(\hat{y}|\mathbf{x})$. Then, the attack $I_{entr}$ is given as:
\begin{align}
    I_{entr}(\hat{p}(y|\mathbf{x}),y) = \mathds{1}(H(p(\hat{y}|x)) \leq \tau)
\end{align}

\paragraph{Modified entropy-based attack \citep{song2021systematic}} Song et al.[15] introduced an enhanced prediction entropy metric that integrates both the entropy metric and the ground truth labels. The modified entropy metric tends to yield lower values for training samples compared to testing samples. To infer membership, either a class-dependent threshold $\tau_y$ or a class-independent threshold $\tau_{attack}$ is applied.
\begin{align}
    I_{Mentr}(\hat{p}(y|\mathbf{x}),y) = \mathds{1}(Mentr(p(\hat{y}|\mathbf{x})) \leq \tau_y)
\end{align}
where $Mentr(p(\hat{y}|\mathbf{x}))$ for (x,y) data sample is given by combination of entropy information and ground truth label as:
\begin{align}
    Mentr(p(\hat{y}|\mathbf{x})) = - ((1 - p(\hat{y}|\mathbf{x})_y) log(p(\hat{y}|\mathbf{x})_y) - \sum_{i \neq y}(p(\hat{y}|\mathbf{x})_i log(1- p(\hat{y}|\mathbf{x})_i)))
\end{align}

\paragraph{Likelihood Ratio Attack (LiRA) \citep{lira}} 
LiRA is a shadow-model based single–query attack that explicitly models the distributions of a scalar score for members and non-members and then performs a likelihood ratio test. 
For a sample $(\mathbf{x},y)$, the attacker first defines a one-dimensional score $s(\mathbf{x},y)$ from the target model, typically the negative cross-entropy loss or the (log-)confidence on the true label $y$. Using multiple shadow models trained with and without $(\mathbf{x},y)$, the attacker estimates two score distributions: one for members (IN) and one for non-members (OUT). 
In practice, LiRA fits parametric Gaussians
\[
s(\mathbf{x},y)\mid \mathrm{IN} \sim \mathcal{N}(\mu_{\mathrm{in}},\sigma_{\mathrm{in}}^2), 
\qquad 
s(\mathbf{x},y)\mid \mathrm{OUT} \sim \mathcal{N}(\mu_{\mathrm{out}},\sigma_{\mathrm{out}}^2),
\]
and computes the log-likelihood ratio
\[
\Lambda_{\mathrm{LiRA}}(\mathbf{x},y)
=
\log\frac{\phi\!\left(s(\mathbf{x},y);\mu_{\mathrm{in}},\sigma_{\mathrm{in}}^2\right)}
         {\phi\!\left(s(\mathbf{x},y);\mu_{\mathrm{out}},\sigma_{\mathrm{out}}^2\right)},
\]
where $\phi(\cdot;\mu,\sigma^2)$ denotes the Gaussian density. 
The LiRA decision rule is then
\begin{align}
    I_{\mathrm{LiRA}}(\hat{p}(y|\mathbf{x}),y)
    =
    \mathds{1}\big(\Lambda_{\mathrm{LiRA}}(\mathbf{x},y) \ge \tau_{\mathrm{LiRA}}\big),
\end{align}
for some threshold $\tau_{\mathrm{LiRA}}$ chosen to trade off between TPR and FPR. 
In the ``online'' variant, the attacker fits both IN and OUT distributions from shadow models; in the ``offline'' variant, only the OUT distribution is estimated and low likelihood under the OUT model is treated as evidence of membership.

\paragraph{Robust MIA (RMIA) \citep{rmia}}
RMIA reframes membership inference as a calibrated hypothesis test based on a \emph{pairwise likelihood ratio} between a query $(\mathbf{x},y)$ and many population samples $(\mathbf{z},y_\mathbf{z})$. For each pair $(\mathbf{x},\mathbf{z})$, RMIA compares how the (approximate) probability of $\mathbf{x}$ and $\mathbf{z}$ change when conditioning on the event that $\mathbf{x}$ was used to train the target model. Concretely, RMIA defines
\[
\mathrm{LR}(\mathbf{x},\mathbf{z})
\;=\;
\frac{\Pr(\mathbf{x} \mid \theta)}{\Pr(\mathbf{z} \mid \theta)}
\;\Big/\;
\frac{\Pr(\mathbf{x})}{\Pr(\mathbf{z})},
\]
where $\Pr(\cdot \mid \theta)$ denotes the target model’s likelihood  and $\Pr(\cdot)$ is a population prior. Intuitively, if including $\mathbf{x}$ in the target training set fits $\mathbf{x}$  disproportionately better than many other population points $\mathbf{z}$, the ratio $\mathrm{LR}(\mathbf{x},\mathbf{z})$ becomes large. RMIA samples many $\mathbf{z}$ from the population and defines a robust membership score
\[
R(\mathbf{x})
\;=\;
\frac{1}{|n_z|}\sum_{\mathbf{z}}\mathds{1}\!\big(\mathrm{LR}(\mathbf{x},\mathbf{z}) > \gamma\big),
\]
where $\gamma$ is a fixed pairwise LR threshold and $n_z$ is number of population (non-member) samples. The attack then declares membership if $R(\mathbf{x})$ exceeds a global threshold $\tau$; by sweeping $\tau$ one obtains a calibrated ROC curve, and for a chosen FPR one can directly pick the corresponding $\tau$. In the \emph{offline} mode, all reference models are OUT models trained once on population data; in the \emph{online} mode, the attacker additionally trains IN reference models that explicitly include $\mathbf{x}$ in their training set, which yields a more accurate approximation of the conditional likelihoods but is more computationally expensive. 

\begin{table}[t]
  \caption{Attack accuracy of direct threshold MIA on SGD, Sharp, ASAM, and GSAM. Analogous to SAM, optimization methods that improve generalization (ASAM, GSAM) through finding flatter minima tend to be more prone to direct threshold attacks, while optimization that looks for sharp minima instead is more robust to MIA attack while being worse off in generalization.}
  \label{dsq-attacks-extended}
  \centering
  \begin{small}
  \begin{tabular}{ccccccc}
    \toprule 
    Dataset       & Algo   & NN   & Confidence    & Entropy & M-entropy &Test Acc\\
    \midrule
    \multirow{4}{*}{CIFAR-100}  
                & SGD    & 76.62\%        & 77.19\%   & 76.61\%   & 77.30\% & 80.30\%\\
                & Sharp  & 57.62\%      & 59.69\%    & 57.88\% & 59.69\% & 76.14\% \\   
                & ASAM   & \textcolor{red}{78.92\%}        & 79.22\%    & 78.86\% & \textcolor{red}{79.31\%} & 81.80\% \\
                & GSAM   & 78.63\%      & \textcolor{red}{79.23\%}    & \textcolor{red}{79.00\%} & 79.23\% & \textcolor{blue}{82.16\%} \\
    \midrule
    \multirow{4}{*}{CIFAR-10}   
                & SGD    & 50.00\%      & 59.37\%    & 59.09\% & 59.51\% & 96.00\%\\                  
                & Sharp  & 50.22\%      & 52.86\%    & 52.47\% & 52.78\% & 92.86\%\\ 
                & ASAM   & \textcolor{red}{50.48\%}        & 61.39\%    & 61.20\% & 61.32\% & \textcolor{blue}{96.66\%} \\
                & GSAM   & 50.00\%      & \textcolor{red}{61.46\%}    & \textcolor{red}{61.38\%} & \textcolor{red}{61.54\%} & 96.64\% \\
    \midrule
    \multirow{4}{*}{Purchase100}    
                & SGD    & 66.00\%      & 66.76\%    & 64.78\% & 67.13\% & 85.50\%\\                  
                & Sharp  & 59.58\%      & 60.96\%    & 58.04\% & 61.16\% & 84.31\%\\ 
                & ASAM   & 66.85\%      & 66.84\%    & 65.39\% & 67.03\% & 85.54\% \\
                & GSAM   & \textcolor{red}{67.45\%}      & \textcolor{red}{67.72\%}    & \textcolor{red}{66.51\%} & \textcolor{red}{67.87\%} & \textcolor{blue}{85.82\%} \\
    \midrule
    \multirow{4}{*}{Texas100}   
                & SGD    & 59.81\%      & 65.20\%    & 55.74\% & 65.13\% & 50.83\%\\                  
                & Sharp  & 51.11\%      & 59.89\%    & 53.46\% & 59.36\% & 49.97\%\\ 
                & ASAM   & \textcolor{red}{60.92\%}        & \textcolor{red}{67.50\%}    & \textcolor{red}{58.80\%} & 67.10\% & \textcolor{blue}{53.17\%} \\
                & GSAM   & 54.89\%      & 67.07\%    & 57.93\% & \textcolor{red}{67.13\%} & 52.04\% \\
    \bottomrule
  \end{tabular}
  \end{small}
\end{table}
\section{Other Sharpness-aware Optimizers}
\label{app:other_sharp}
In this section, we discuss other variants of SAM, namely Adaptive SAM (ASAM) \citep{kwon21b}, Guided SAM (GSAM) \citep{gsam}, and custom designed optimizer, namely Sharp. Sharp objective is designed to explicitly find a sharper minima. The objective function of Sharp is,

\begin{align}\label{eq:sharpminima}  
\mathcal{L}_{\mathrm{Sharp}}(\mathbf{w}) 
= L(\mathbf{w}) - \beta \max_{\epsilon \in B(\rho)} L(\mathbf{w}+\epsilon).  
\end{align}

This objective can be seen as minimizing the loss at current $w$ while maximizing the loss in the vicinity. We empirically verify that this objective does lead to a sharper minima measuring its hessian trace. Results and discussion about Sharp are available in \Cref{sharpdetail}.

The results on CIFAR10, CIFAR100, Purchase100, and Texas100 are reported in Table~\ref{dsq-attacks-extended}. Other sharpness-aware optimizers are shown to achieve similar generalization gain, albeit at the cost of higher membership attack accuracy. On the other hand, optimizer that explicitly looks for a sharp minima does worse in terms of generalization, but has better membership privacy. 

\begin{table}[t]
\centering
\caption{Comparison of \textbf{offline} shadow model MIA on SGD and SAM. In \textcolor{blue}{blue} we highlight the best performing model on the test set, and in \textcolor{red}{red} the model with higher privacy leakage (higher AUC, Attack Accuracy, and TPR@0.1\%FPR).}
\label{tab:mia_offline}
\begin{small}
\setlength{\tabcolsep}{4pt}
\begin{tabular}{llcccccccc}
\toprule
 & & \multicolumn{4}{c}{SGD} & \multicolumn{4}{c}{SAM} \\
\cmidrule(lr){3-6} \cmidrule(lr){7-10}
Dataset & Attack & Test Acc & AUC & Attack Acc & TPR@.1 & Test Acc & AUC & Attack Acc & TPR@.1 \\
\midrule

\multirow{2}{*}{CIFAR-100} 
 & RMIA  & \multirow{2}{*}{67.7\%} & 86.8\% & 77.3\% & 17.3\% & \multirow{2}{*}{\textcolor{blue}{69.1\%}} & \textcolor{red}{87.7\%} & \textcolor{red}{78.1\%} & \textcolor{red}{18.9\%} \\
 & LiRA  &                         & 76.2\% & 71.9\% & 16.9\% &                                            & \textcolor{red}{77.8\%} & \textcolor{red}{73.2\%} & \textcolor{red}{19.4\%} \\
\midrule

\multirow{2}{*}{CIFAR-10} 
 & RMIA  & \multirow{2}{*}{92.3\%} & 69.4\% & 62.3\% & 4.3\% & \multirow{2}{*}{\textcolor{blue}{93.1\%}} & \textcolor{red}{72.7\%} & \textcolor{red}{64.6\%} & \textcolor{red}{5.7\%} \\
 & LiRA  &                         & 54.1\% & 55.9\% & 4.1\% &                                            & \textcolor{red}{58.7\%} & \textcolor{red}{58.3\%} & \textcolor{red}{7.0\%} \\
\midrule

\multirow{2}{*}{Purchase100} 
 & RMIA  & \multirow{2}{*}{76.5\%} & 68.7\% & 62.6\% & 1.7\% & \multirow{2}{*}{\textcolor{blue}{77.4\%}} & \textcolor{red}{70.2\%} & \textcolor{red}{63.6\%} & \textcolor{red}{1.9\%} \\
 & LiRA  &                         & 52.9\% & 53.4\% & 0.1\% &                                            & \textcolor{red}{53.7\%} & \textcolor{red}{54.2\%} & 0.1\% \\
\midrule

\multirow{2}{*}{Texas100} 
 & RMIA  & \multirow{2}{*}{46.9\%} & 74.8\% & 67.4\% & 3.6\% & \multirow{2}{*}{\textcolor{blue}{49.2\%}} & \textcolor{red}{78.8\%} & \textcolor{red}{69.8\%} & \textcolor{red}{5.6\%} \\
 & LiRA  &                         & 56.9\% & 58.3\% & 0.8\% &                                            & \textcolor{red}{61.7\%} & \textcolor{red}{61.5\%} & \textcolor{red}{2.6\%} \\

\bottomrule
\end{tabular}
\end{small}
\end{table}

\section{Offline Shadow Model Attacks}
\label{app:offline}
We report the results for offline shadow model attacks in Table~\Cref{tab:mia_offline}. The results are in line and support the finding that SAM tends to incur higher membership privacy leakage. Excluding cases for tabular datasets where TPR at 0.1\% FPR is near zero for both models, SAM has higher values for all other attack metrics. Consistent with the literature, RMIA is more effective for offline setting compared to LiRA~\citep{rmia}. For online setting, we use a different experimental setup (WideResNet targets and shadows, our own training pipeline, and a different choice of auxiliary z-points). In this setting, LiRA is slightly stronger than RMIA across most metrics (Table~\ref{dsq-attacks-extended}), but the gaps are modest and much smaller than those reported for the offline comparison in~\citet{rmia}. We therefore view our results as broadly consistent with prior work: RMIA and LiRA are competitive state-of-the-art shadow-model attacks, and their exact ranking can depend on architectural and training choices. Our main conclusions—in particular, that SAM consistently increases vulnerability to both LiRA and RMIA compared to SGD—are unaffected by these small differences.

\section{Datasets} \label{data}
Here we introduce the four benchmark datasets used in the experiments and they have been widely used in prior works on MI attacks:
\paragraph{CIFAR-10}\footnote{https://www.cs.toronto.edu/~kriz/cifar.html}
This is a benchmark dataset for image classification task. The dataset consists of 60,000 color images of 32x32 size. There are 6,000 images from 10 classes where 5,000 images per class belong to the training dataset and 1,000 images per class belong to the test dataset. 

\paragraph{CIFAR-100}\footnote{https://www.cs.toronto.edu/~kriz/cifar.html}  The dataset is designed to be more challenging than CIFAR-10 as it contains a greater number of classes and more fine-grained distinctions between objects. There are a total of 60,000 images from 100 classes. Each subclass consists of 600 images, and within each subclass, there are 500 training images and 100 testing images. This distribution ensures a balanced representation of each class in both the training and testing sets.
 
\paragraph{Purchase-100} \footnote{https://www.kaggle.com/c/acquire-valued-shoppers-challenge} This a 100 class classification task with 197,324 data samples and consists of 600 binary feature; each dimension corresponds to a product and its value states if corresponding customer purchased the product; the corresponding label represents the shopping habit of the customer. We use the pre-processed and simplified version provided by \citet{shokri2017membership} and used by \citet{tang2022mitigating}.

\paragraph{Texas-100} \footnote{https://www.dshs.texas.gov/THCIC/Hospitals/Download.shtm.} This dataset is based on the Hospital Discharge Data public files with information about inpatients stays in several health facilities released by the Texas Department of State Health Services from 2006 to 2009. We used a prepossessed and simplified version of this dataset provided by \citep{shokri2017membership} and used by \citep{tang2022mitigating} which is composed of 67,330 data samples with 6,170 binary features. Each feature represents a patient's medical attribute like the external causes of injury, the diagnosis and other generic information.The classification task is to classify patients into 100 output classes which represent the main procedure that was performed on the patient.

\paragraph{EyePacs}
\footnote{https://www.kaggle.com/datasets/mariaherrerot/eyepacspreprocess} The pre-processed version of this dataset is obtained from Kaggle and it was originally used for a Diabetic Retinopathy Detection challenge. The dataset consists of 88,702 colour fundus images, including 35,126 samples for training and 53,576 samples for testing. The images were captured under various conditions by various devices at multiple primary care sites throughout California and elsewhere. For each subject, two images of the left and right eyes were collected, with the same resolution. A clinician was asked to rate each image for the presence of DR with a scale of 0–4 according to the Early Treatment Diabetic Retinopathy Study (ETDRS) scale. Note that for this dataset only training set (35k images) is used since the labels for testing set is not publicly available. The images in the dataset vary in their image resolution and we resized all the images to 128x128 pixels for our experiments.

\begin{figure}
  \centering
  \includegraphics[scale=0.28]{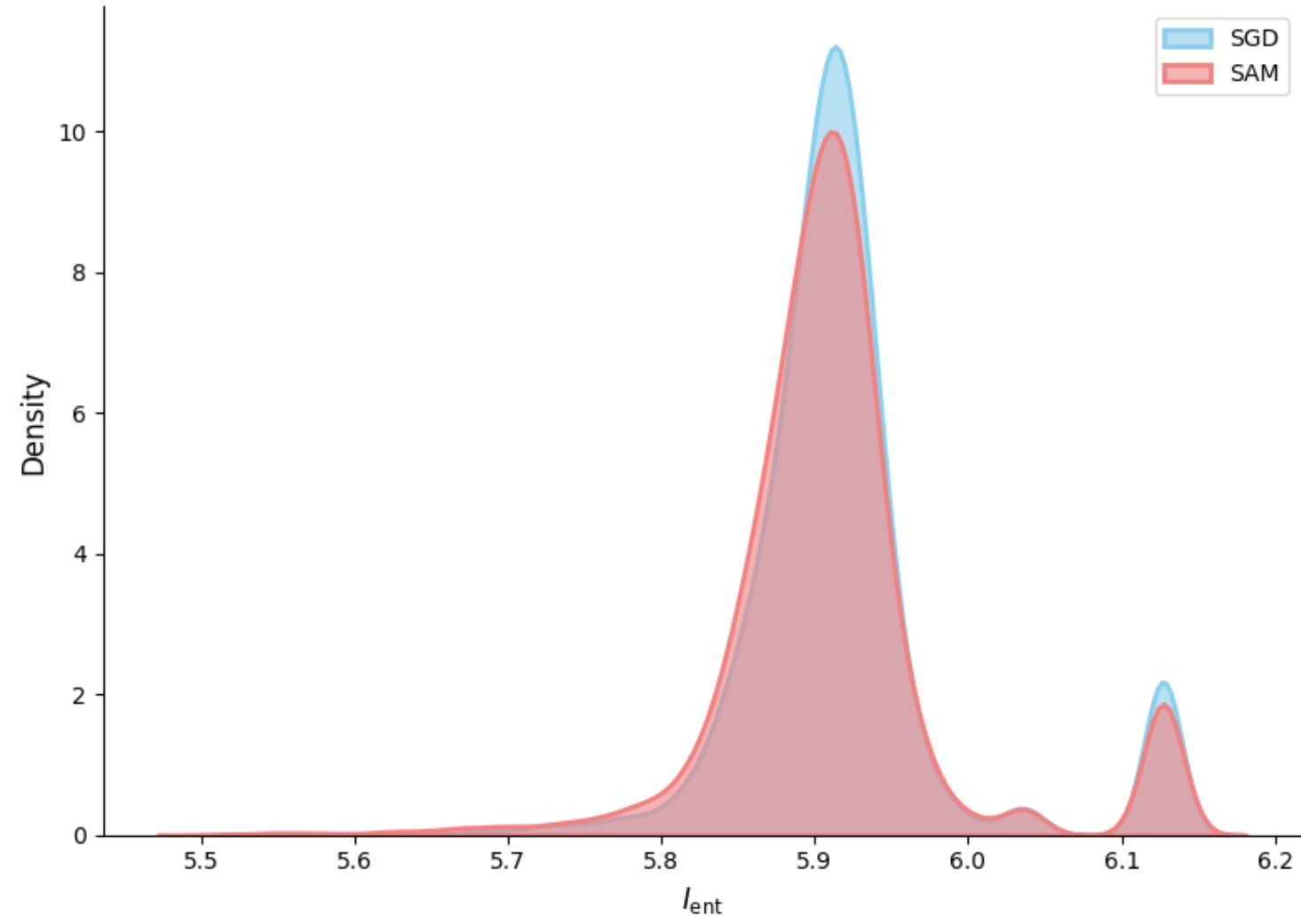}
  \caption{$\mathcal{I}_{ent}$ distribution excluding bucket 5 for SGD and SAM}\label{fig:entdist}
\end{figure}

\section{Experimental setup} \label{setup}
\subsection{Models} For CIFAR-100 and CIFAR-10, we use WideResNet (WRN) \citep{zagoruyko2016wide} with 16 layer depth and 8 as width factor. For Purchase-100 and Texas-100, we follow the setting in ~\citet{tang2022mitigating} and use a 4-layer fully connected neural network with layer sizes [1024, 512, 256, 100]. For EyePacs, we use ResNet-18.

\subsection{$\mathcal{I}_{ent}$ experiment}
Here we discuss how test data points were grouped into 5 buckets according to different $\mathcal{I}_{ent}$ levels. Bucket 5 contains highest $\mathcal{I}_{ent}$ level, and is composed of test points where all 500 training points have 0 influence score. This means that the prediction output for that test point does not change had the model been trained without any one particular training data point. Because influence scores for all training points are equal, these test points have highest $\mathcal{I}_{ent}$ \footnote{When actually calculating $\mathcal{I}_{ent}$ with \Cref{entr}, this evaluates to 0 due to probability normalization, but represents highest value.}. \Cref{fig:entdist} displays distribution of $\mathcal{I}_{ent}$ for remaining test data points. We group those above 6.1 into bucket 4. For the rest of the points, we calculate the mean and standard deviation and use them for grouping. We group points below $-0.4\sigma$ from the mean into bucket 1, points between $-0.4\sigma$ and $0.4\sigma$ into bucket 2, and points above $0.4\sigma$ into bucket 3. Final number of test points in each buckets are [Bucket 1: 1924, Bucket 2: 2996, Bucket 3: 2392, Bucket 4: 535, Bucket 5: 2153]. For SAM's buckets, final number of test points are [Bucket 1: 1913, Bucket 2: 3181, Bucket 3: 2548, Bucket 4: 502, Bucket 5: 1856]. Number of overlapping indices were [Bucket 1: 1116, Bucket 2: 1625, Bucket 3: 1199, Bucket 4: 133, Bucket 5: 1678].

\subsection{Attack setup \& size of data splits} We adopt the attack setting from \citep{tang2022mitigating, nasr2018machine} to determine the partition between training data and test data and to determine the subset of the training and test data that constitutes attacker’s prior knowledge for CIFAR-100, Purchase-100 and Texas-100 datasets. We use similar strategy to determine the data split for CIFAR-10. Specifically, the attacker’s knowledge corresponds to half of the training and test data, and the MIA success is evaluated over the remaining half. For shadow model attacks, the total sample pool used is 50000 for CIFAR10 and CIFAR100, 40000 for Purchase100, and 20000 for Texas100. For RMIA attack, we used $\gamma =1$ and selected all of the $z$ samples within the training pool that were not part of the target model's training set. On CIFAR10, for example, number of $z$ samples was 25000.

\subsection{Hyperparameter tuning and empirical validation of flatness for Sharp}
\label{sharpdetail}
For the Sharp objective, we fine-tuned $\beta$ and $\rho$ that resulted in a model that exhibited sufficient difference in test accuracy and sharpness of the minima compared to SAM and SGD. The final hyperparameters of the model reported were $\rho = 0.01, \beta=0.6818$ for CIFAR-100 and CIFAR-10, $\rho = 0.01, \beta=0.83$ for Purchase-100, $\rho = 0.001, \beta=0.513$ for Texas-100, and $\rho = 0.001, \beta=0.18$ for EyePacs.

To verify that Sharp actually finds a sharper minima, we computed the trace of the hessian matrix using Hutchinson's method for SGD, SAM, and Sharp models on CIFAR-100. The results are in \Cref{tab:hessian-trace}. Higher trace indicates a sharper minima and vice versa. The trace is the largest for Sharp and smallest for SAM. 

\begin{table}[h]
\centering
\begin{tabular}{lc}
\hline
\textbf{Method} & \textbf{Trace of the Hessian} \\
\hline
Sharp & 1556.54 \\
SGD  & 307.87  \\
SAM  & 84.18   \\
\hline
\end{tabular}
\caption{Comparison of Hessian trace values across methods.}
\label{tab:hessian-trace}
\end{table}

\subsubsection{Ball of radius $\rho$}
For SAM loss, sharp minima loss, and our proposed loss, we approximate the maximum loss in the ball of radius $\rho$ around the minima. \cite{norton2021diametrical} have found that the type of norm that is used for defining the ball has large impact along with actual $\rho$ value. For all our experiments, we use L2 norm for our ball of radius $\rho$.

\subsubsection{Hyperparameter tuning for CIFAR-10 \& CIFAR-100}
We trained each model for 200 epochs and chose the model with highest validation accuracy on a held-out validation set. We used initial learning rate of 0.1 with learning rate decay of 0.2 at 60th, 120th, and 160th epoch with batch size of 128. We trained the models with weight decay 0.0005 and Nesterov momentum of 0.9. For SWA on CIFAR-100, we trained first 150 epoch with vanilla SGD and used weight averaging for the rest of the epochs. \\

\subsubsection{Hyperparameter tuning for Texas-100 \& Purchase-100}
We chose the best model as discussed before for CIFAR-10/100. We trained models with a learning rate of 0.1 with weight decay 0.0005 and Nesterov momentum of 0.9. We trained the models on Purchase-100 for a total of 100 epochs and on Texas-100 for a total on 75 epochs. During training, we employed a batch size of 512 for the Purchase-100 dataset and a batch size of 128 for the Texas-100 dataset.

\subsubsection{Hyperparameter tuning for EyePacs}
We trained ResNet-18 with SGD, SAM and our proposed loss using EyePacs dataset for 100 epochs. Since, the dataset is highly imbalanced with about 25k data points out of 35k training data points belonging to one of the five classes, we used the balanced batch sampling strategy and a lower learning rate of 0.01 with learning rate decay of 0.2 at 60th epoch. As before, we also used weight decay 0.0005 and Nesterov momentum of 0.9.  For our experiments, we utilized a batch size of 100, consisting of 12 samples from each of the 5 classes.

\section{Ablation Study: Comparison of different architectures} \label{diffarch}
To validate consistency across different model architectures, we report direct threshold attack results in \Cref{diffarchitectures} using InceptionV4 \footnote{https://github.com/weiaicunzai/pytorch-cifar100/blob/master/models/} and resnet18 \footnote{https://github.com/inspire-group/MIAdefenseSELENA/tree/main} for CIFAR-100 and CIFAR-10. We kept our $\rho$ the same across all model architectures with value 0.1. The results are consistent with our findings that SAM tends to have higher test accuracy while having higher membership attack accuracy at the same time. Overall best attack accuracy is higher for SAM for all the cases although we find mixed findings for multi-query attack accuracy specifically.

\begin{table}

  \caption{Privacy vs Generalization tradeoff for SAM and SGD using InceptionV4 and Resnet18 }
  \centering
  \label{diffarchitectures}
  \begin{tabular}{ccccc}
    \toprule 
Dataset     & Model & Optimizer &Test Acc  & Best Attack Acc     \\
    \midrule
    \multirow{4}{*}{CIFAR-100}  & \multirow{2}{*}{Resnet18} & SGD     &78.42\%    & 74.31\%   \\
               &       & SAM     &78.74\%    & 77.45\%     \\

               & \multirow{2}{*}{InceptionV4} & SGD     &77.44\%    & 77.22\%    \\
               &       & SAM     &79.60\%    & 80.82\%    \\

    \midrule
    \multirow{4}{*}{CIFAR-10}   & \multirow{2}{*}{Resnet18} & SGD     &95.18\%    & 57.90\%    \\
               &       & SAM     &96.16\%    & 60.05\%     \\

               & \multirow{2}{*}{InceptionV4} & SGD     &94.26\%  &61.60\%   \\   
               &       & SAM     &95.76\%    & 64.41\%     \\

    \bottomrule
  \end{tabular}
\end{table}

\begin{figure}[p]
    \centering
    \begin{minipage}{0.24\textwidth}
        \centering
        \includegraphics[width=\linewidth]{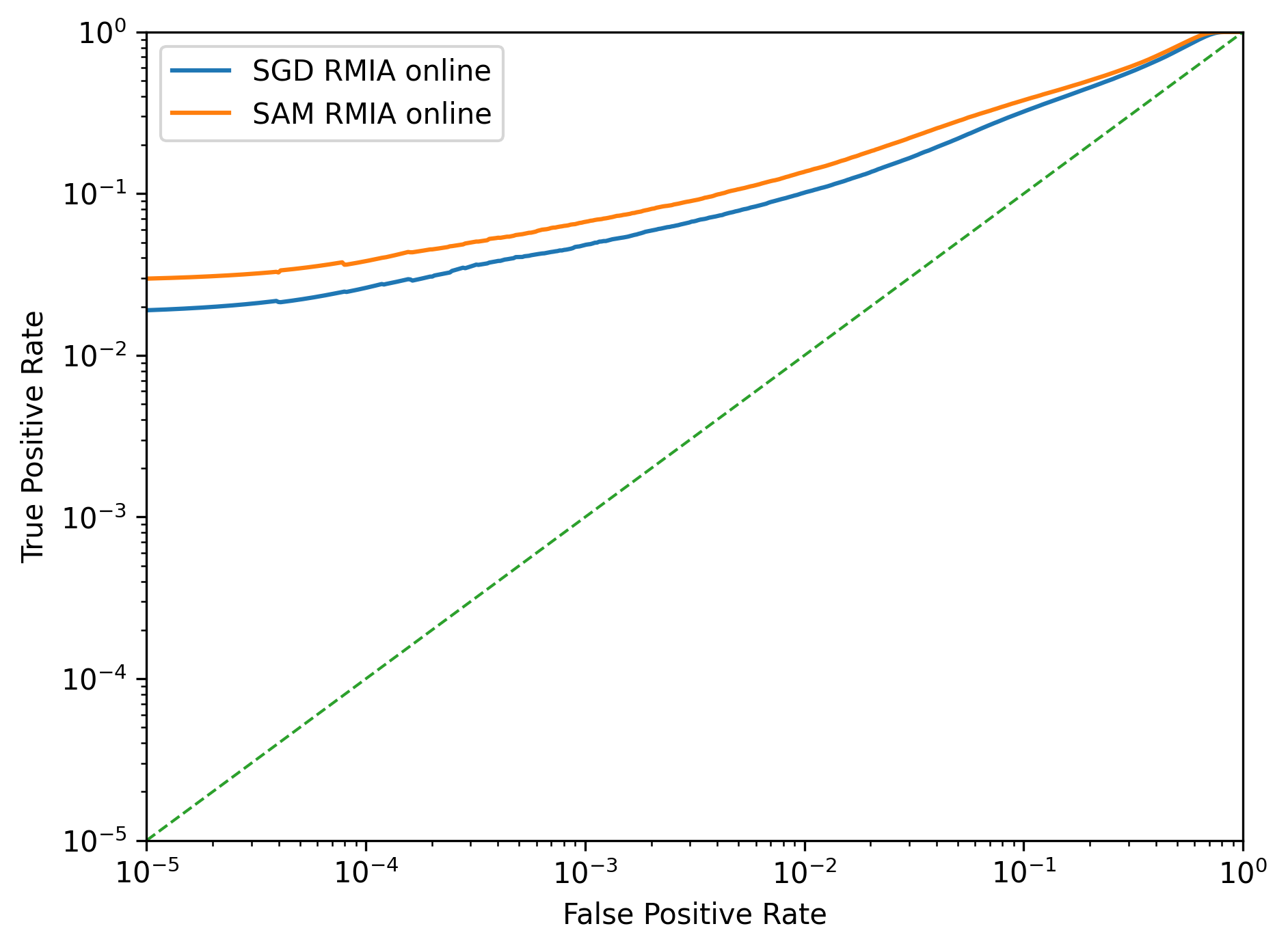}
        \par\vspace{2pt}\tiny C10: RMIA (Online)
    \end{minipage}
    \hfill
    \begin{minipage}{0.24\textwidth}
        \centering
        \includegraphics[width=\linewidth]{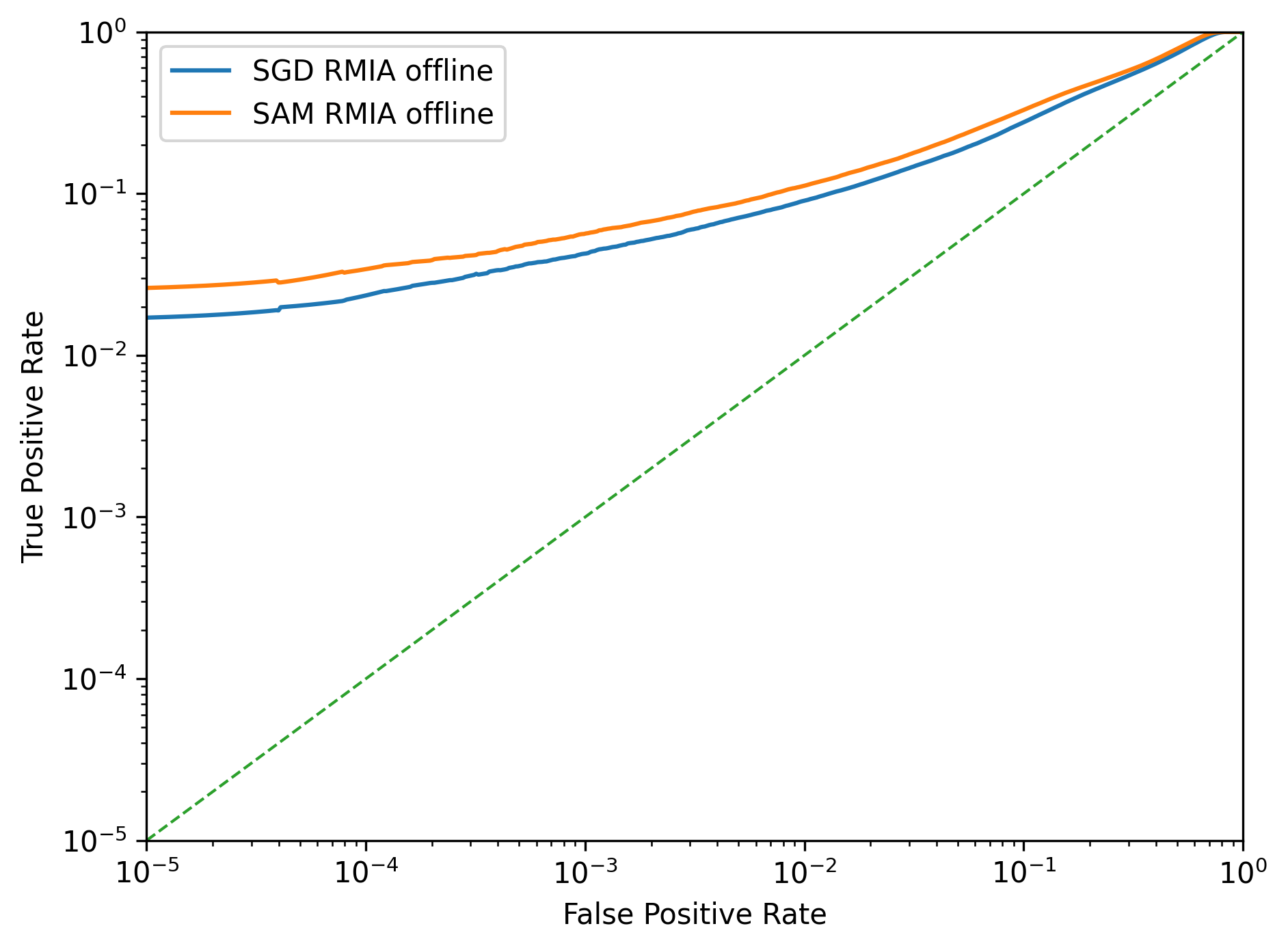}
        \par\vspace{2pt}\tiny C10: RMIA (Offline)
    \end{minipage}
    \hfill
    \begin{minipage}{0.24\textwidth}
        \centering
        \includegraphics[width=\linewidth]{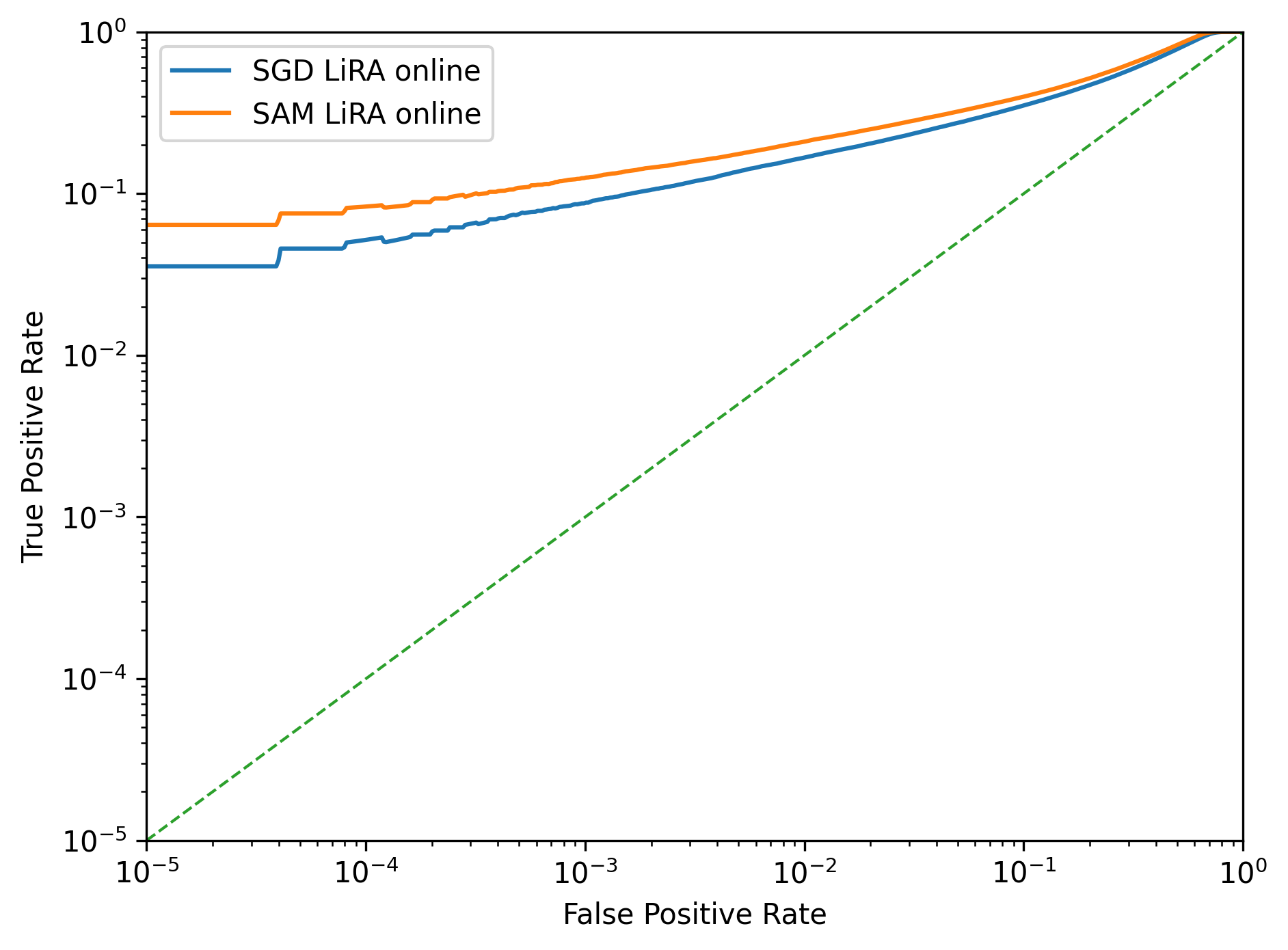}
        \par\vspace{2pt}\tiny C10: LiRA (Online)
    \end{minipage}
    \hfill
    \begin{minipage}{0.24\textwidth}
        \centering
        \includegraphics[width=\linewidth]{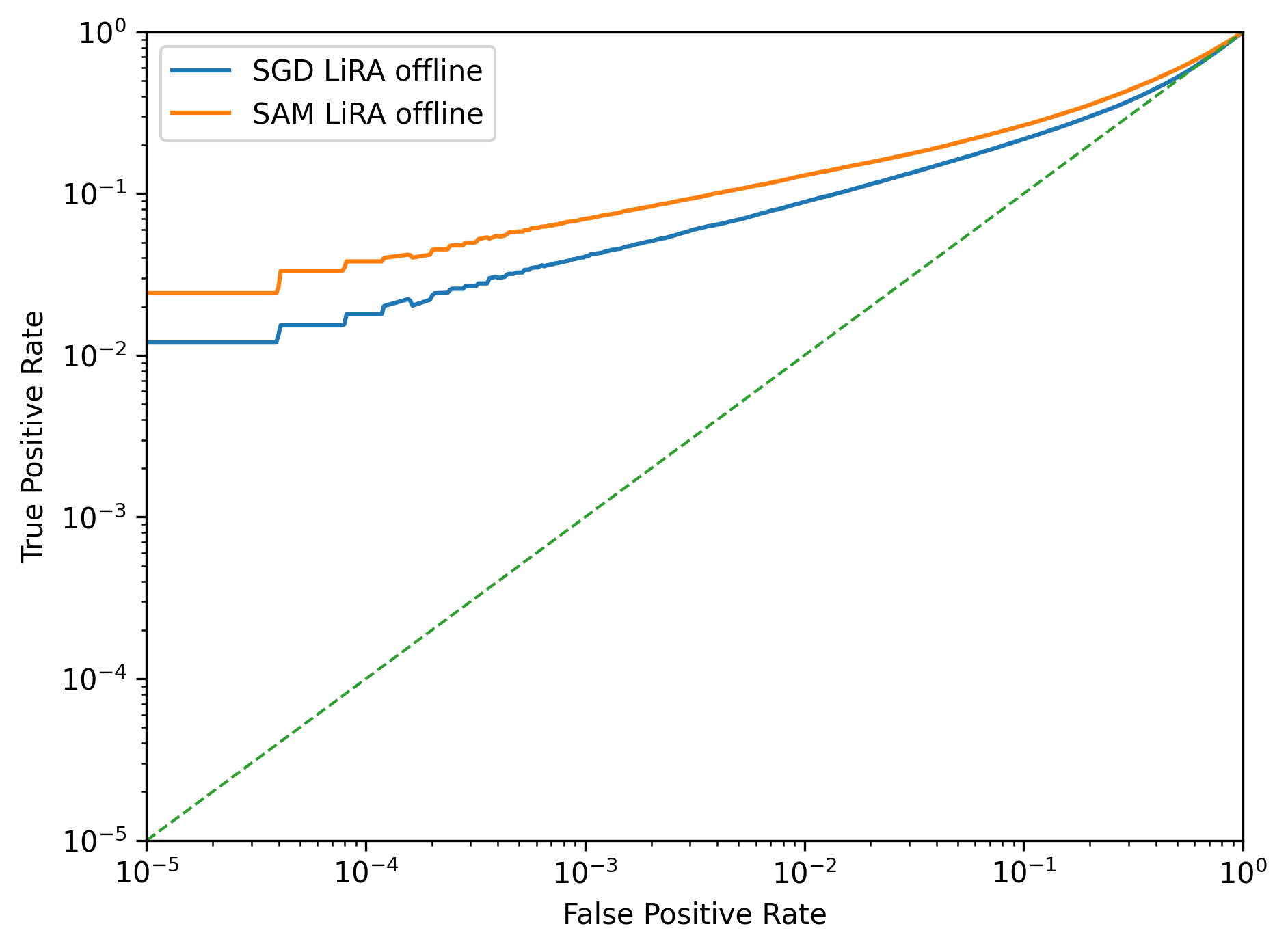}
        \par\vspace{2pt}\tiny C10: LiRA (Offline)
    \end{minipage}

    \vspace{1em} 

    \begin{minipage}{0.24\textwidth}
        \centering
        \includegraphics[width=\linewidth]{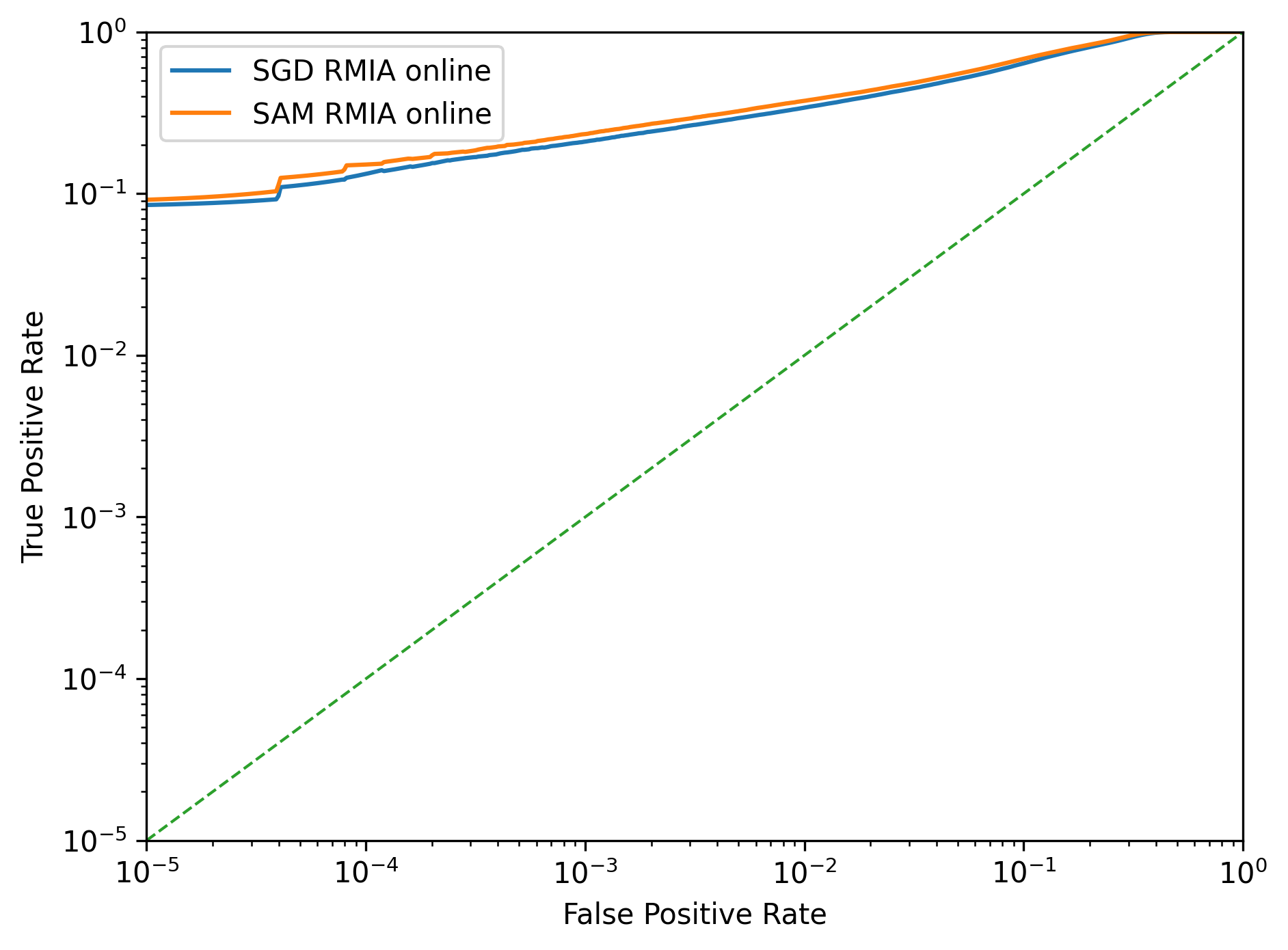}
        \par\vspace{2pt}\tiny C100: RMIA (Online)
    \end{minipage}
    \hfill
    \begin{minipage}{0.24\textwidth}
        \centering
        \includegraphics[width=\linewidth]{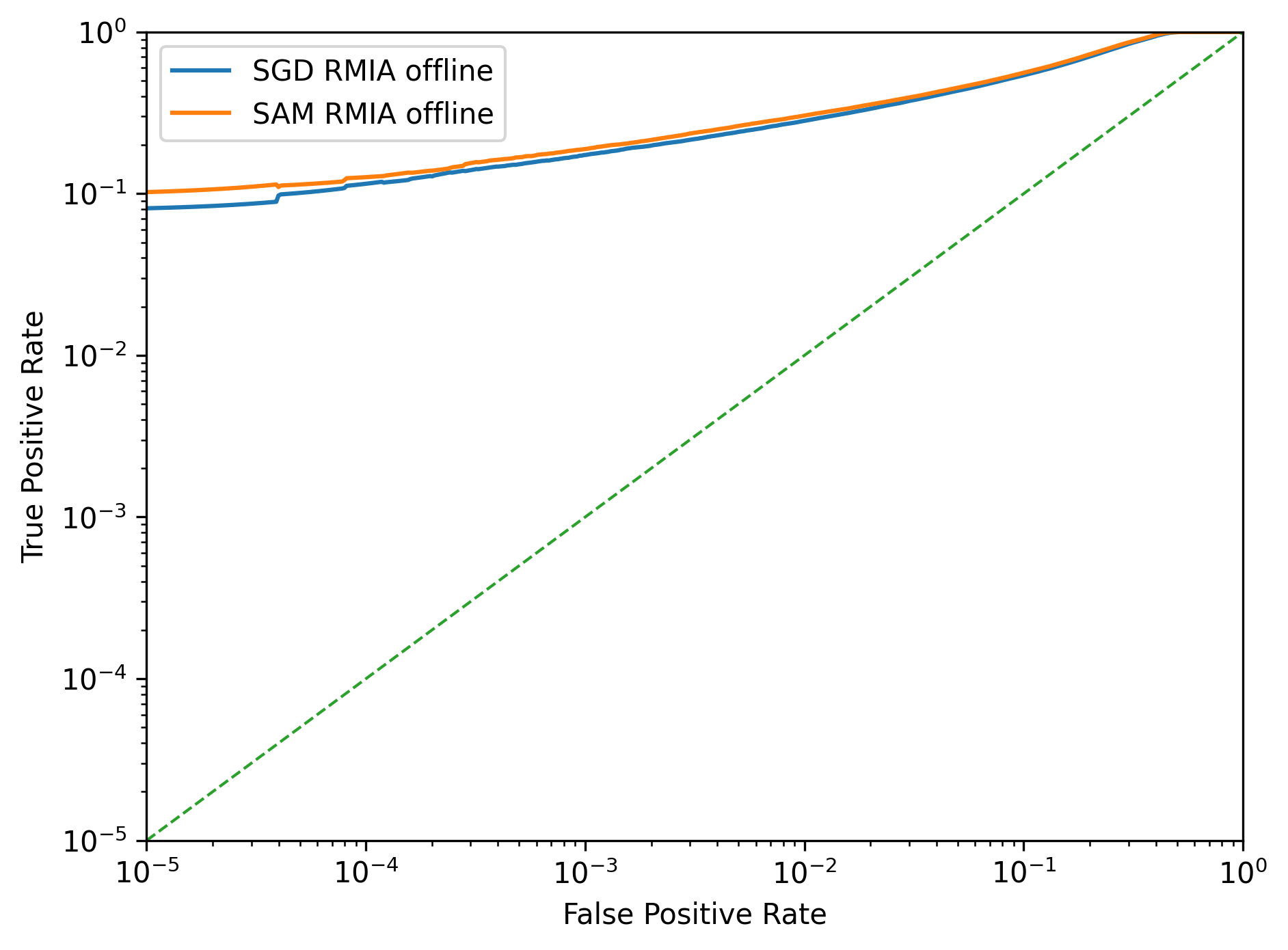}
        \par\vspace{2pt}\tiny C100: RMIA (Offline)
    \end{minipage}
    \hfill
    \begin{minipage}{0.24\textwidth}
        \centering
        \includegraphics[width=\linewidth]{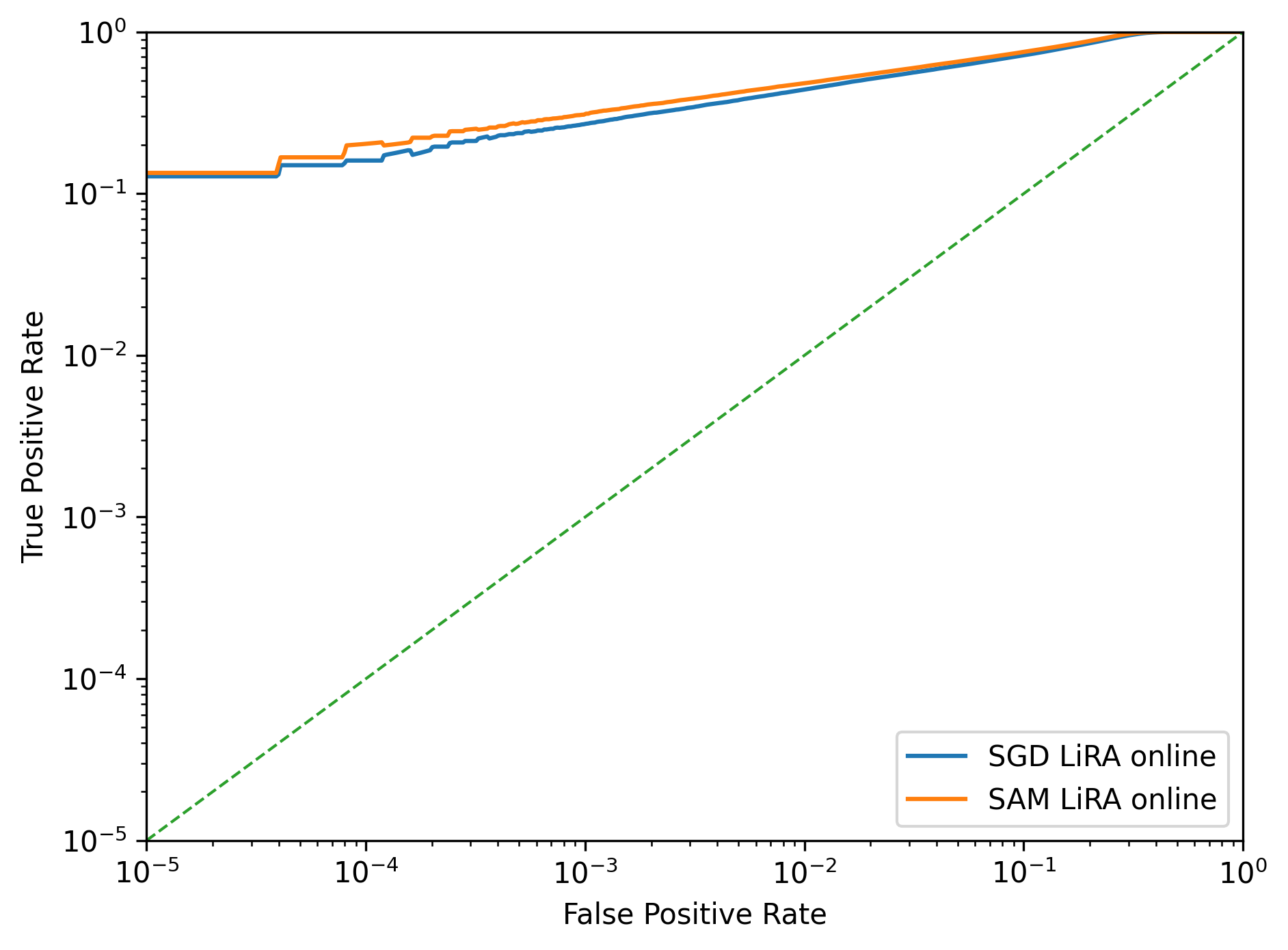}
        \par\vspace{2pt}\tiny C100: LiRA (Online)
    \end{minipage}
    \hfill
    \begin{minipage}{0.24\textwidth}
        \centering
        \includegraphics[width=\linewidth]{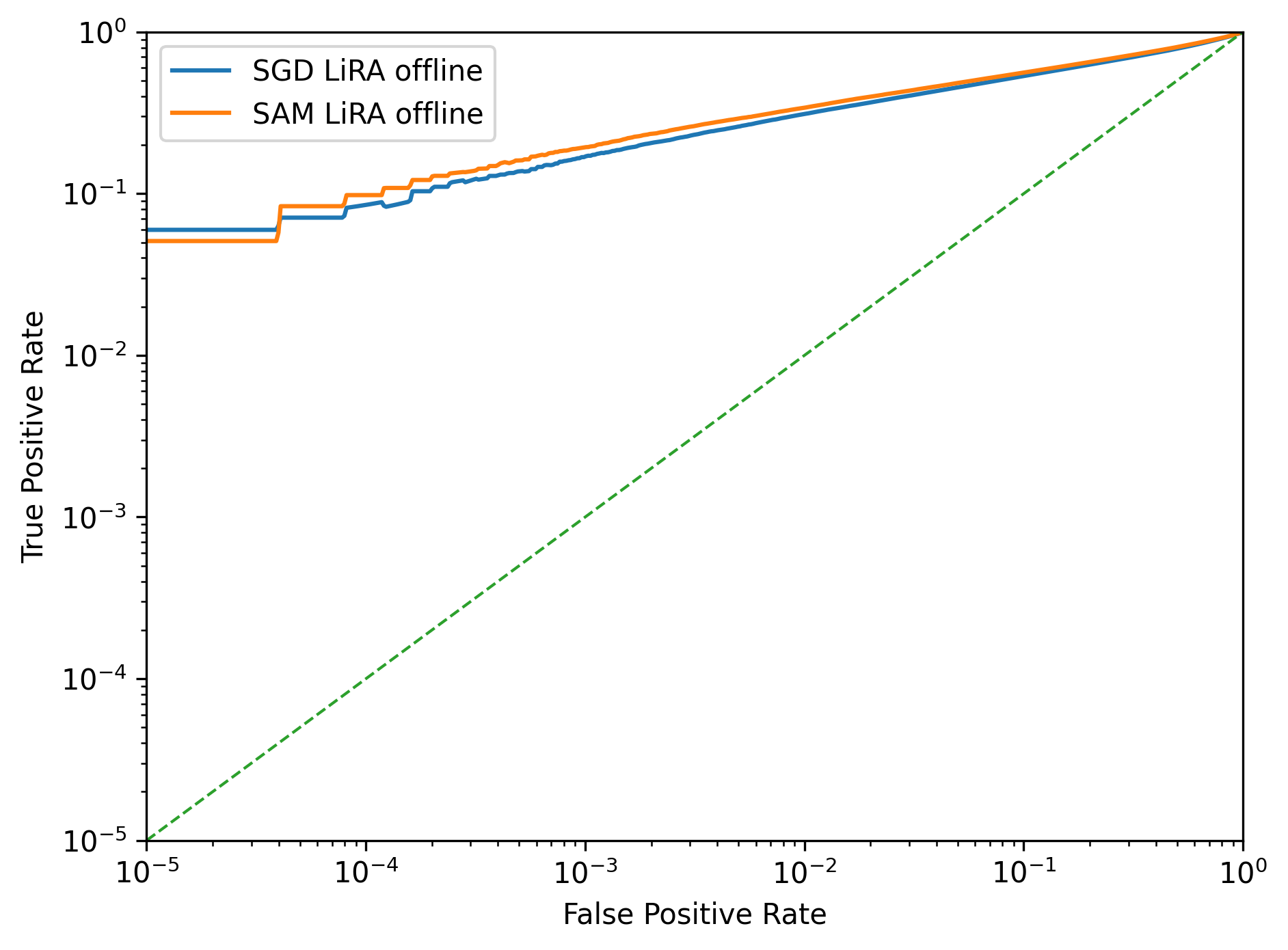}
        \par\vspace{2pt}\tiny C100: LiRA (Offline)
    \end{minipage}

    \vspace{1em}

    \begin{minipage}{0.24\textwidth}
        \centering
        \includegraphics[width=\linewidth]{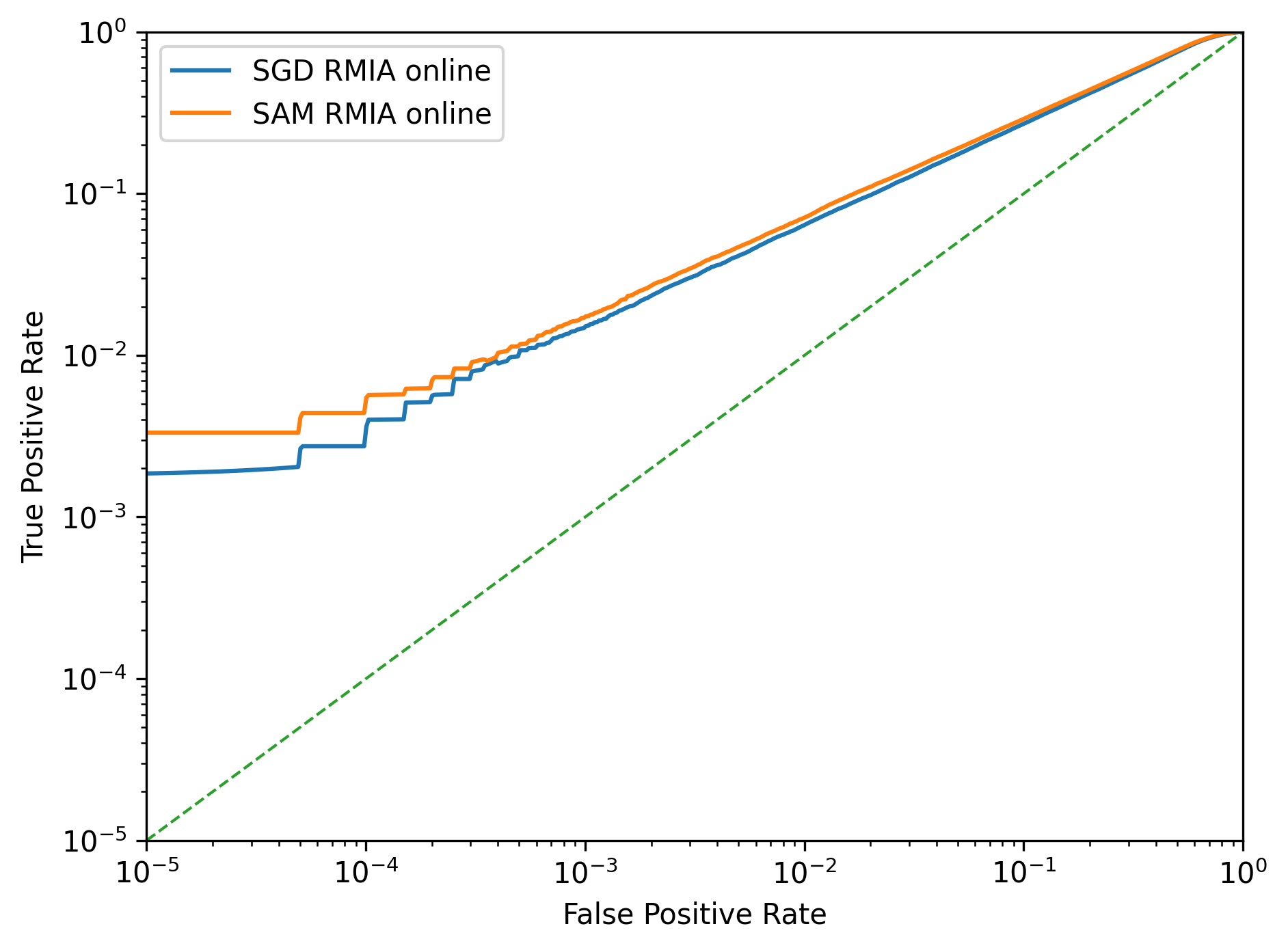}
        \par\vspace{2pt}\tiny Purch: RMIA (Online)
    \end{minipage}
    \hfill
    \begin{minipage}{0.24\textwidth}
        \centering
        \includegraphics[width=\linewidth]{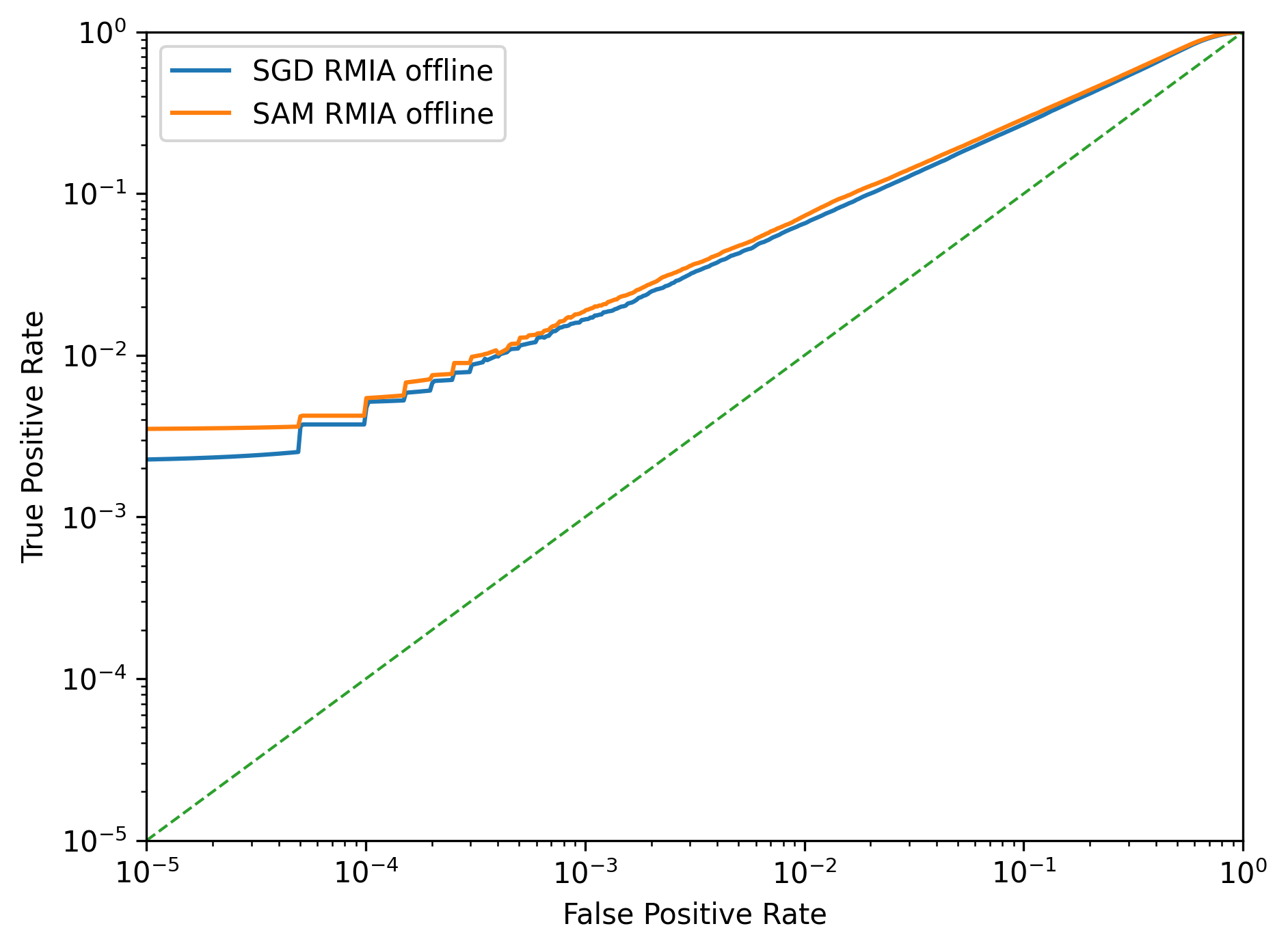}
        \par\vspace{2pt}\tiny Purch: RMIA (Offline)
    \end{minipage}
    \hfill
    \begin{minipage}{0.24\textwidth}
        \centering
        \includegraphics[width=\linewidth]{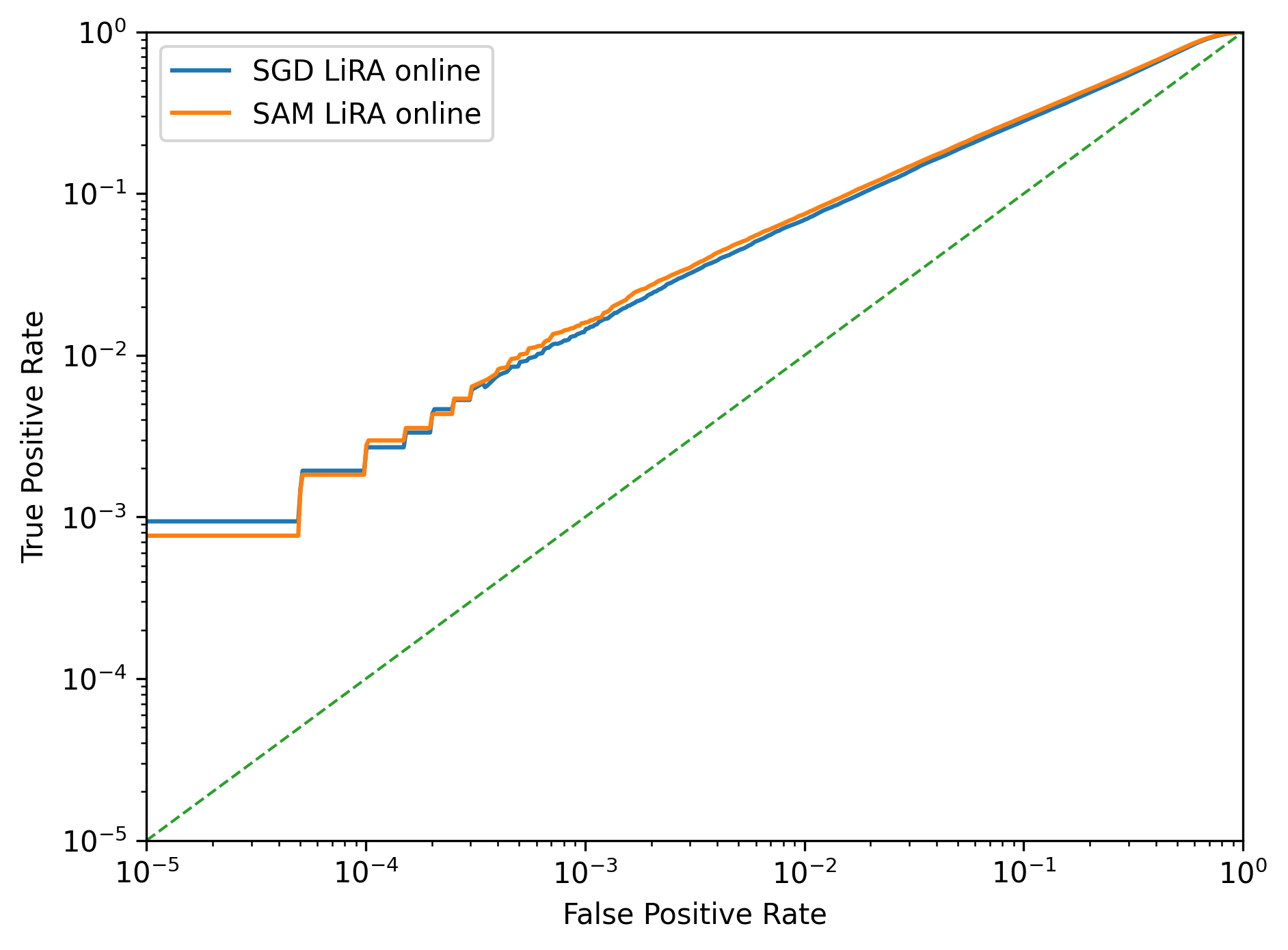}
        \par\vspace{2pt}\tiny Purch: LiRA (Online)
    \end{minipage}
    \hfill
    \begin{minipage}{0.24\textwidth}
        \centering
        \includegraphics[width=\linewidth]{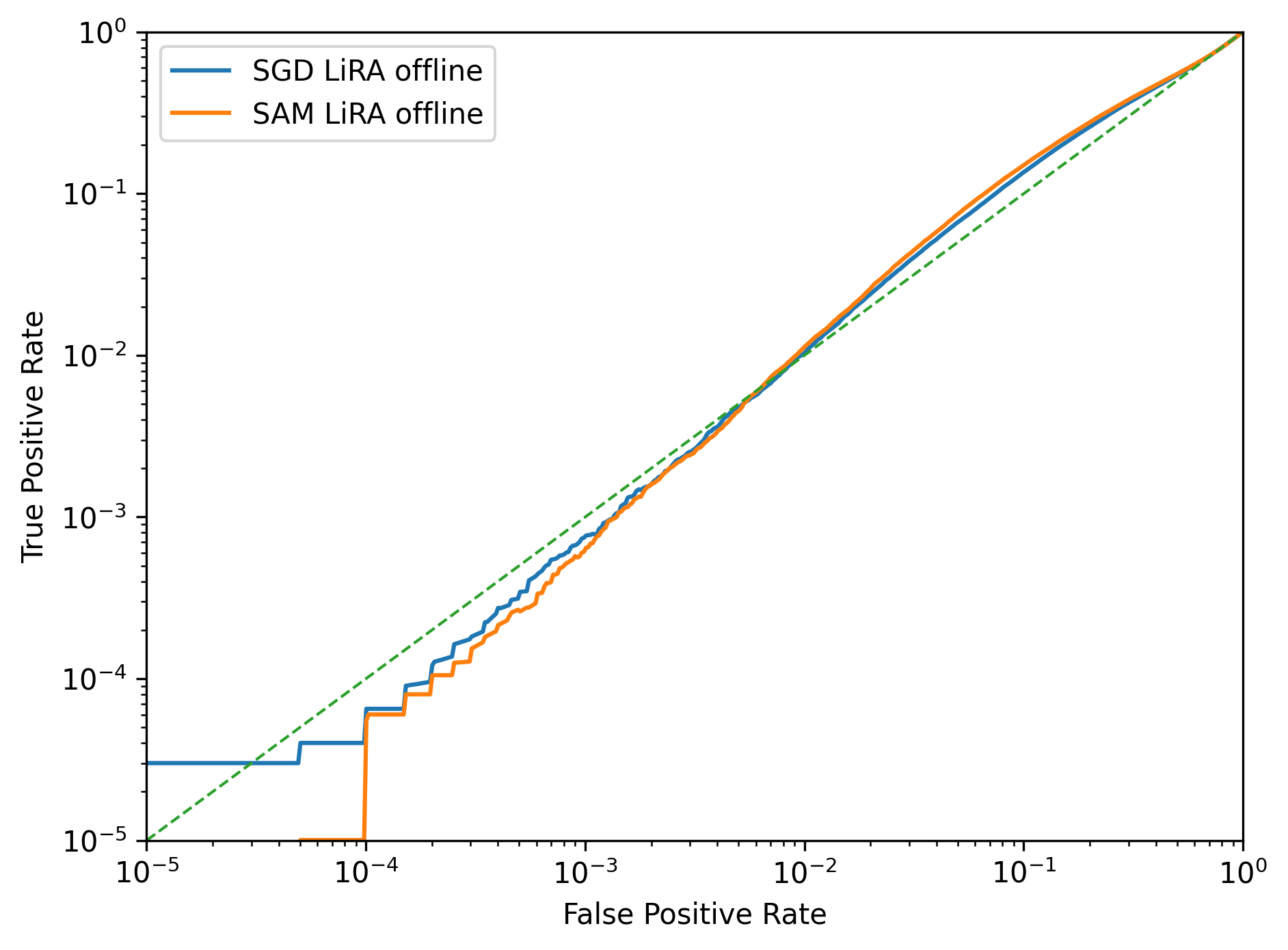}
        \par\vspace{2pt}\tiny Purch: LiRA (Offline)
    \end{minipage}

    \vspace{1em}

    \begin{minipage}{0.24\textwidth}
        \centering
        \includegraphics[width=\linewidth]{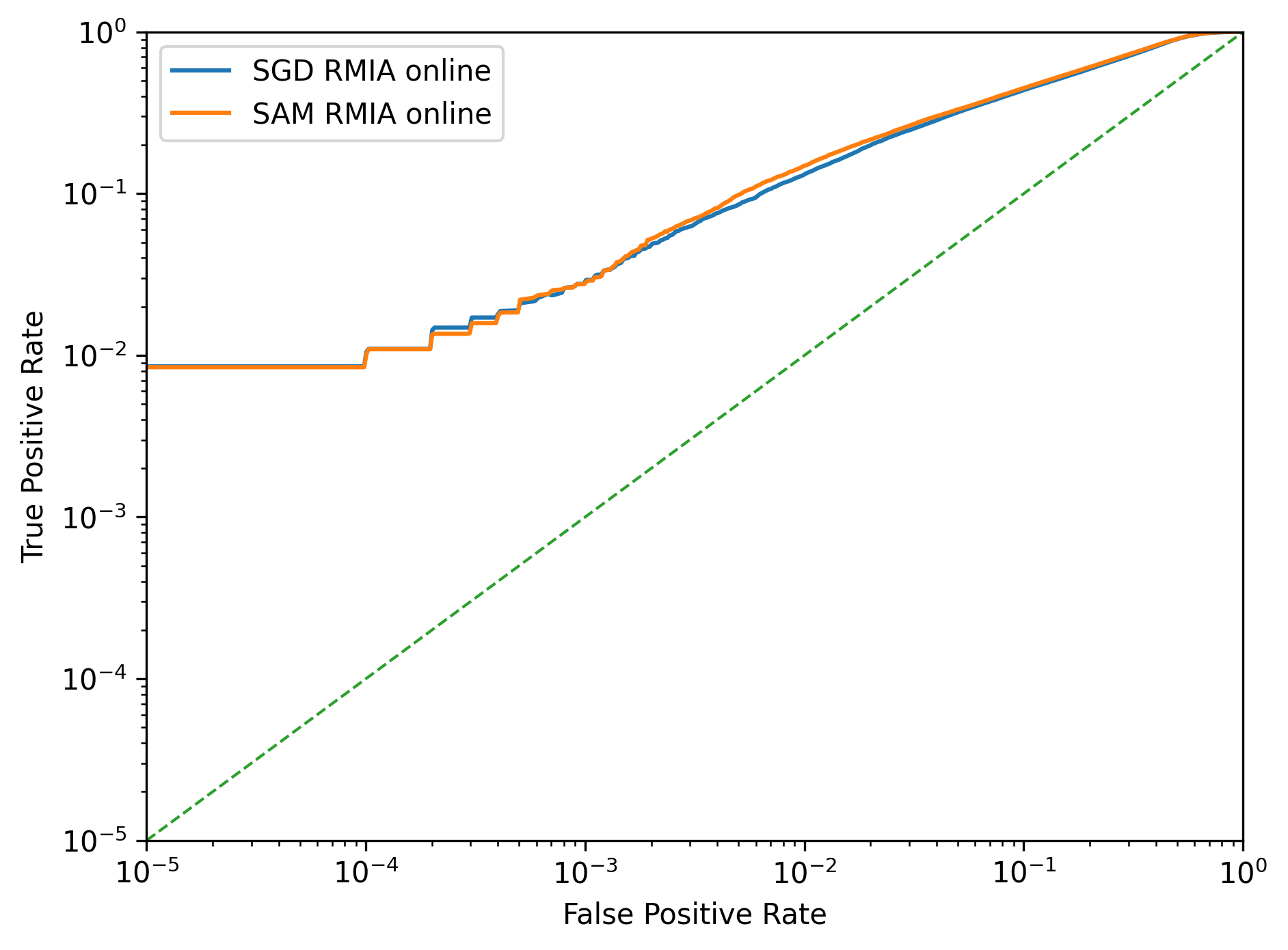}
        \par\vspace{2pt}\tiny Texas: RMIA (Online)
    \end{minipage}
    \hfill
    \begin{minipage}{0.24\textwidth}
        \centering
        \includegraphics[width=\linewidth]{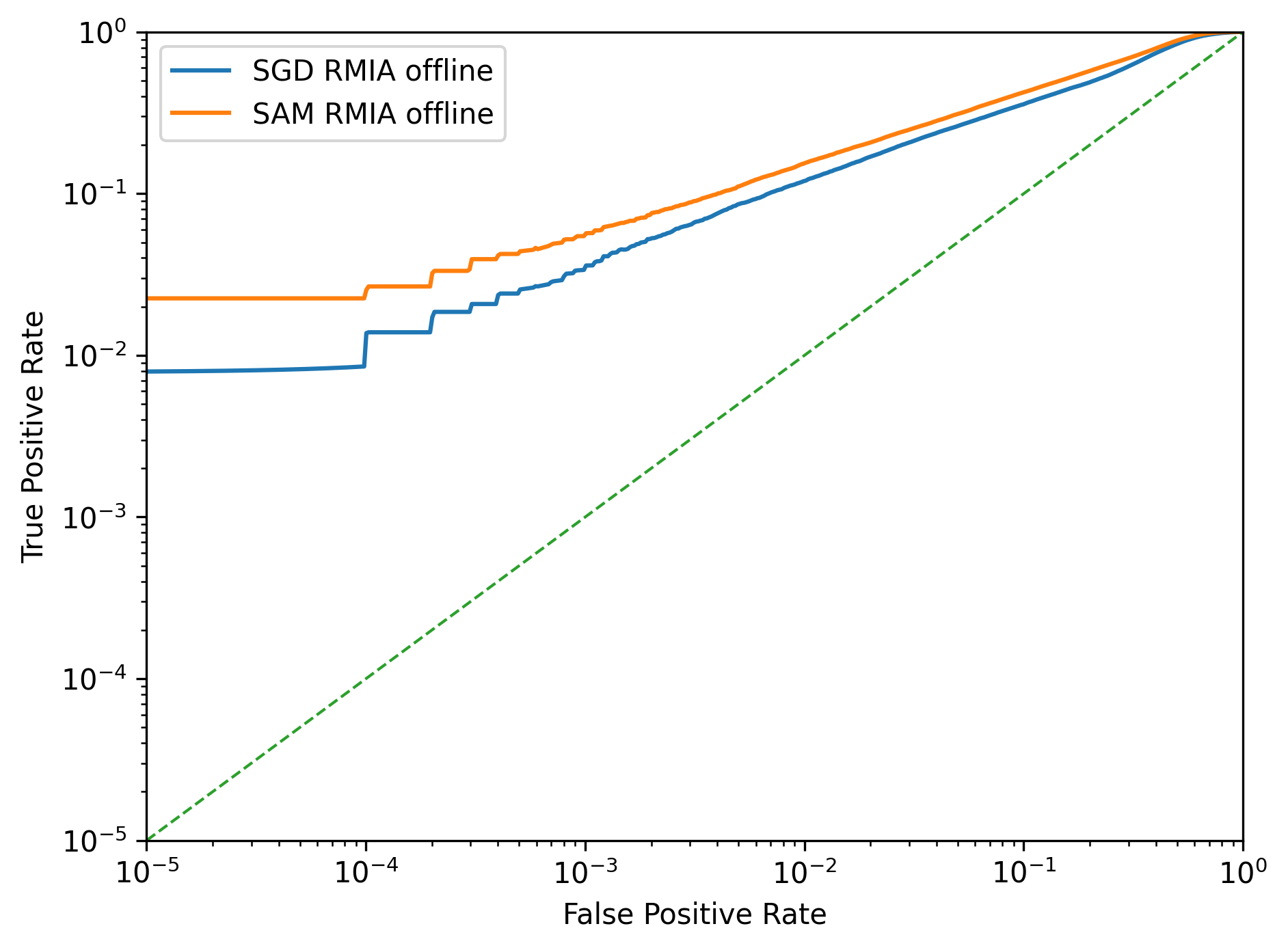}
        \par\vspace{2pt}\tiny Texas: RMIA (Offline)
    \end{minipage}
    \hfill
    \begin{minipage}{0.24\textwidth}
        \centering
        \includegraphics[width=\linewidth]{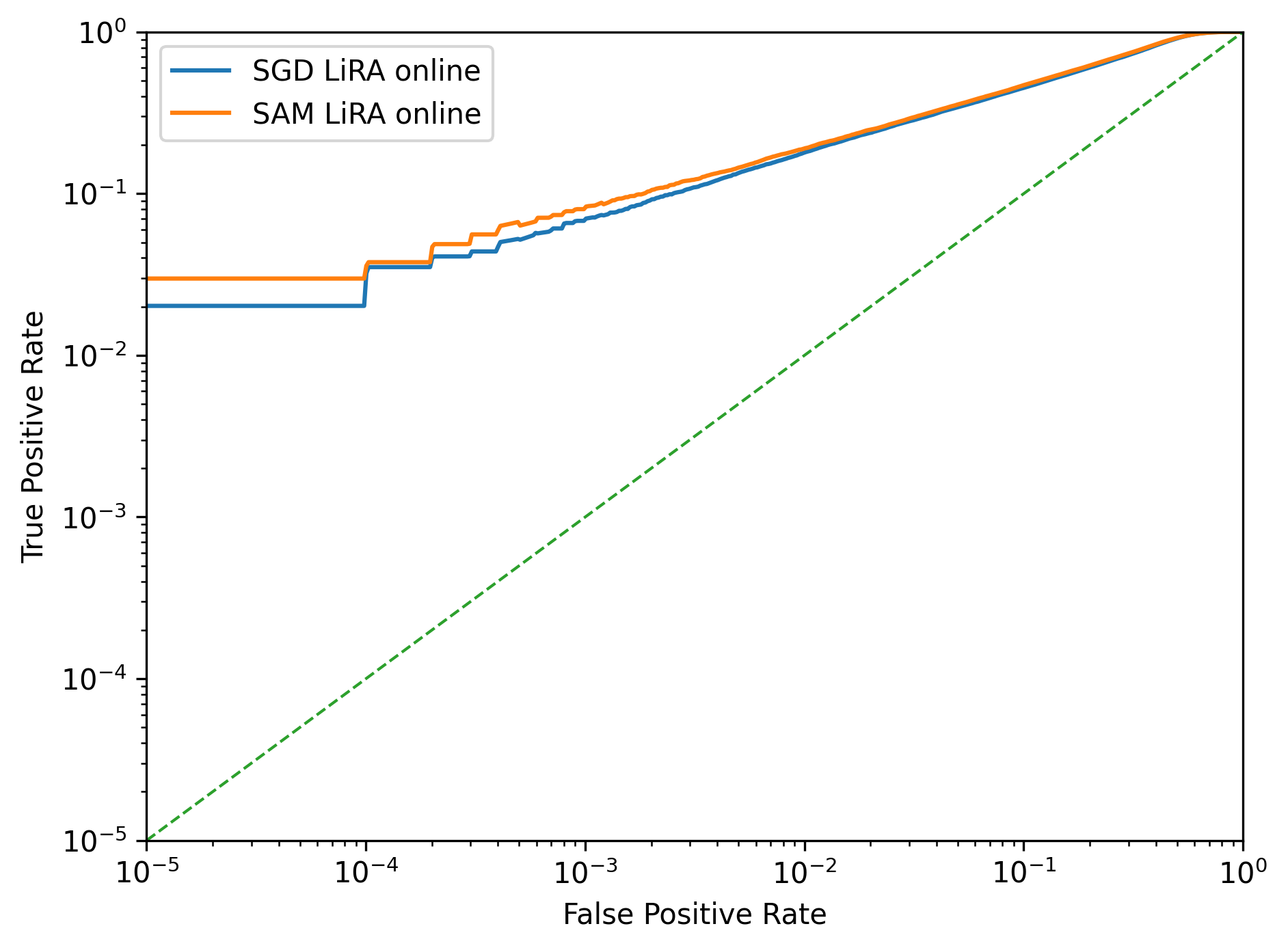}
        \par\vspace{2pt}\tiny Texas: LiRA (Online)
    \end{minipage}
    \hfill
    \begin{minipage}{0.24\textwidth}
        \centering
        \includegraphics[width=\linewidth]{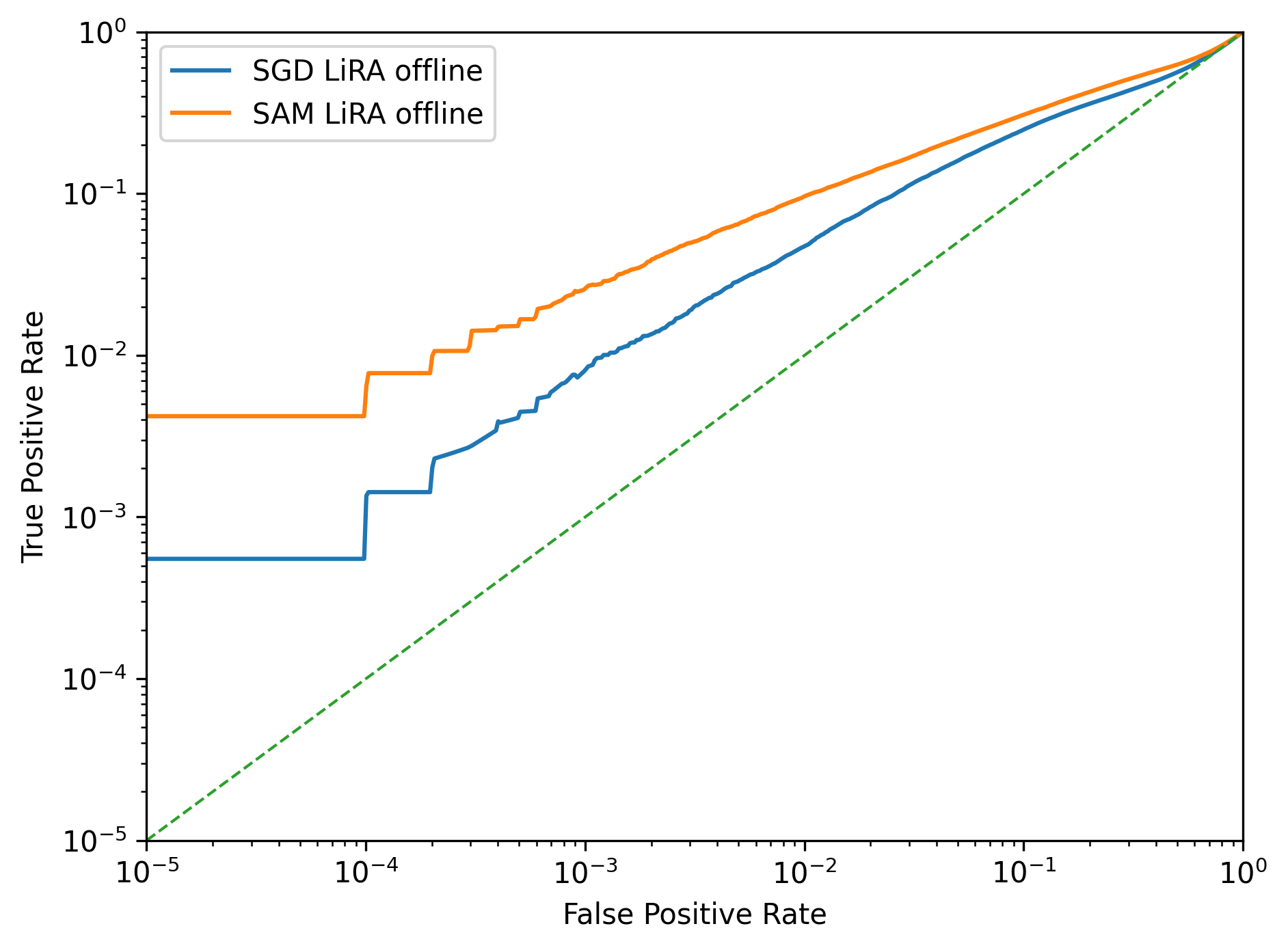}
        \par\vspace{2pt}\tiny Texas: LiRA (Offline)
    \end{minipage}

    \caption{ROC curves comparing SAM (Orange) vs. SGD (Blue) across all datasets and attack modes on log-log scale. Rows represent datasets (CIFAR-10, CIFAR-100, Purchase100, Texas100). Columns represent the attack configuration. The ROC curve for SAM (orange) is above the ROC curve for SGD (blue) for nearly the entire range for most settings.}
    \label{fig:roc}
\end{figure}

\end{document}